\documentclass[10pt,twocolumn,letterpaper]{article}

\usepackage[pagenumbers]{iccv} 

\definecolor{iccvblue}{rgb}{0.21,0.49,0.74}
\usepackage[pagebackref,breaklinks,colorlinks,allcolors=iccvblue]{hyperref}

\usepackage{amsthm}
\usepackage{amsmath}
\usepackage{amssymb}
\usepackage{subcaption}
\usepackage{multirow}
\usepackage{algorithm}
\usepackage[switch]{lineno}

\usepackage{enumitem}
\setlist[itemize]{align=parleft,left=0pt..1em}

\theoremstyle{definition}

\usepackage[noend]{algpseudocode}

\usepackage{pifont}
\newcommand{\cmark}{\ding{51}}%
\newcommand{\xmark}{\ding{55}}%

\usepackage[accsupp]{axessibility}  

\def\confName{ICCV}
\def\confYear{2025}

\title{Class-Wise Federated Averaging for Efficient Personalization}

\author{Gyuejeong Lee\\
SAKAK Inc.\\
{\tt\small regulation.lee@sakak.co.kr}
\and
Daeyoung Choi\thanks{Corresponding author}\\
The Cyber University of Korea\\
{\tt\small choidy@cuk.edu}
}

\begin{document}
\maketitle
\begin{abstract}
Federated learning (FL) enables collaborative model training across distributed clients without centralizing data. However, existing approaches such as Federated Averaging (\texttt{FedAvg}) often perform poorly with heterogeneous data distributions, failing to achieve personalization owing to their inability to capture class-specific information effectively.
We propose Class-wise Federated Averaging (\texttt{cwFedAvg}), a novel personalized FL (PFL) framework that performs Federated Averaging for each class, to overcome the personalization limitations of \texttt{FedAvg}.
\texttt{cwFedAvg} creates class-specific global models via weighted aggregation of local models using class distributions, and subsequently combines them to generate personalized local models.
We further propose Weight Distribution Regularizer (\texttt{WDR}), which encourages deep networks to encode class-specific information efficiently by aligning empirical and approximated class distributions derived from output layer weights, to facilitate effective class-wise aggregation.
Our experiments demonstrate the superior performance of \texttt{cwFedAvg} with \texttt{WDR} over existing PFL methods through efficient personalization while maintaining the communication cost of \texttt{FedAvg} and avoiding additional local training and pairwise computations.
\end{abstract}

\section{Introduction} \label{sec:intro}
Deep networks demand large-scale training data for superior performance \citep{lecun2015deep}; however, data collection faces significant challenges due to prohibitive costs and privacy constraints \citep{kairouz2021advances}. This limitation necessitates developing communication-efficient and privacy-preserving approaches to effectively utilize distributed data from sources such as data silos and edge devices \citep{yang2019federated,li2020federated}.
Federated learning (FL) addresses this challenge by enabling client collaboration through model aggregation without collecting client data \citep{mcmahan2017communication,kairouz2021advances}. A foundational approach, Federated Averaging (\texttt{FedAvg}) \citep{mcmahan2017communication}, creates a single global model by aggregating local models weighted by client sample sizes. However, this approach performs poorly with non-independent and identically distributed (non-IID) data due to its lack of personalization capability \citep{zhao2018federated,li2020federated,hsieh2020non}.
This limitation stems from how deep networks encode information. Deep networks develop pathways, where a pathway represents a union of weights from input to output \citep{khakzar2021neural}, and encode class-specific information in these pathways \citep{anand1993improved,khakzar2021neural,kang2020decoupling}. Pathways demonstrate distinct patterns across different classes based on the class proportion \citep{wang2018interpret,yu2018distilling,qiu2019adversarial}. However, \texttt{FedAvg}'s aggregation weighting factor, which only considers client sample sizes, fails to reflect class-specific pathways, limiting its personalization capability.

This study addresses the personalization limitation of \texttt{FedAvg}, motivated by class-specific pathways. We change the aggregation weighting factor of \texttt{FedAvg} using the class distributions of clients and create multiple global models specialized for each class using an adapted weighting factor. Based on this, we propose Class-wise Federated Averaging (\texttt{cwFedAvg}), a class-wise extension of \texttt{FedAvg} that performs Federated Averaging separately for each class. \texttt{cwFedAvg} implements a two-step aggregation process: first, we create \textit{class-specific global models} by aggregating local models weighted by their respective client and class sample proportions; then, we generate personalized local models by aggregating these class-specific global models weighted by the class distribution of each client.

We identify two requirements to further improve the effectiveness of \texttt{cwFedAvg}: model weights (pathways) must strongly correlate with empirical class distributions to capture class-specific information effectively, and privacy-sensitive class distributions must be securely shared with the server.
We analyze how empirical class distribution affects pathways and propose Weight Distribution Regularizer (\texttt{WDR}) to address these requirements. \texttt{WDR} strengthens the correlation between empirical class distributions and model weights by minimizing the distance between the empirical and approximated class distributions derived from the $\ell_2$-norms of output layer weight vectors. Additionally, it enables privacy-preserving sharing of class distributions using the approximated distribution as a proxy.
Finally, we extend the proposed approach by applying \texttt{cwFedAvg} selectively to upper layers, thereby reducing memory requirements while maintaining high performance.

Our work focuses on learning individualized models, one of the main approaches in personalized FL (PFL) \citep{tan2022towards}. This approach achieves personalization by modifying the model aggregation process, enabling adaptation to the unique data distribution of each client without post-training steps—a crucial advantage for resource-constrained devices \citep{zhang2023fedala}. Although the proposed method is a straightforward extension of \texttt{FedAvg}, it effectively addresses key limitations of existing learning individualized model approaches. Specifically, current methods either incur substantial overhead, such as weighted model aggregation \citep{luo2022adapt,zhang2020personalized} and client-pair collaboration \citep{huang2021personalized}, or rely on strong assumptions about client group structures  \citep{sattler2020clustered,ghosh2020efficient,briggs2020federated,duan2021fedgroup}. Our proposed \texttt{cwFedAvg} delivers efficient personalization without requiring additional model downloads, pairwise computations, or clustering assumptions.

Our contributions and benefits of our approach are summarized as follows:

\noindent \textbf{Our Contributions.} 
\begin{itemize}[itemsep=-0.25em, topsep=-0.25em]
\item We propose \texttt{cwFedAvg}, a novel PFL framework that performs Federated Averaging per class to create personalized models through class-specific global models.
\item We develop \texttt{WDR} to enable effective class-wise model aggregation and ensure secure class distribution estimation.
\item Our extensive experiments on four datasets and various levels of data heterogeneity demonstrate \texttt{cwFedAvg}'s superior performance over existing PFL methods and provide insights into personalization through visualization of output layer weight distributions and pathways.
\end{itemize}
\vspace{5pt}
\noindent \textbf{Benefits of Our Approach.} 
\begin{itemize}[itemsep=-0.25em, topsep=-0.25em]
\item Efficient Personalization: \texttt{cwFedAvg} with \texttt{WDR} achieves personalization while maintaining the communication efficiency of \texttt{FedAvg} without requiring pairwise model collaborations or additional local training, making it particularly suitable for resource-constrained devices.
\item Enhanced Privacy: Our distribution estimation method with \texttt{WDR} enables privacy-preserving class distribution sharing, making this approach applicable beyond personalization to other FL contexts, such as client selection.
\end{itemize}
\section{Related Work}  \label{sec:related_work}
\textbf{Personalized Federated Learning.}  
Among PFL approaches, our work aligns with recent methods that personalize models through modified aggregation techniques. \texttt{FedFomo} \citep{zhang2020personalized} encourages FL among relevant clients by utilizing an optimally weighted combination of models. \texttt{FedAMP} \citep{huang2021personalized} uses attention-based techniques to promote stronger collaboration between clients with similar data distributions. However, these methods often require heavy computation for weight learning or additional communication to download other clients' models.
Clustering-based FL methods address this issue by performing FL within client clusters. \texttt{CFL} \citep{sattler2020clustered} employs hierarchical clustering using cosine similarity of clients' gradient updates as a post-processing step. \texttt{IFCA} \citep{ghosh2020efficient} assigns clients to pre-determined clusters and performs model aggregation within each cluster.
Regularization-based methods eliminate both extensive computation and clustering assumptions. \texttt{FedNH} \citep{dai2023tackling} adds normalization layers to ensure the uniformity and semantics of class prototypes. \texttt{FedUV} \citep{son2024feduv} introduces weight and representation regularization to emulate IID settings. Like these approaches, \texttt{cwFedAvg} leverages relevant clients for personalization with weight regularization. However, it avoids complex client collaboration and does not require assuming clients can be grouped into discrete clusters.

\noindent  \textbf{Correlating Model Parameters and Class Distribution.} 
In centralized machine learning, several studies have observed relationships between gradients or weights of deep networks and empirical class distribution ~\cite{anand1993improved, kang2020decoupling}. \citet{anand1993improved} revealed a correlation between the number of class samples and the magnitude of gradients associated with that class. They proposed an algorithm to accelerate learning by exploiting this correlation. 
\citet{kang2020decoupling} observed the positive proportional relationship between class and weight distribution.
In the realm of FL, utilizing class distribution is pivotal as it can be employed for client selection \cite{yang2021federated} and loss function modification on clients \citep{wang2021addressing,zhang2022federated}. However, owing to privacy concerns, directly transmitting class distribution information to the server is typically prohibited. Consequently, several proxy methods have been proposed to estimate class distribution from a deep network. 
\citet{yang2021federated} utilized the gradient magnitude as a proxy to estimate the class distributions of clients and employed it for client selection. 
\citet{wang2021addressing} developed a monitoring scheme that estimates class distribution based on the work of \cite{anand1993improved}.
Our distribution estimation is motivated by \citet{anand1993improved} but uses weights instead of gradients and incorporates \texttt{WDR} to strengthen the class-weight distribution correlation.
\section{Problem Formulation and Motivation}
This section presents the problem formulation and motivation based on theoretical and empirical analysis.
\subsection{Problem Formulation}  
The objective of traditional FL can be summarized as follows.
\begin{align}
    \label{eq:fl_objective}
    \min _{\boldsymbol{w}} f_G(\boldsymbol{w})=\min _{\boldsymbol{w}} \sum_{i=1}^M p_i F_i(\boldsymbol{w}),
\end{align}
where $f_G(\cdot)$ and $F_i(\cdot)$ denote the global objective and the local objective of client $i$, respectively. 
The global objective $f_G(\boldsymbol{w})$ is the weighted sum of $M$ local objectives, with $M$ being the number of clients. The weight $p_i$ for each client is defined as the ratio of the number of data samples $n_i$ on that client to the total number of data samples $n = \sum_{i=1}^M n_i$ across all clients, thus $p_i = \frac{n_i}{n}$. The local objective $F_i(\cdot)$ for each client $i$ can be defined as the expected loss over the data distribution $\mathcal{D}_i$ specific to that client. 
We approximate this expected loss using the empirical risk calculated over the local training data $\mathcal{D}_i^{tr}$ available to the client. 
This empirical risk minimization is expressed as $\mathbb{E}_{z \sim \mathcal{D}_i}[\mathcal{L}(\cdot ; z)]\approx \frac{1}{n_i} \sum_{z \in \mathcal{D}_i^{tr}}\mathcal{L}(\cdot ; z)$,
where $z$ represents the data under local distribution $\mathcal{D}_i$. 

In PFL, the global objective can take a more flexible form. The goal is to optimize a set of personalized models, one for each client, rather than a single global model. This objective can be expressed as follows.
\begin{align}
    \label{eq:pfl_objective}
    \min _{\boldsymbol{W}} f_P(\boldsymbol{W})
    &=\min _{\boldsymbol{w}_{\boldsymbol{i}}, i \in[M]}f_P\left(\boldsymbol{w}_1, \ldots, \boldsymbol{w}_M\right),
\end{align}
where $f_P(\boldsymbol{W})$ denotes the global objective for the PFL algorithm, and $\boldsymbol{w}_i$ denote personalized models. 
The goal is to find the optimal set of personalized models $\boldsymbol{W}^*$ that minimizes the global objective function $f_P(\boldsymbol{W})$.

\subsection{Motivation}
The extension from \texttt{FedAvg} to \texttt{cwFedAvg} is made possible by methods to quantify class-specific information encoded in deep networks. Our work is motivated by empirical and theoretical findings showing that deep networks encode class-specific information in their weights and gradients~\citep{anand1993improved,khakzar2021neural,kang2020decoupling}. A deep network develops pathways, where a pathway represents a union of paths (weights) from input to output~\citep{khakzar2021neural}. Critical pathways, consisting of large-magnitude weights, demonstrate distinct patterns across different classes~\citep{wang2018interpret,yu2018distilling,qiu2019adversarial}.

\noindent \textbf{A Motivational Example.}
We first analyze how pathways diverge across clients due to non-IID data distributions through an example. 
We construct a synthetic dataset with two centers in $\mathbb{R}^3$ space generated from a Gaussian distribution for a binary classification task.
The dataset contains 6000 samples: 3400 for class $k=0$ and 2600 for class $k=1$.
The data are partitioned across three clients with distinct distributions. Client 1 exhibits a high imbalance, with 2700 samples of class $k=0$ and 300 samples of class $k=1$. Client 2 presents an inverse distribution, with 200 samples of class $k=0$ and 1800 samples of class $k=1$. Client 3 maintains a balanced distribution, with 500 samples per class. We employ a single-layer neural network with four neurons and ReLU activation functions.

\begin{figure}[bt]
    \begin{subfigure}[b]{0.085\textwidth}
        \includegraphics[width=\textwidth]{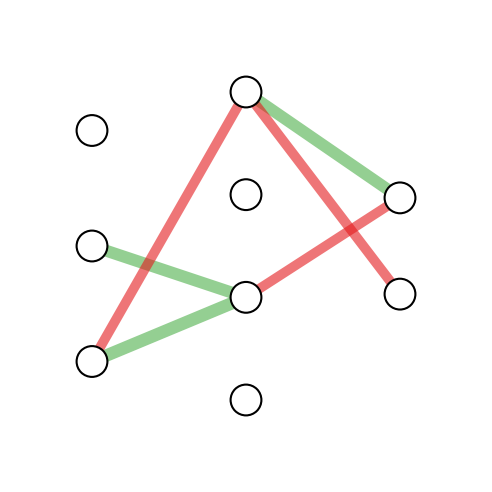}
        \caption{$\boldsymbol{w}_{1}^{L}$}\label{fig:w_1l}
    \end{subfigure} 
    \!
    \begin{subfigure}[b]{0.085\textwidth}
        \includegraphics[width=\textwidth]{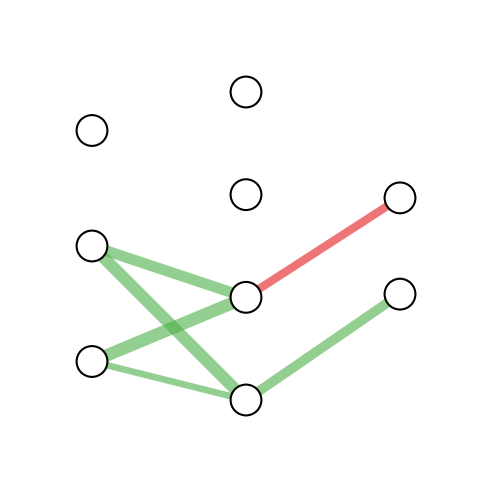}
        \caption{$\boldsymbol{w}_{2}^{L}$}\label{fig:w_2l}
    \end{subfigure} 
    \!
    \begin{subfigure}[b]{0.085\textwidth}
        \includegraphics[width=\textwidth]{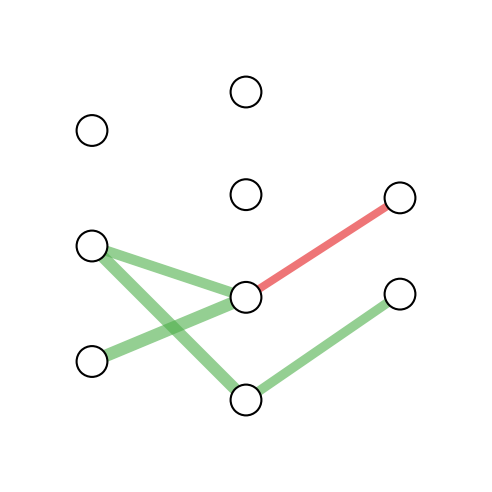}
        \caption{$\boldsymbol{w}_{3}^{L}$}\label{fig:w_3l}
    \end{subfigure} 
    \!
    \begin{subfigure}[b]{0.085\textwidth}
        \includegraphics[width=\textwidth]{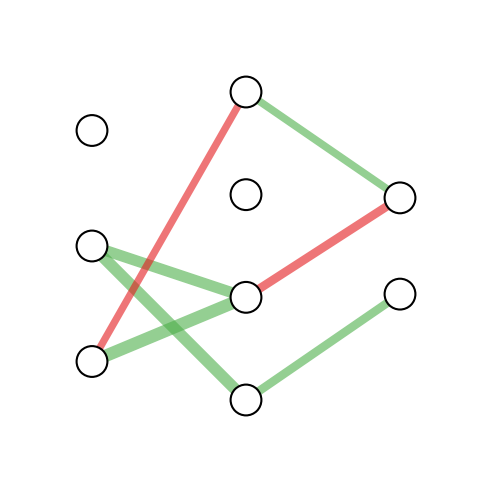}
        \caption{$\boldsymbol{w}^{G}$}\label{fig:w_g}
    \end{subfigure} 
    \\
    \begin{subfigure}[b]{0.085\textwidth}
        \includegraphics[width=\textwidth]{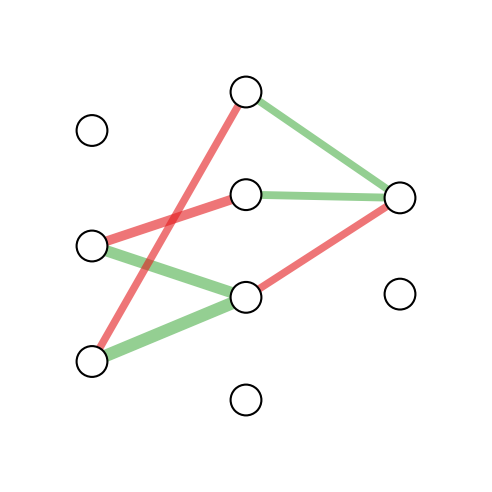}
        \caption{$\boldsymbol{\tilde{w}}_{1}^{L}$}\label{fig:w_1l_wdr}
    \end{subfigure} 
    \!
    \begin{subfigure}[b]{0.085\textwidth}
        \includegraphics[width=\textwidth]{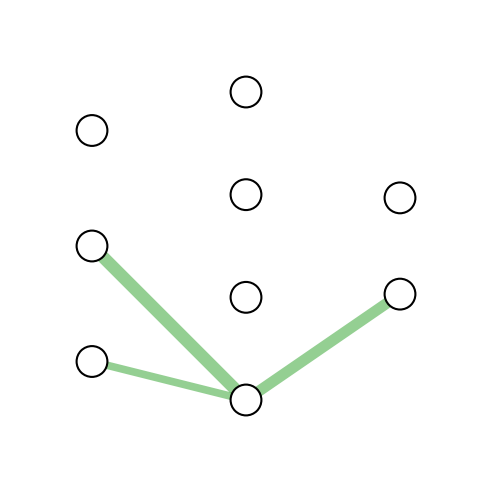}
        \caption{$\boldsymbol{\tilde{w}}_{2}^{L}$}\label{fig:w_2l_wdr}
    \end{subfigure} 
    \!
    \begin{subfigure}[b]{0.085\textwidth}
        \includegraphics[width=\textwidth]{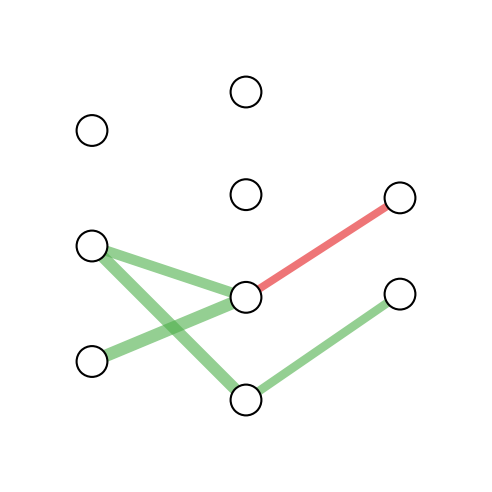}
        \caption{$\boldsymbol{\tilde{w}}_{3}^{L}$}\label{fig:w_3l_wdr}
    \end{subfigure} 
    \!
    \begin{subfigure}[b]{0.085\textwidth}
        \includegraphics[width=\textwidth]{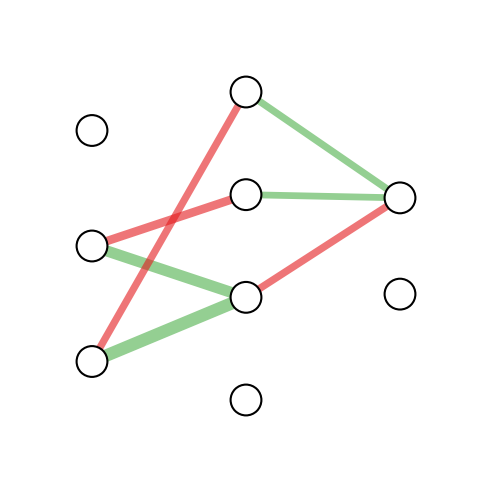}
        \caption{$\boldsymbol{\tilde{w}}_{1}^{G}$}\label{fig:w_1g_wdr}
    \end{subfigure} 
    \!
    \begin{subfigure}[b]{0.085\textwidth}
        \includegraphics[width=\textwidth]{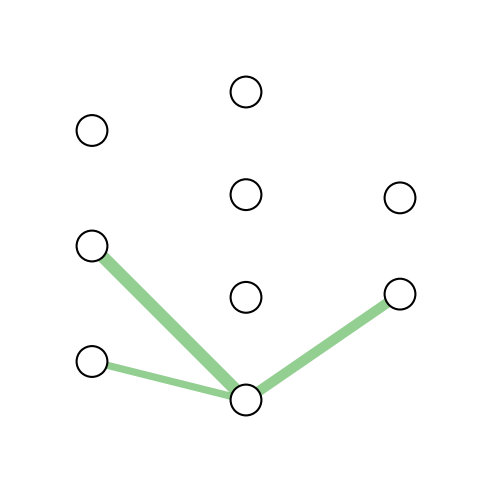}
        \caption{$\boldsymbol{\tilde{w}}_{2}^{G}$}\label{fig:w_2g_wdr}
    \end{subfigure} 
    \\
    \begin{subfigure}[b]{0.085\textwidth}
        \includegraphics[width=\textwidth]{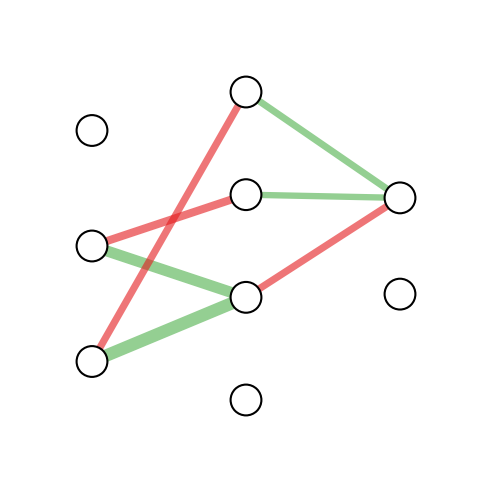}
        \caption{$\boldsymbol{\tilde{w}}_{1}^{L^*}$}\label{fig:w_1l_wdr_a}
    \end{subfigure} 
    \!
    \begin{subfigure}[b]{0.085\textwidth}
        \includegraphics[width=\textwidth]{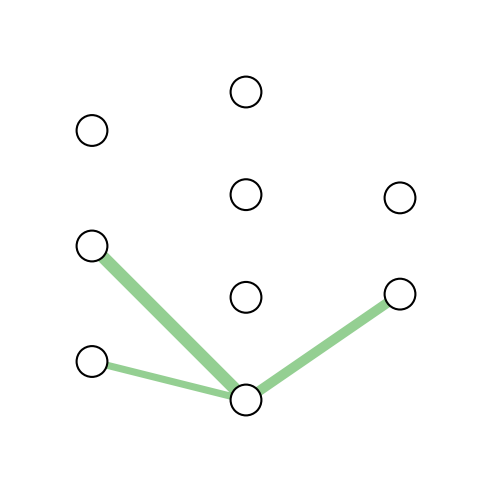}
        \caption{$\boldsymbol{\tilde{w}}_{2}^{L^*}$}\label{fig:w_2l_wdr_a}
    \end{subfigure} 
    \!
    \begin{subfigure}[b]{0.085\textwidth}
        \includegraphics[width=\textwidth]{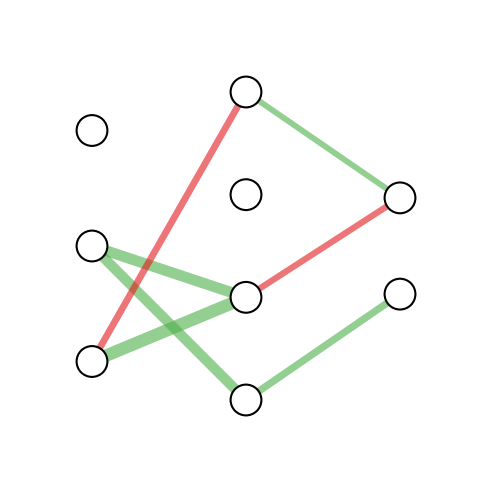}
        \caption{$\boldsymbol{\tilde{w}}_{3}^{L^*}$}\label{fig:w_3l_wdr_a}
    \end{subfigure} 
    \caption{Pathway comparison of \texttt{FedAvg} and \texttt{cwFedAvg}. $\boldsymbol{\tilde{w}}$ denotes the model trained with \texttt{WDR}, and $*$ denotes the local models updated using class-specific global models (h) and (i).}
    \label{fig:pathway}
\end{figure}

We visualize the network architecture, where green and red lines represent positive and negative weights, respectively, and line width indicates magnitude. 
We draw critical pathways created from the top 8 out of 20 weights and denote ${{\boldsymbol{w}}_{i}^{L}}$ as the $i$-th local model and ${{\boldsymbol{w}}^{G}}$ as a global model of \texttt{FedAvg}. Figures~\ref{fig:w_1l}--\ref{fig:w_3l} show the learned pathways of local models for clients 1, 2, and 3, respectively. 
Notably, we observe that both output neurons contribute critical pathways in the three figures, and pathways in Figures~\ref{fig:w_2l} and \ref{fig:w_3l} remain similar regardless of data imbalances. Figure~\ref{fig:w_g} shows the pathways of the global model after the aggregation of \texttt{FedAvg}. This demonstrates why \texttt{FedAvg} fails to personalize, as the single averaged global model cannot capture the unique patterns of each client.

\noindent \textbf{Theoretical Background and Observations.} 
Although the magnitude of the weights in critical pathways varies with class proportions, its relationship with class distribution remains partially understood.
The analysis of \citet{anand1993improved} revealed a correlation between gradients of weights in the output layer and class sample sizes.
Let $\mathbf{w}_{i,j}$ denote the weight vector from penultimate layer neurons to output neuron $j$ of client $i$. \citet{anand1993improved} formalized the correlation through a theorem: for a neural network classifier, the squared $\ell_2$-norms of the gradients of the weight vectors satisfy:
  \begin{align}
    \frac{\mathbb{E}\left\|\nabla \mathcal{L}\left(\mathbf{w}_{i,j}\right)\right\|_2^2}{\mathbb{E}\left\|\nabla \mathcal{L}\left(\mathbf{w}_{i,k}\right)\right\|_2^2} \approx \frac{n_{i,j}^2}{n_{i,k}^2},
    \label{eq:correlation_gradient}
  \end{align}
where $n_{i,j}$ denotes the number of samples belonging to class $j$ on client $i$. 

\begin{figure}[t]
    \centering
    \begin{subfigure}[b]{0.29\columnwidth}
        \includegraphics[width=\columnwidth]{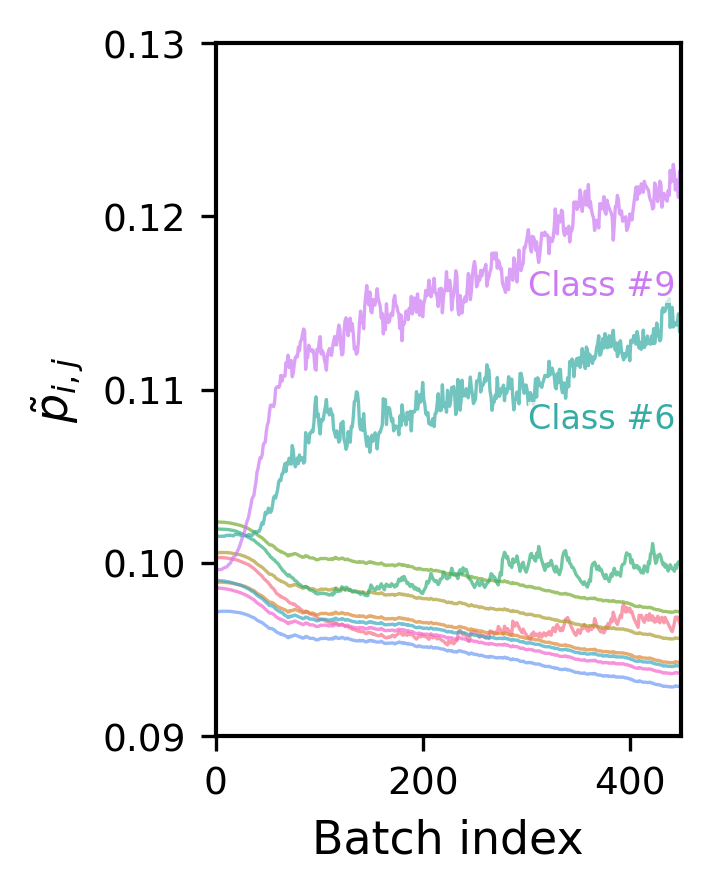} 
        \caption{Train w/o \texttt{WDR}} \label{fig:line_p}
    \end{subfigure} 
    \begin{subfigure}[b]{0.29\columnwidth}
        \includegraphics[width=\columnwidth]{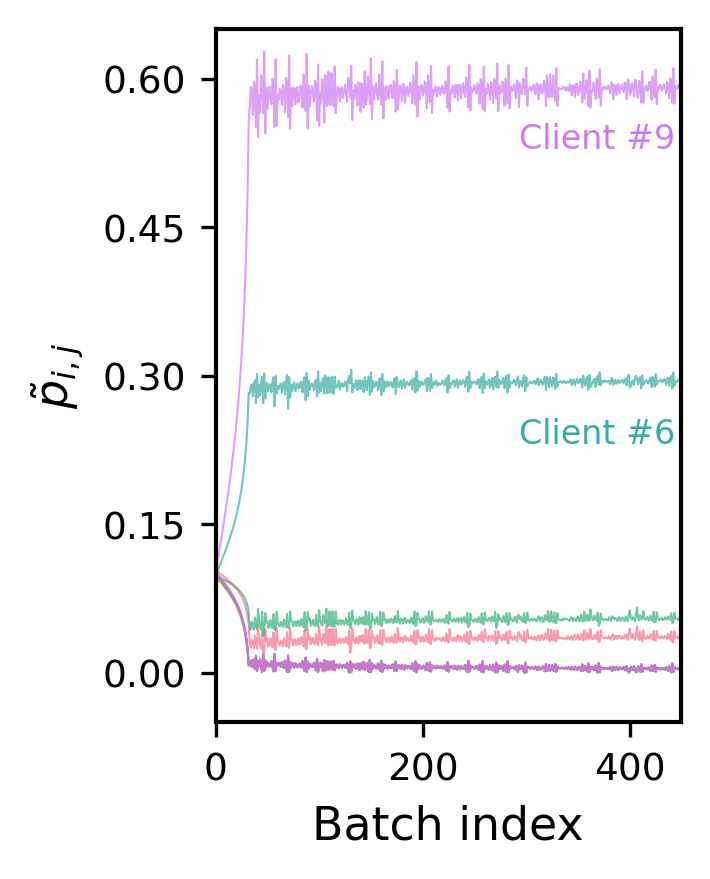} 
        \caption{Train w/ \texttt{WDR}} \label{fig:line_p_wdr}
    \end{subfigure} 
    \begin{subfigure}[b]{0.385\columnwidth}
        \includegraphics[width=\columnwidth]{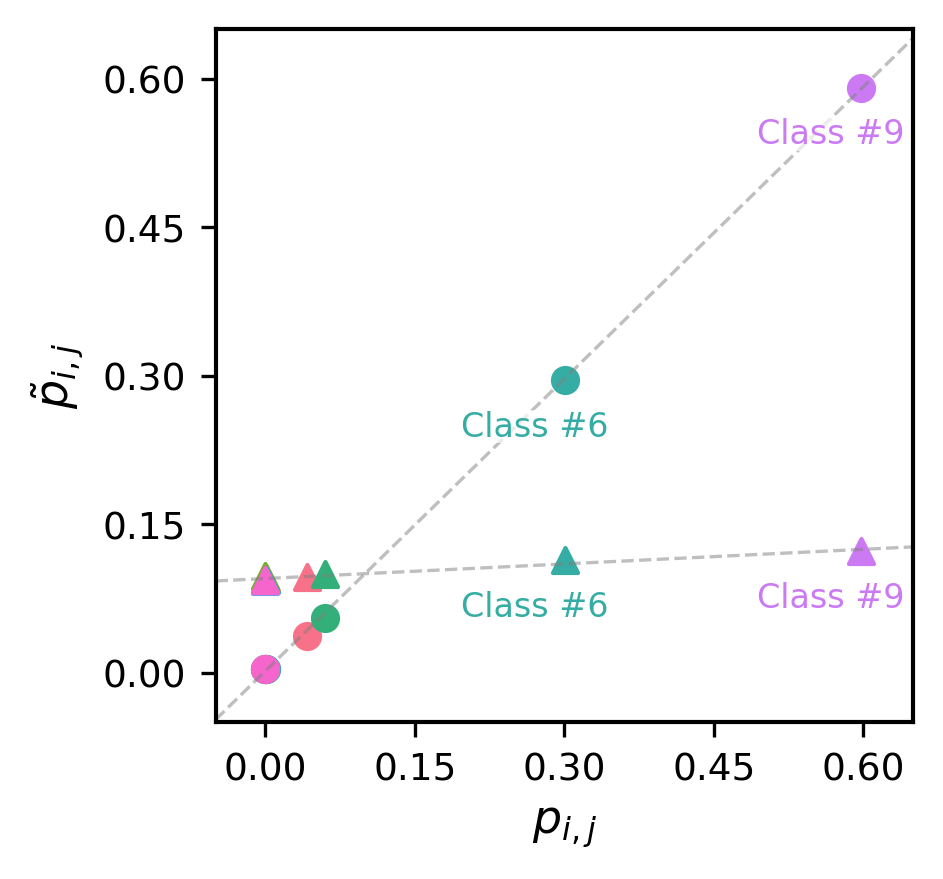} 
        \caption{Correlation} \label{fig:scatter_p}
    \end{subfigure}
    \caption{Evolution of $\tilde{p}_{i,j}$ and its correlation with ${p}_{i,j}$.}
    \label{fig:norm_correlation}
    \vspace{-10pt}    
\end{figure}

Based on the theorem, we found a positive correlation between $\left\|\mathbf{w}_{i,j}\right\|_2$ and $n_{i,j}$. We further extend this relationship to demonstrate that the distribution derived from normalized $\ell_2$-norms of the weight vectors $\tilde{p}_{i,j}$ correlates with the empirical class distribution $p_{i,j} = \frac{n_{i,j}}{n_i}$. In this work, for $K$-class classification, $\tilde{p}_{i,j}$ is defined as:
\begin{align}
\tilde{p}_{i,j} = \frac{\left\|\mathbf{w}_{i,j}\right\|_2}{\sum_{k=1}^K \left\|\mathbf{w}_{i,k}\right\|_2}.
\label{eq:correlation_weight}
\end{align}
Indeed, we found that a correlation exists between them. We observed the evolution of $\tilde{p}_{i,j}$ for a client during the first FL round's epoch on CIFAR-10, where each color represents a different class in Figure \ref{fig:line_p}. Classes 6 and 9 with larger samples exhibit higher $\tilde{p}_{i,j}$ values than the others. The triangular markers in Figure \ref{fig:scatter_p} indicate a positive correlation. 

However, proving Eq. (\ref{eq:correlation_gradient}) and (\ref{eq:correlation_weight}) for all network weights remains challenging.
Unlike weights in the output layer that directly correspond to specific classes, weights in other layers lack this explicit class association. Nevertheless, we argue that these weights correlate with class proportions due to the cascading effect of backpropagation from the output layer. Based on this, we hypothesize that the contribution of each class to the trained model can be quantified as the
product of weights and the class proportion $p_{i,j}\boldsymbol{w}_i$.
This is analogous to the quantification of the contribution of each client to the global model in \texttt{FedAvg}, quantified as $p_i\boldsymbol{w}_i$. 

\section{The \texttt{cwFedAvg} Algorithm} \label{sec:cwFedAVG}
In this section, we describe \texttt{cwFedAvg} with \texttt{WDR} and conclude by analyzing its relationship to \texttt{FedAvg}.

\subsection{Class-Wise Aggregation}
\noindent \textbf{Class-Wise Local Model Aggregation.} 
We denote ${\boldsymbol{w}}_{i}^{L}$ as the $i$-th local model and ${\boldsymbol{w}}_{j}^{G}$ as the $j$-th class-specific global model for $M$ clients in a $K$-class classification task. The server aggregates local models by creating class-specific global models as follows (Figure \ref{fig:local_aggregation} in the supplementary materials). 
\begin{align}
\boldsymbol{w}_{j}^{G}
&= \sum_{i=1}^M q_{i,j} \boldsymbol{w}_{i}^{L}
\label{eq:q_ij}
\\
&= \sum_{i=1}^M \dfrac{ p_i \cdot p_{i,j}  }{\sum_{i=1}^M p_i \cdot p_{i,j}} \boldsymbol{w}_{i}^{L},  \label{eq:q_ij_from_p}
\end{align}
where ${q}_{i,j}$ is a weighting factor.
After expressing ${q}_{i,j}$ in terms of sample counts in Eq. (\ref{eq:q_ij_from_p}), we get ${q}_{i,j} = \frac{n_{i,j}}{\sum_{i=1}^M n_{i,j}}$, which represents client $i$'s proportion of total class $j$ samples. Therefore, we consider this term as client $i$'s contribution to class $j$ in the system, and this aggregation becomes analogous to \texttt{FedAvg} performed separately for each class.

\noindent \textbf{Class-Wise Global Model Aggregation.} 
In contrast to \texttt{FedAvg}, where a single global model is copied to each local model, 
\texttt{cwFedAvg} performs a weighted summation of $K$ class-specific global models as follows (Figure \ref{fig:global_aggregation} in the supplementary materials). 
\begin{align}
    \label{eq:w_L_i}
    \boldsymbol{w}_{i}^{L}=\sum_{j=1}^K p_{i,j}\boldsymbol{w}_{j}^{G}.
\end{align}
This weighted aggregation based on the class distributions of each client leads to personalization.

\subsection{Weight Regularization for \texttt{cwFedAvg}}\label{sec:wdr}
In this section, we argue that \texttt{cwFedAvg} requires two key conditions for effectiveness and privacy protection: (1) model weights that strongly correlate with the empirical class distribution to enable effective personalization, and (2) secure sharing of privacy-sensitive class distribution ${p}_{i,j}$ without compromising client information ${n}_{i,j}$. We present a method that simultaneously addresses both.

Regarding the first condition, we observed that $\tilde{\boldsymbol{p}}_i=[\tilde{p}_{i,1}, \tilde{p}_{i,2}, ..., \tilde{p}_{i,K}]$ correlates with the empirical distribution $\boldsymbol{p}_i=[p_{i,1}, p_{i,2}, ..., p_{i,K}]$, but they differ significantly as shown in Figure~\ref{fig:scatter_p} (triangular markers). Notably, even when class $j$ has zero or near-zero samples, $\tilde{p}_{i,j}$ maintains a non-negligible value approximately 0.1 (the leftmost triangles).
To strengthen the correlation between $\boldsymbol{p}_i$ and $\tilde{\boldsymbol{p}}_i$, we propose \texttt{WDR}, which minimizes the Euclidean distance between them:
\vspace{-0.5em}
\begin{align}
\label{eq:wdr}
\mathcal{R}_i=\left\|\boldsymbol{p}_i - \tilde{\boldsymbol{p}}_i\right\|_2.
\end{align}
With the regularization term, the total cost function $\tilde{\mathcal{L}}_i$ is denoted as 
$\tilde{\mathcal{L}}_i=\mathcal{L}_i+\lambda \mathcal{R}_i$,
where $\lambda \in [0, \infty)$ is the regularization coefficient. 
When training with \texttt{WDR} (Figure \ref{fig:line_p_wdr}) $\tilde{p}_{i,j}$ evolves distinctly from the patterns in Figure \ref{fig:line_p}, with more apparent separation between classes. This improved separation is further illustrated in Figure \ref{fig:scatter_p} (circular markers), where \texttt{WDR} enables $\tilde{p}_{i,j}$ to accurately approximate ${p}_{i,j}$.
This precise approximation enables replacing ${p}_{i,j}$ with $\tilde{p}_{i,j}$ as a reliable substitute, addressing our second condition. Clients, therefore, do not need to directly share $n_{i,j}$.

The pathway analysis provides additional validation of applying \texttt{cwFedAvg} with \texttt{WDR}. Figures~\ref{fig:w_1l_wdr}--\ref{fig:w_3l_wdr} show how \texttt{WDR} aligns weight patterns with the dominant class of each client, unlike Figures~\ref{fig:w_1l}--\ref{fig:w_3l}.
For example, Figure~\ref{fig:w_1l_wdr} shows pathways connecting to the first output neuron with most samples from class 0, whereas Figure~\ref{fig:w_1l} shows pathways connecting to both output neurons.
Through the class-wise aggregations, class-specific global models (Figures~\ref{fig:w_1g_wdr} and \ref{fig:w_2g_wdr}) mirror the pathways of clients with the majority of samples for each respective class (Figures~\ref{fig:w_1l_wdr} and \ref{fig:w_2l_wdr}). The updated client models (Figures~\ref{fig:w_1l_wdr_a}--\ref{fig:w_3l_wdr_a}) maintain their original pathway structures, showing minimal deviation from their pre-update models, thus indicating personalization.

\begin{algorithm}[tbh]
\caption{\texttt{cwFedAvg} with \texttt{WDR}}
\hspace*{0.02in} {\bf Input:} $M$ clients, regularization coefficient $\lambda$ \\
\hspace*{0.02in} {\bf Server executes:}
\begin{algorithmic}[1]
\State Initialize local model $\boldsymbol{w}_{i}^{L}$ and global model $\boldsymbol{w}_{j}^{G}$ 
\State Initialize $\tilde{p}_{i,j}$ to $\frac{1}{K}$ 
\For{iteration $t = 1, \ldots, T$}
    \State Sample a client subset $\mathcal{C}^t$
    \For{client $i \in \mathcal{C}^t$ in parallel}
        \State Compute $\boldsymbol{w}_{i}^{L}$ via Eq. (\ref{eq:w_L_i}) replacing ${p}_{i,j}$ with $\tilde {p}_{i,j}$
        \State $\boldsymbol{w}_{i}^{L} \leftarrow$ ClientUpdate$\left(i,\boldsymbol{w}_{i}^{L}\right)$
    \EndFor
    \State Compute $\tilde {p}_{i,j}$ via Eq. (\ref{eq:correlation_weight}) 
    \State Compute $\boldsymbol{w}_{j}^{G}$ via Eq. (\ref{eq:q_ij_from_p}) replacing ${p}_{i,j}$ with $\tilde {p}_{i,j}$
\EndFor
\end{algorithmic}
\hspace*{0.01in} \\
\hspace*{0.01in} {\bf ClientUpdate$\left(i, \boldsymbol{w}_{i}^{L}\right)$:}
\begin{algorithmic}[1]
\For{each local epoch}
    \For{batch $b \in \mathcal{B}_i$}
        \State Update $\boldsymbol{w}_{i}^{L}$ according to the loss via Eq. (\ref{eq:wdr})
    \EndFor
\EndFor
\State \Return $\boldsymbol{w}_{i}^{L}$ to the server
\end{algorithmic}
\label{alg:cwFedAvg}
\end{algorithm}

Algorithm \ref{alg:cwFedAvg} presents the complete FL process of \texttt{cwFedAvg} with \texttt{WDR}. Building upon the \texttt{FedAvg} framework, the proposed method introduces two key features: class-wise aggregation (lines 6 and 9 on server side) and class distribution estimation using \texttt{WDR} (line 8 on server side and line 3 on client side).

\subsection{Comparative Analysis with \texttt{FedAvg}} \label{sec:cwfedavg_comparative}
\noindent {\bf Insight on Relation between Global Models.} 
We analyze class-specific global models of \texttt{cwFedAvg} compared with the global model of \texttt{FedAvg} under two scenarios. (1) In the IID case, one can prove that $w_{j}^G = w^G$, where $w^G$ is the global model of \texttt{FedAvg} if all clients have the complete set of classes in their local datasets and maintain uniform class distribution. This implies that \texttt{cwFedAvg} performs similarly to \texttt{FedAvg} as data heterogeneity decreases. (2) In the extreme non-IID case, when each client $i$ has data from exactly one class $k_i$, we demonstrate that
\begin{align}
    w_{j}^G = \sum_{i:k_i=j} \frac{n_i}{\sum_{l:k_l=j} n_l} w_i^L.
\end{align}
This equation indicates that the $j$-th class-specific global model is equivalent to the global model created by \texttt{FedAvg} of local models containing only class $j$. Both cases can be trivially proven through straightforward mathematical derivations. Although most practical scenarios fall between these two cases, theoretically proving how the differences between global models impact performance remains challenging.

\begin{figure}[th]
    \centering
    \begin{subfigure}[b]{0.15\textwidth}
        \includegraphics[width=\textwidth]{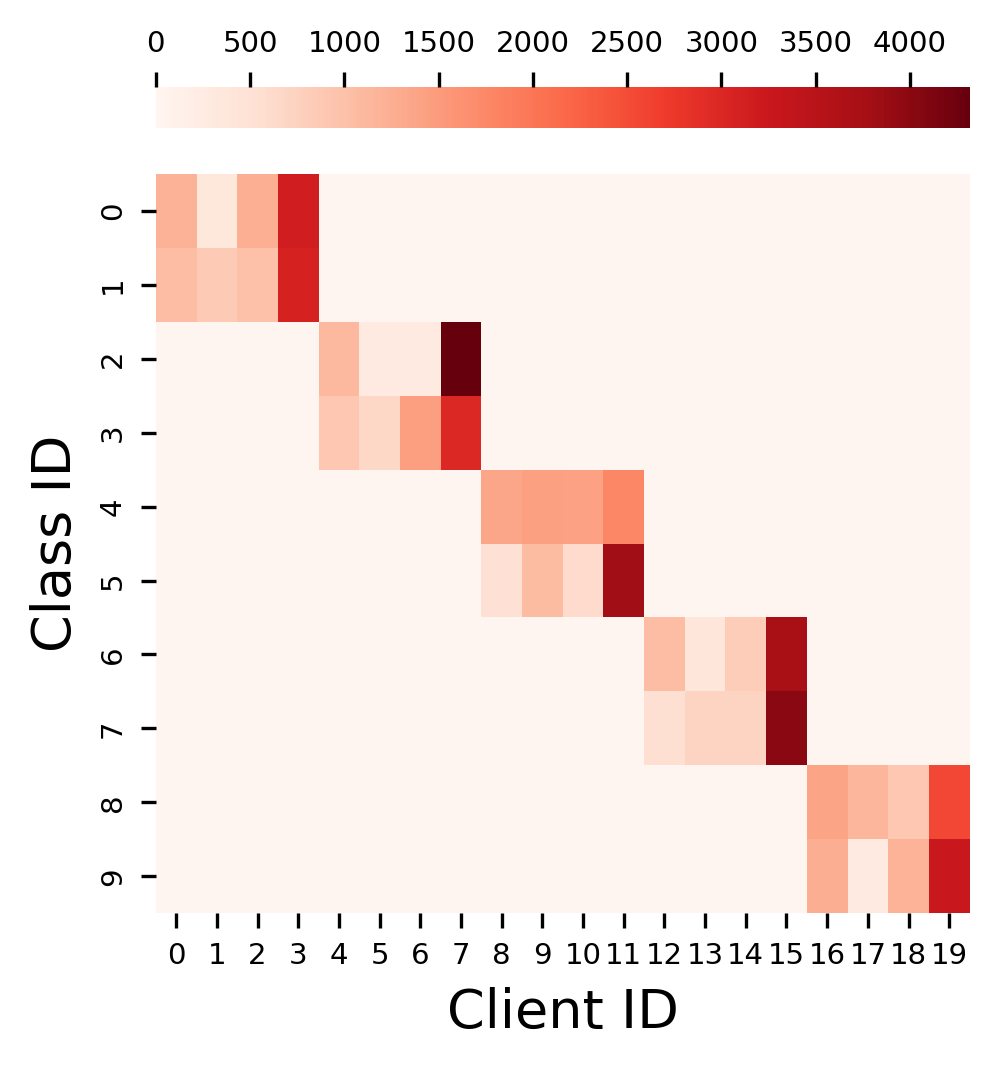}
        \caption{Data distribution of clients }\label{fig:heatmap_cifar10_pat_data}
    \end{subfigure} 
    \hfill
    \begin{subfigure}[b]{0.15\textwidth}
        \includegraphics[width=\textwidth]{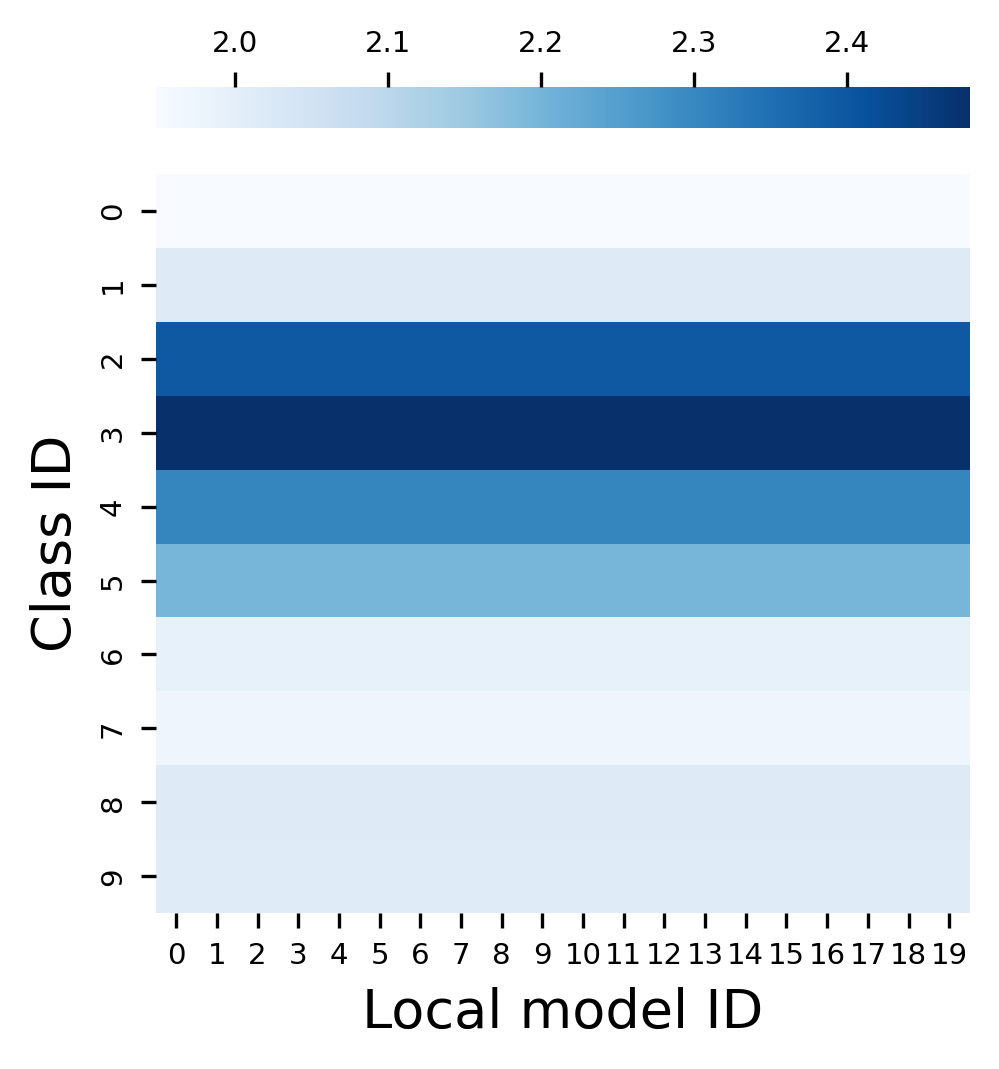}
        \caption{Local models of \texttt{FedAvg} }\label{fig:heatmap_cifar10_pat_fedavg_local}
    \end{subfigure}
    \hfill
    \begin{subfigure}[b]{0.15\textwidth}
        \includegraphics[width=\textwidth]{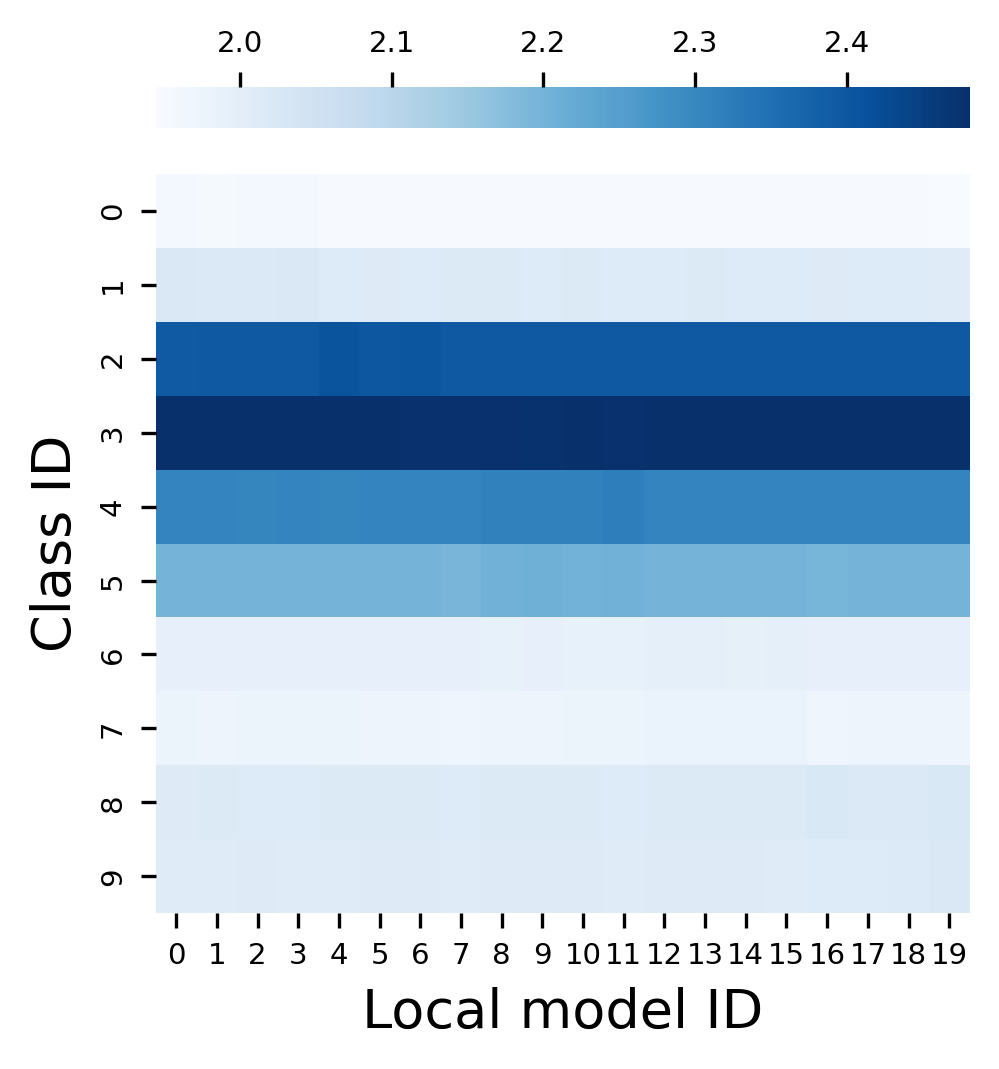}
        \caption{Local models of \\ fine-tuned \texttt{FedAvg}}\label{fig:heatmap_cifar10_pat_fedavg_ft_local}
    \end{subfigure} 
    \\
    \begin{subfigure}[b]{0.15\textwidth}
        \includegraphics[width=\textwidth]{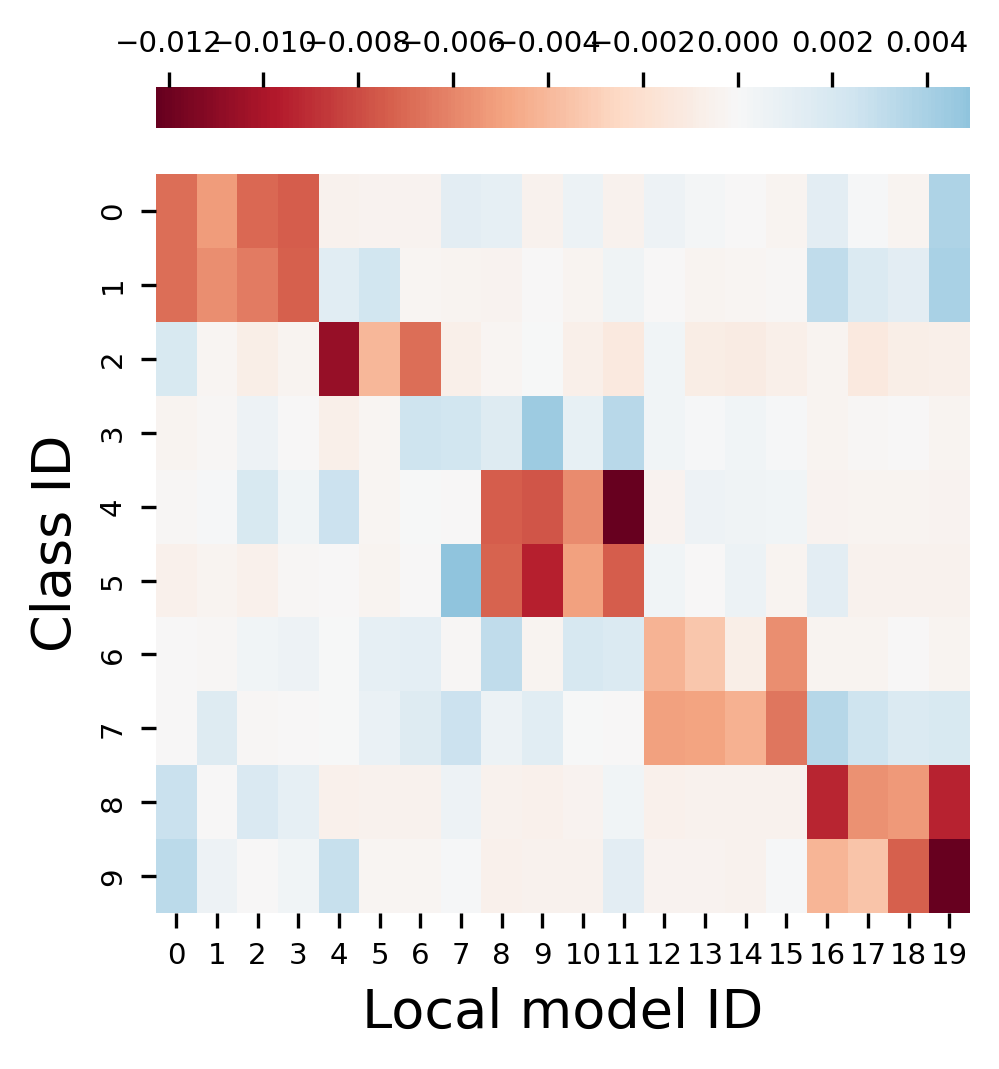}
        \caption{Difference between \\ (b) and (c) }\label{fig:heatmap_cifar10_pat_diff}
    \end{subfigure}
    \hfill
    \begin{subfigure}[b]{0.15\textwidth}
        \includegraphics[width=\textwidth]{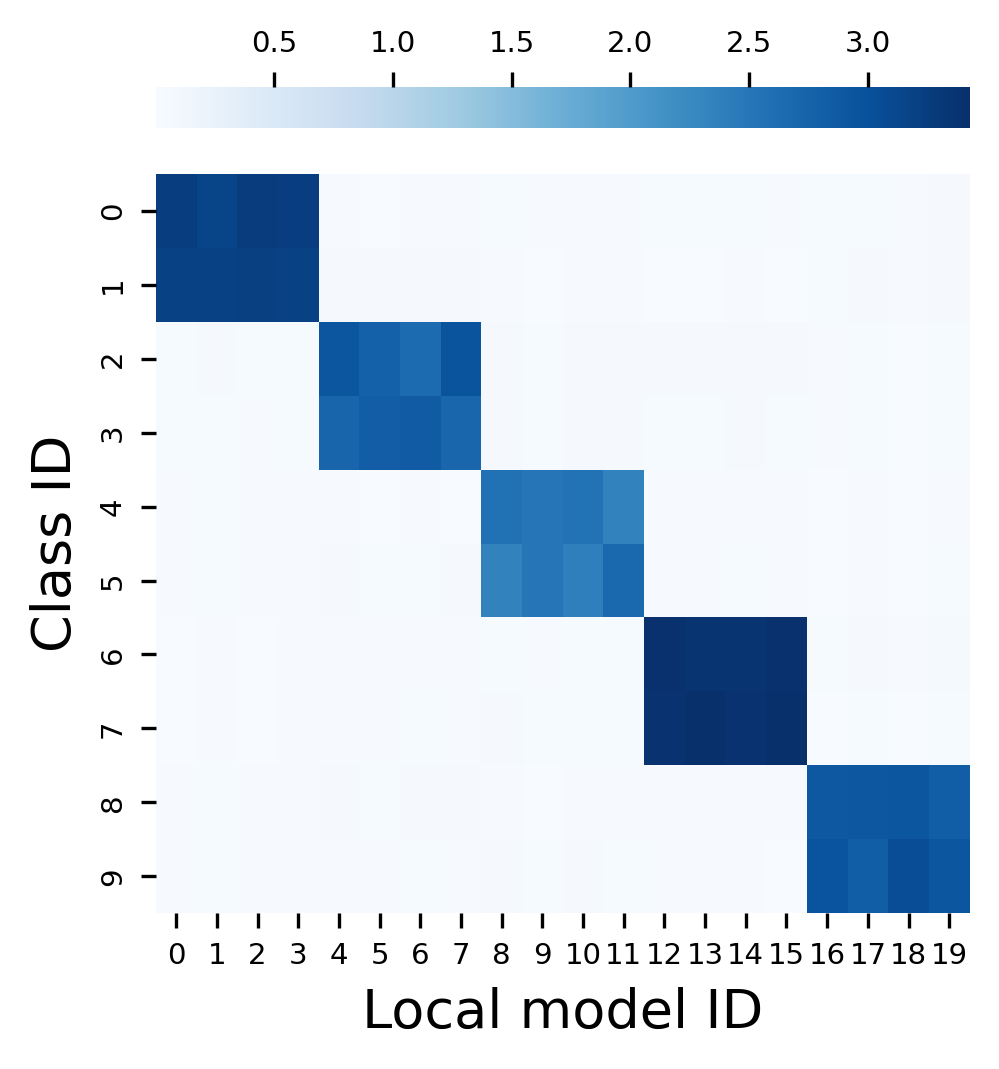}
        \caption{Local models of \\ \texttt{cwFedAvg} w/ \texttt{WDR}}\label{fig:heatmap_cifar10_pat_cwfedavg_wdr_local}
    \end{subfigure}
    \hfill
    \begin{subfigure}[b]{0.15\textwidth}
        \includegraphics[width=\textwidth]{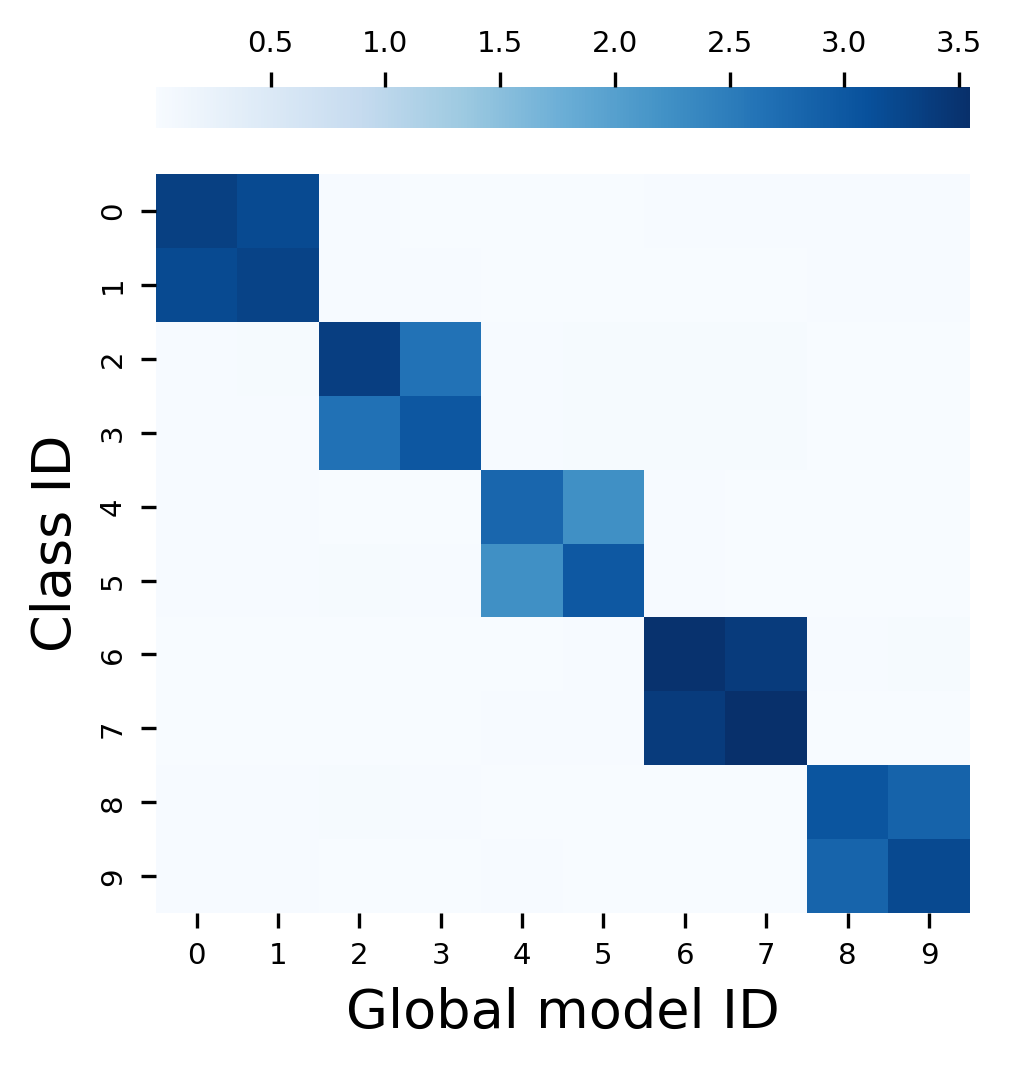}
        \caption{Global models of \\ w/ \texttt{WDR}}\label{fig:heatmap_cifar10_pat_cwfedavg_wdr_global}
    \end{subfigure}   
    \caption{Heatmaps depicting the data distribution and $\ell_2$-norms of output layer weight vectors for the CIFAR-10 pathological setting. (a) Each cell represents the number of data samples belonging to class $j$ for client $i$. (b)--(f) Each cell shows the $\ell_2$-norm of output layer weight vector, $\left\|\mathbf{w}_{i,j}\right\|_2$.}
    \label{fig:heatmap_cifar10_pat}
    \vspace{-10pt}
\end{figure}

\begin{table*}[th]
\centering
\setlength{\tabcolsep}{6.5pt}
{\fontsize{9}{10}\selectfont
\begin{tabular}{@{}lrrrrrrr@{}}
\toprule
\multirow{2}{*}[-2pt]{Algorithm} & \multicolumn{2}{c}{Pathological setting} & \multicolumn{5}{c}{Practical setting ($\alpha=0.1$)} \\
\cmidrule(l){2-3} \cmidrule(l){4-8}
& CIFAR-10 & CIFAR-100 & MNIST & CIFAR-10 & CIFAR-100 & Tiny ImageNet & Tiny ImageNet\textsuperscript{*} \\
\midrule
\texttt{FedAvg}  & 60.68$\pm$0.84 & 28.22$\pm$0.32 & 98.70$\pm$0.04 & 61.94$\pm$0.56 & 32.44$\pm$0.42 & 21.35$\pm$0.12 & 24.71$\pm$0.15 \\
\texttt{FedProx} & 60.65$\pm$0.92 & 28.59$\pm$0.28 & 98.68$\pm$0.09 & 62.48$\pm$0.86 & 32.26$\pm$0.26 & 20.65$\pm$0.12 & 24.06$\pm$0.16 \\
\midrule
\texttt{FedAMP}  & 88.82$\pm$0.15 & 63.29$\pm$0.49 & 99.26$\pm$0.01 & \textbf{89.46$\pm$0.11} & 47.65$\pm$0.62 & 29.95$\pm$0.10 & 31.38$\pm$0.18\\
\texttt{FedFomo} & 90.76$\pm$0.59 & 63.12$\pm$0.59 & 99.13$\pm$0.04 & 88.05$\pm$0.08 & 44.62$\pm$0.37 & 26.22$\pm$0.25 & 26.12$\pm$0.31\\
\texttt{CFL}  & 60.58$\pm$0.15 & 28.55$\pm$0.30 & 98.70$\pm$0.01 & 61.40$\pm$0.51 & 44.19$\pm$0.69 & 29.62$\pm$0.43 & 33.47$\pm$0.68\\
\texttt{IFCA} & 72.84$\pm$4.80 & 58.98$\pm$2.38 & 99.10$\pm$0.06 & 70.12$\pm$0.13 & 34.86$\pm$1.02 & 19.93$\pm$0.59 & 26.68$\pm$0.16\\
\texttt{FedNH}  & 50.82$\pm$0.33 & 26.26$\pm$0.36 & 98.85$\pm$0.29 & 56.38$\pm$0.17 & 32.98$\pm$0.88 & 17.04$\pm$0.07 & 24.24$\pm$0.76\\
\texttt{FedUV}  & 88.11$\pm$0.13 & 62.72$\pm$0.28 & 99.25$\pm$0.09 & 88.59$\pm$0.09 & 46.80$\pm$0.20 & 28.09$\pm$0.06 & 25.45$\pm$0.03\\
\midrule
\texttt{cwFedAvg} (Output)  & \textbf{91.23$\pm$0.04} & \textbf{67.50$\pm$0.14} & \textbf{99.52$\pm$0.03} & 88.65$\pm$0.19 & \textbf{56.29$\pm$0.18} & \textbf{41.38$\pm$0.12} & \textbf{43.51$\pm$0.14}\\
\bottomrule
\end{tabular}
}
\caption{Classification accuracy (\%) across datasets. Tiny ImageNet\textsuperscript{*} indicates experiments using ResNet-18. \texttt{cwFedAvg} (Output) denotes \texttt{cwFedAvg} selectively applied to the output layer.}
\label{table:total_result}
\vspace{-10pt}
\end{table*}

\noindent {\bf Personalization by \texttt{cwFedAvg}.} 
We visualize heatmaps of both the empirical data distribution and the $\ell_2$-norms of output layer weight vectors ($\left\|\mathbf{w}_{i,j}\right\|_2$) to demonstrate how \texttt{cwFedAvg} achieves effective personalization.
Figure \ref{fig:heatmap_cifar10_pat_data} presents the class-wise sample distribution across 20 clients for the CIFAR-10 pathological setting, where each client only has data from two classes.
Next, Figures \ref{fig:heatmap_cifar10_pat_fedavg_local} (\texttt{FedAvg}) and \ref{fig:heatmap_cifar10_pat_fedavg_ft_local} (fine-tuned \texttt{FedAvg} with local dataset) do not show any distinctions between local models in contrast to Figure \ref{fig:heatmap_cifar10_pat_data} and are visually similar. However, a closer inspection reveals subtle differences between the two at positions corresponding to the diagonal darker areas in Figure \ref{fig:heatmap_cifar10_pat_data}. 
Figure \ref{fig:heatmap_cifar10_pat_diff} visualizes the difference between the two heatmaps to highlight this distinction, where the diagonal areas appear darker, indicating that personalized models with local data (fine-tuned \texttt{FedAvg}) can induce changes in the norms. 
In contrast, Figure \ref{fig:heatmap_cifar10_pat_cwfedavg_wdr_local} (\texttt{cwFedAvg} with \texttt{WDR}) exhibits a pattern similar to Figure \ref{fig:heatmap_cifar10_pat_data}, suggesting that each model has undergone personalization tailored to its possessed classes. As designed, each global model in Figure \ref{fig:heatmap_cifar10_pat_cwfedavg_wdr_global} specializes in specific classes, as each client possesses data from only two classes in this setting. 
Consequently, the norm can be a quantitative measure for assessing model personalization.
Additional visualizations for more practical settings are provided in the supplementary materials.

\noindent {\bf Communication and Resource Requirements.} 
\texttt{cwFedAvg} maintains identical communication overhead to \texttt{FedAvg}, as all class-wise aggregations are performed on the server-side. This can be verified by comparing the dotted arrows in Figures~\ref{fig:fedavg_upload} and~\ref{fig:local_aggregation} for local model aggregation, and Figures~\ref{fig:fedavg_download} and~\ref{fig:global_aggregation} for global model aggregation in the supplementary materials. 
The client-side operations in \texttt{cwFedAvg} preserve the same memory and computational requirements as \texttt{FedAvg}. When coupled with \texttt{WDR}, the only additional cost is the negligible computation for regularization.
On the server side, although \texttt{cwFedAvg} requires more extensive storage to maintain multiple global models, its runtime memory allocation during aggregation matches \texttt{FedAvg} (Figure~\ref{fig:fedavg_upload} and~\ref{fig:local_aggregation}). Memory efficiency can be achieved by applying \texttt{cwFedAvg} selectively to upper layers, and computational efficiency can be achieved through parallel execution of class-wise model aggregations.

\noindent \textbf{Selective Application of \texttt{cwFedAvg} to Upper Layers.} 
Deep networks exhibit layer-wise characteristics, where upper layers tend to learn class-specific features while lower layers capture more general, class-agnostic features. Based on this property, we examine a selective application of \texttt{cwFedAvg}. This approach applies \texttt{cwFedAvg} exclusively to the upper layers while maintaining \texttt{FedAvg} for the lower layers. We analyze the impact of this selective application on performance in the server, with detailed results presented in Section \ref{sec:experiment}. 
\section{Experiments} \label{sec:experiment}
In this section, we comprehensively evaluate \texttt{cwFedAvg} with \texttt{WDR} across various settings and analyze its convergence behavior against \texttt{FedAvg}.

\subsection{Experimental Setup}
{\bf Datasets.}  
We evaluate our approach on four standard benchmark datasets: MNIST~\citep{deng2012mnist}, CIFAR-10/100~\citep{krizhevsky2009learning}, and Tiny ImageNet~\citep{chrabaszcz2017downsampled}. Each is partitioned into 75\% training and 25\% test splits. We examine two data heterogeneity settings: (1) pathological setting, where each client holds data from only a subset of classes---specifically two classes for MNIST and CIFAR-10, 10 classes for CIFAR-100, and 20 classes for Tiny ImageNet---and (2) practical setting, where client data distributions follow a Dirichlet distribution parameterized by $\alpha$, with smaller values of $\alpha$ corresponding to higher statistical heterogeneity across clients. We use $\alpha=0.1$ as the default setting. Data distributions varying $\alpha$ are provided in the supplementary materials.

\noindent {\bf Evaluation Protocol.} 
We measure performance using average test accuracy across all clients per FL round and report results from the best-performing round~\citep{t2020personalized, huang2021personalized, zhang2023fedala}.
We compare our method against eight baselines across four categories: (1) traditional methods: \texttt{FedAvg} and \texttt{FedProx}, (2) aggregation-based methods: \texttt{FedAMP} and \texttt{FedFomo}, (3) clustering-based methods: \texttt{CFL} and \texttt{IFCA}, and (4) regularization-based methods: \texttt{FedNH} and \texttt{FedUV}.

\noindent {\bf Implementation.} 
Our experiments employ two model architectures: a 4-layer CNN with ReLU activation functions (for MNIST, CIFAR-10/100, and Tiny ImageNet) and ResNet-18 \citep{he2016deep} (for Tiny ImageNet). Following the settings of \texttt{FedAvg}~\cite{mcmahan2017communication}, we set the training configuration to 20 clients as the default participating in each round, learning rate of 0.005, batch size of 10, and 1 local epoch. For regularizing clients with \texttt{WDR}, we set $\lambda$ to 10 for MNIST and CIFAR-10, 1000 for CIFAR-100, and 2000 for Tiny ImageNet. Each experiment runs for 1,000 communication rounds to ensure convergence. All results are averaged over three independent runs with different random seeds. Experimental details are provided in the supplementary materials, and code is available at: https://github.com/regulationLee/cwFedAvg

\begin{table*}[t]
  \centering
  \setlength{\tabcolsep}{12pt}
  {\fontsize{9}{11}\selectfont
  \begin{tabular}{@{}llrrrrr@{}}
    \toprule
    \multirow{2}{*}{Algorithm} & \multirow{2}{*}{Comm. Cost} & \multicolumn{2}{c}{Number of clients} & \multicolumn{3}{c}{Data heterogeneity} \\
    \cmidrule(l){3-4} \cmidrule(l){5-7}
    &  & 50 Clients & 100 Clients & $\alpha{=}0.01$ & $\alpha{=}0.5$ & $\alpha{=}1.0$ \\
    \midrule
    \texttt{FedAvg} & $2 \cdot \Sigma$ & 
    32.63$\pm$0.34 & 32.32$\pm$0.30 &
    28.00$\pm$0.92 & 36.18$\pm$0.28 & 36.75$\pm$0.34 \\
    
    \texttt{FedProx} & $2 \cdot \Sigma$ &
    33.22$\pm$0.20 & 32.64$\pm$0.21 &
    27.89$\pm$0.24 & 35.93$\pm$0.31 & 36.65$\pm$0.39 \\

    \midrule
    
    \texttt{FedAMP} & $2 \cdot \Sigma$ &
    44.97$\pm$0.27 & 41.37$\pm$0.35 &
    73.46$\pm$0.40 & 25.41$\pm$0.14 & 21.23$\pm$0.40 \\
    
    \texttt{FedFomo} & $(1{+}M) \cdot \Sigma$ &
    42.62$\pm$0.62 & 38.62$\pm$0.08 &
    71.30$\pm$0.03 & 25.43$\pm$0.58 & 18.95$\pm$0.34 \\
    
    \texttt{CFL} & $2 \cdot \Sigma$ &
    32.83$\pm$0.78 & 32.88$\pm$0.23 &
    27.67$\pm$0.17 & 38.32$\pm$0.47 & 36.80$\pm$0.07 \\
    
    \texttt{IFCA} & $(1{+}C) \cdot \Sigma$ &
    29.17$\pm$0.20 & 26.56$\pm$0.45 &
    53.89$\pm$3.58 & 25.87$\pm$0.57 & 22.27$\pm$1.14 \\
    
    \texttt{FedNH} & $2 \cdot \Sigma$ & 
    33.14$\pm$0.46 & 32.73$\pm$0.24 &
    25.48$\pm$0.25 & 37.13$\pm$0.41 & 20.41$\pm$0.15 \\
    
    \texttt{FedUV} & $2 \cdot \Sigma$ & 
    44.30$\pm$0.14 & 40.91$\pm$0.22 &
    72.67$\pm$0.12 & 27.23$\pm$0.25 & 37.41$\pm$0.34 \\

    \midrule
    
    \texttt{cwFedAvg} (Output) & $2 \cdot \Sigma$ &
    \textbf{55.90$\pm$0.35} & \textbf{53.54$\pm$0.79} &
    \textbf{75.20$\pm$0.21} & \textbf{40.78$\pm$0.93} & \textbf{37.50$\pm$0.10} \\
    \bottomrule
  \end{tabular}
  }
  \caption{Communication cost formulation and classification accuracy (\%) across different settings for CIFAR-100. $\Sigma$ denotes total model parameters, and $C$ denotes the number of clusters. \texttt{cwFedAvg} (Output) denotes \texttt{cwFedAvg} selectively applied to the output layer.}
  \label{table:comprehensive}
\end{table*}

\begin{table*}[th]
\centering
\setlength{\tabcolsep}{6.pt}
{\fontsize{9}{11}\selectfont
\begin{tabular}{@{}llrrrrr@{}}
\toprule
\multicolumn{2}{c}{Applied algorithm and layer} & \multicolumn{2}{c}{Pathological setting} & \multicolumn{3}{c}{Practical setting ($\alpha=0.1$)} \\
\cmidrule(lr){1-2} \cmidrule(l){3-4} \cmidrule(l){5-7}
\texttt{FedAvg} & \texttt{cwFedAvg} & CIFAR-10 & CIFAR-100 & CIFAR-10 & CIFAR-100 & Tiny ImageNet \\
\midrule
-  & Conv1-Conv2-FC-Output     & 90.98$\pm$0.15 & 65.91$\pm$0.29 & 88.40$\pm$0.13 & 54.99$\pm$0.27 & 38.94$\pm$0.38 \\
Conv1  & Conv2-FC-Output       & 90.99$\pm$0.11 & 65.71$\pm$0.19 & 88.55$\pm$0.07 & 55.01$\pm$0.25 & 38.76$\pm$0.51 \\
Conv1-Conv2 & FC-Output       & 90.93$\pm$0.06 & 65.22$\pm$0.16 & 88.45$\pm$0.06 & 54.98$\pm$0.28 & 38.78$\pm$0.68 \\
Conv1-Conv2-FC  & Output       & \textbf{91.23$\pm$0.04} & \textbf{67.50$\pm$0.14} & \textbf{88.65$\pm$0.19} & \textbf{56.29$\pm$0.18} & \textbf{41.38$\pm$0.12} \\
Conv1-Conv2-FC-Output  & -       & 60.68$\pm$0.84 & 28.22$\pm$0.32 & 61.94$\pm$0.56 & 32.44$\pm$0.42 & 21.35$\pm$0.12 \\
\bottomrule
\end{tabular}
}
\caption{Classification accuracy (\%) with selective application of \texttt{cwFedAvg} to a 4-layer CNN (Input-Conv1-Conv2-FC-Output).}
\label{table:seletive_result}
\vspace{-10pt}
\end{table*}

\subsection{Performance Comparison and Analysis}
\noindent {\bf Classification Performance.}  
Table \ref{table:total_result} illustrates that PFL methods typically outperform traditional FL methods on non-IID data owing to the poor personalization ability of the global model. Among the PFL methods, \texttt{cwFedAvg} consistently outperforms all other settings except for one case in the CIFAR-10 practical setting, where the performance gap is minimal. Notably, \texttt{cwFedAvg} exhibits significant improvements over the others when the class count is large, typically indicating high data heterogeneity. The results for a text dataset are provided in the supplementary materials.

\noindent {\bf Communication Cost.} 
Table \ref{table:comprehensive} presents the communication cost formulation per iteration. \texttt{cwFedAvg} maintains the same communication overhead as \texttt{FedAvg} as explained in Section \ref{sec:cwFedAVG}. In contrast, other PFL approaches often incur larger communication costs: \texttt{FedFomo} and \texttt{IFCA} require downloading additional models, which increases downstream communication.

\noindent {\bf Client Scalability.}  
Table \ref{table:comprehensive} demonstrates the performance scaling to the number of clients. All PFL methods exhibit performance degradation as the number of clients increases. However, \texttt{cwFedAvg} maintains high performance even with larger client counts compared with other methods, thus beneficial for large-scale deployments of devices.

\noindent {\bf Data Heterogeneity and Convergence Analysis.}   
We evaluate the model accuracy under non-IID data distributions controlled by the Dirichlet distribution parameter $\alpha$. Table~\ref{table:comprehensive} shows that \texttt{cwFedAvg} maintains robust performance across different $\alpha$ values. In contrast, several PFL methods underperform compared with traditional approaches as data becomes more IID (larger $\alpha$). Moreover, the performance of \texttt{cwFedAvg} converges to that of \texttt{FedAvg} as $\alpha$ increases, which validates our analysis of global model relationships in Section~\ref{sec:cwfedavg_comparative}.
\begin{figure}[th]
    \centering
    \begin{subfigure}[b]{0.485\columnwidth}
        \includegraphics[width=\columnwidth]{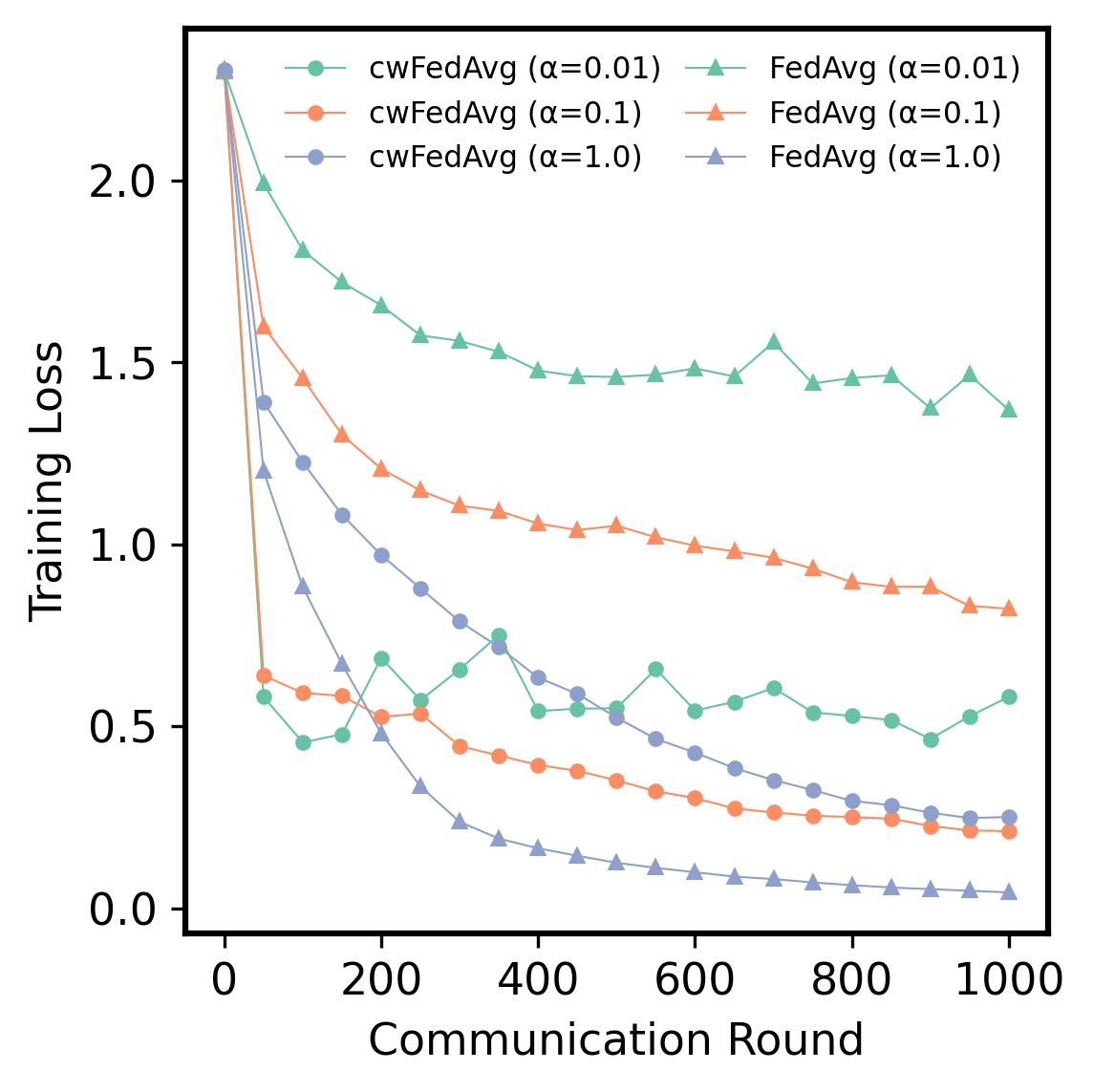} 
        \caption{CIFAR-10}  
    \end{subfigure}  
    \hfill
    \begin{subfigure}[b]{0.485\columnwidth}
        \includegraphics[width=\columnwidth]{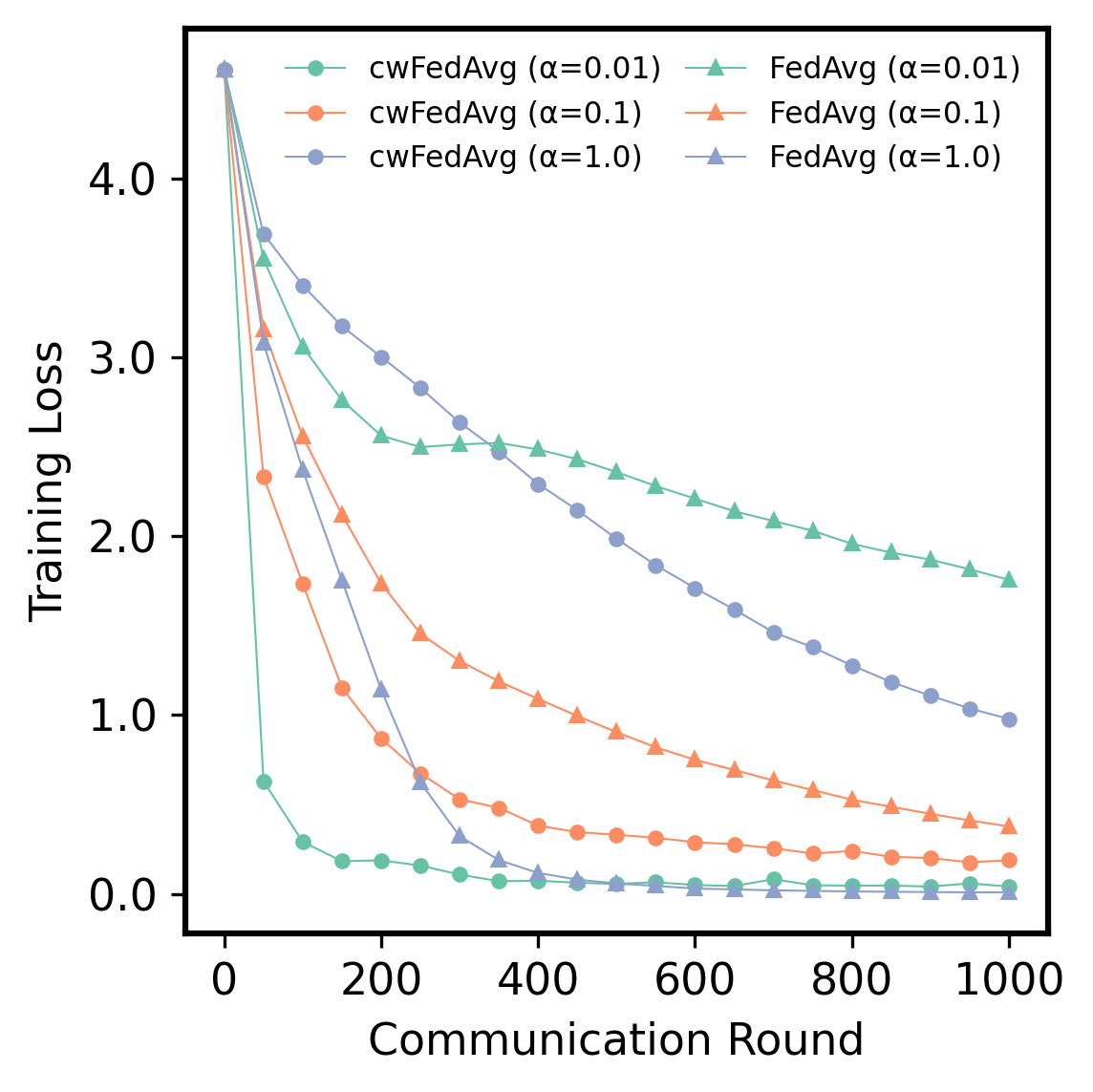} 
        \caption{CIFAR-100}  
    \end{subfigure}
    \vspace{-5pt}
    \caption{Convergence comparison of \texttt{cwFedAvg} and \texttt{FedAvg}.}
    \label{fig:convergence}
    \vspace{-15pt}
\end{figure}
Figure~\ref{fig:convergence} shows the average training loss for 20 clients under varying data heterogeneity.
The results demonstrate that \texttt{cwFedAvg} converges significantly faster than \texttt{FedAvg} on both CIFAR-10 and CIFAR-100 datasets when data are highly heterogeneous ($\alpha=0.01$ and $0.1$).
This aligns with our pathway analysis in Figure~\ref{fig:pathway}, as local models in \texttt{cwFedAvg} adapt to their class distributions.
However, when data is nearly IID ($\alpha=1.0$), \texttt{FedAvg} converges faster.
We provide per-client training loss comparisons in the supplementary materials.

\noindent {\bf Selective Application of \texttt{cwFedAvg}.} 
Unlike \texttt{FedAvg}, PFL methods often require additional memory for storing multiple models, such as cluster models in \texttt{IFCA} and class-specific global models in \texttt{cwFedAvg}. We selectively apply \texttt{cwFedAvg} to upper layers only to reduce memory overhead. Across various datasets, applying \texttt{cwFedAvg} solely to the output layer achieves optimal performance (Table~\ref{table:seletive_result}). This result aligns with the conventional understanding that upper layers capture class-specific features. Memory cost comparison is provided in the supplementary materials.

\begin{table}[tbh]
  \centering
  \setlength{\tabcolsep}{7.5pt}
  {\fontsize{9}{11}\selectfont
  \begin{tabular}{@{}llrrr@{}}
    \toprule
    Distribution & \texttt{WDR} &  CIFAR-100 & Tiny ImageNet \\
    \midrule
    Empirical ($\boldsymbol{p}$)            &  \xmark & 48.27$\pm$0.74 & 31.09$\pm$0.14 \\
    Approximation ($\boldsymbol{\tilde{p}}$)    & \xmark & 32.25$\pm$0.56 & 20.08$\pm$0.96 \\
    Approximation ($\boldsymbol{\tilde{p}}$)    & \cmark   & \textbf{54.99$\pm$0.27} & \textbf{38.94$\pm$0.38} \\
    \bottomrule
  \end{tabular} 
  }
  \caption{Ablation study results showing classification accuracy (\%) of \texttt{cwFedAvg} under practical settings.}
  \label{table:ablation}
\end{table}

\begin{figure}[th]
    \centering
    \begin{subfigure}[b]{0.48\columnwidth}
        \includegraphics[width=\columnwidth]{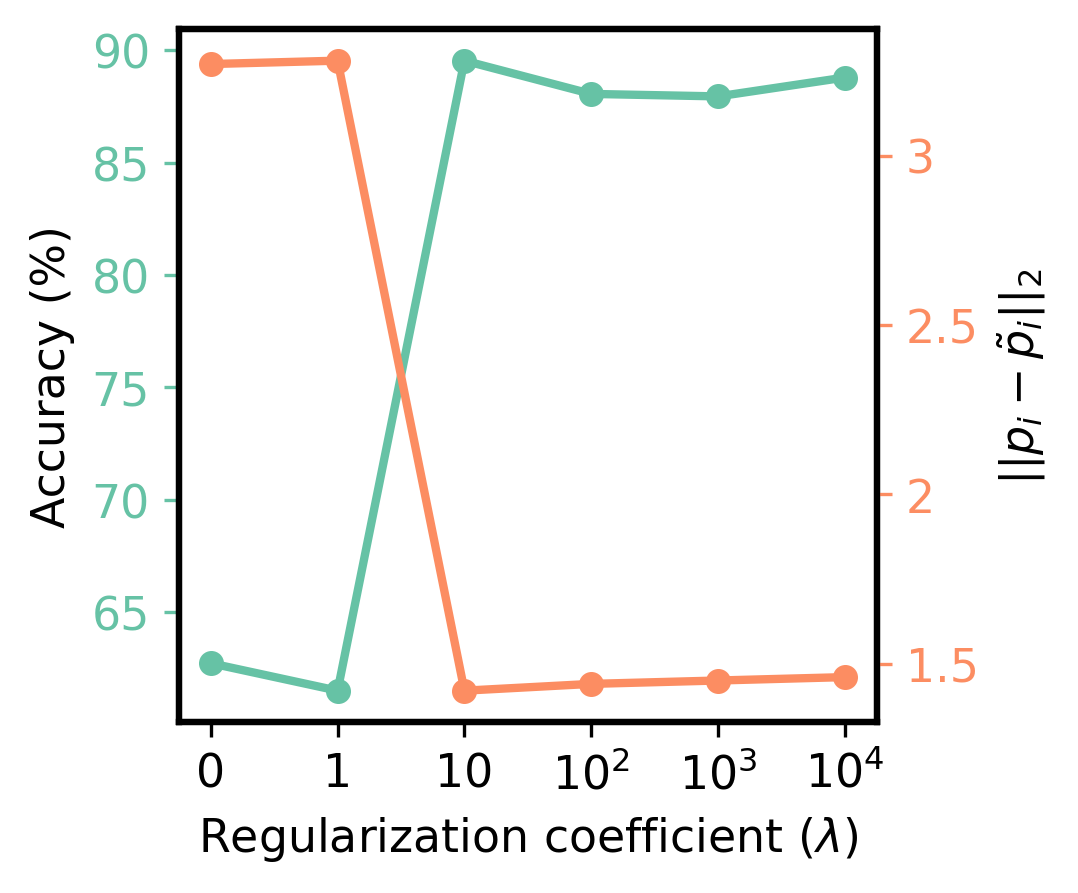} 
        \caption{CIFAR-10} \label{fig:lambda_cifar10}
    \end{subfigure}  
    \,
    \begin{subfigure}[b]{0.48\columnwidth}
        \includegraphics[width=\columnwidth]{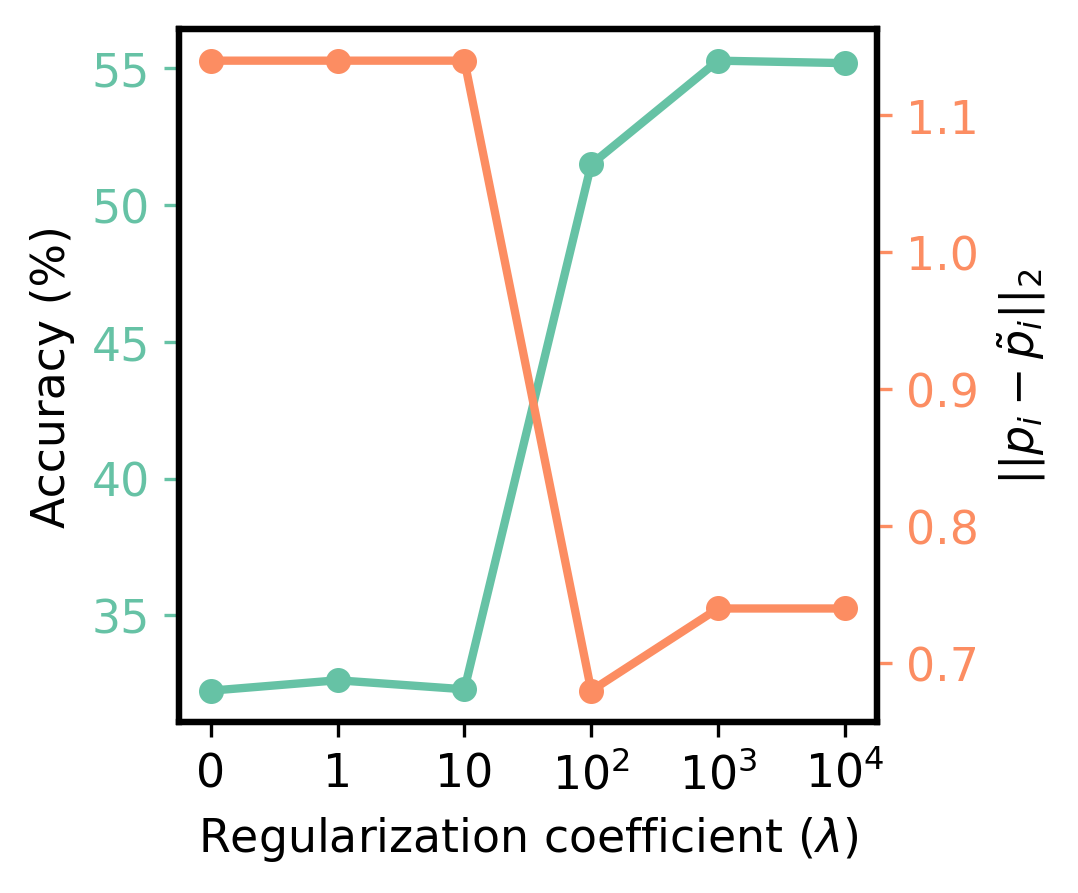} 
        \caption{CIFAR-100} \label{fig:lambda_cifar100}
    \end{subfigure}
    \caption{
    Influence of regularization coefficient ($\lambda$) on accuracy and regularization term ($\|\boldsymbol{p}_i - \tilde{\boldsymbol{p}}_i\|_2$) in practical settings.}
    \label{fig:lambda_relation}
\end{figure}

\noindent {\bf Ablation Study.}
We examine the effect of empirical data distribution $\boldsymbol{p}$, approximated distribution $\boldsymbol{\tilde{p}}$, and \texttt{WDR} in practical settings when \texttt{cwFedAvg} is applied to all layers. First, as expected, with $\boldsymbol{\tilde{p}}$ and without \texttt{WDR}, \texttt{cwFedAvg} (the second row of Table~\ref{table:ablation}) shows performance comparable to \texttt{FedAvg} (the first row of Table~\ref{table:total_result}) as it fails to encode and extract class-specific information properly. Notably, \texttt{cwFedAvg} with $\boldsymbol{\tilde{p}}$ and \texttt{WDR} significantly outperforms \texttt{cwFedAvg} with $\boldsymbol{p}$ without \texttt{WDR} when the class count is large. This result suggests that \texttt{WDR} effectively approximates $\boldsymbol{p}$ while regularizing pathways to be class-specific.

\noindent {\bf Impact of Regularization Coefficient.} 
Figure \ref{fig:lambda_relation} illustrates the impact of regularization coefficient ($\lambda$) on accuracy and the regularization term ($\left\|\boldsymbol{p}_i - \tilde{\boldsymbol{p}}_i\right\|_2$) for CIFAR-10/100 practical heterogeneous settings. Increasing $\lambda$ reduces the regularization term, enabling \texttt{cwFedAvg} to effectively utilize $\tilde{\boldsymbol{p}}_i$ for class-wise aggregation. Based on these results, we select the optimal $\lambda$ that maximizes regularization strength while maintaining accuracy.

\begin{figure}[th]
    \centering
    \begin{subfigure}[b]{0.15\textwidth}
        \includegraphics[width=\textwidth]{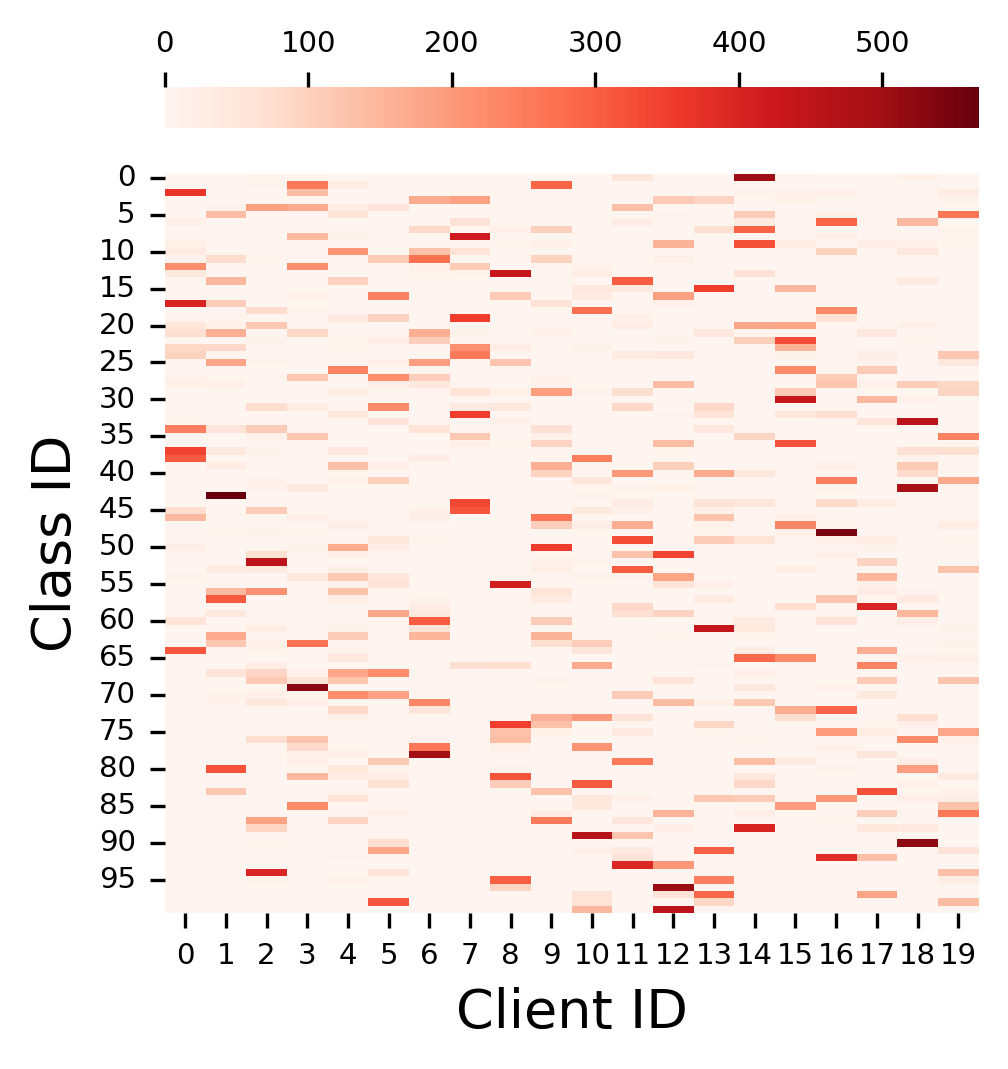}
        \caption{Data distribution of \\ clients} \label{fig:heatmap_cifar100_prac1_dist}
    \end{subfigure}
    \hfill
    \begin{subfigure}[b]{0.15\textwidth}
        \includegraphics[width=\textwidth]{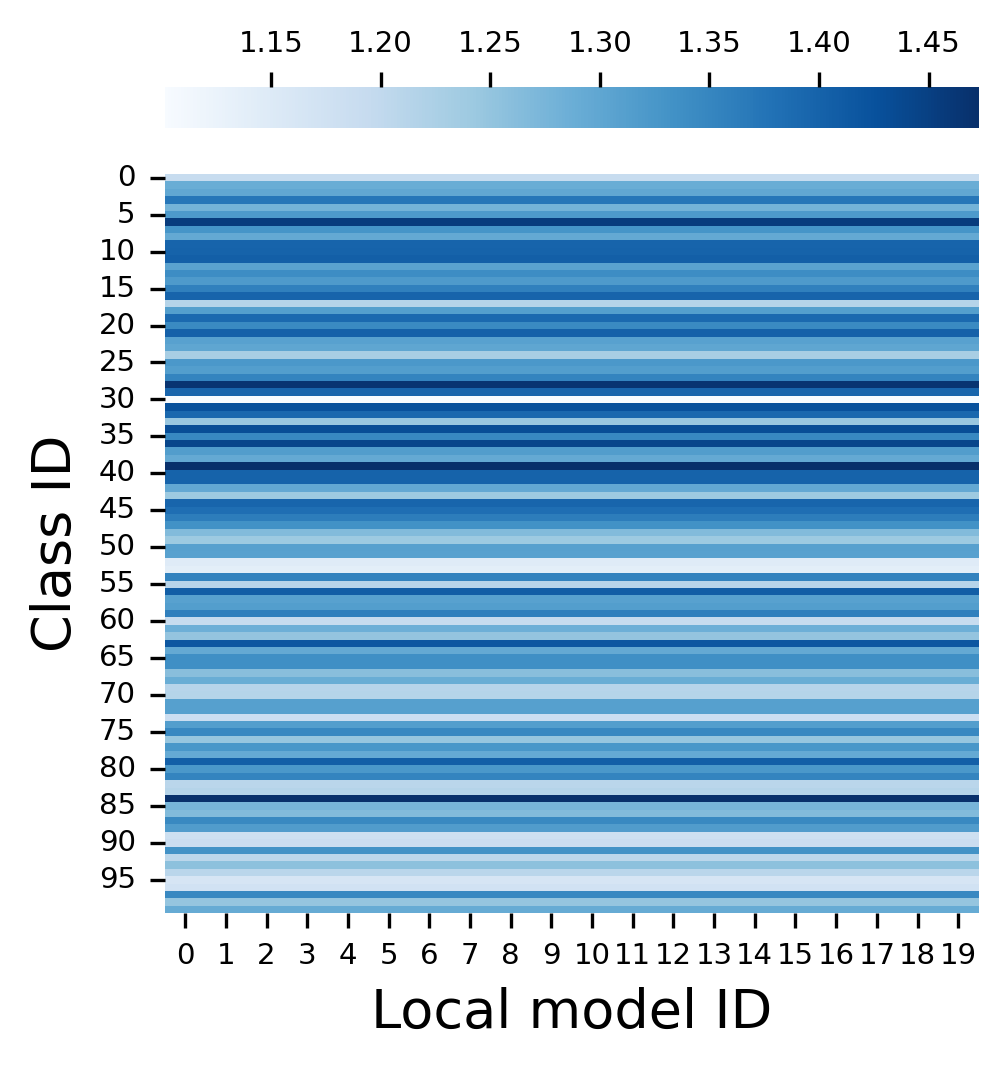}
        \caption{Local models of \\ \texttt{FedAvg}} 
    \end{subfigure}
    \hfill
    \begin{subfigure}[b]{0.15\textwidth}
        \includegraphics[width=\textwidth]{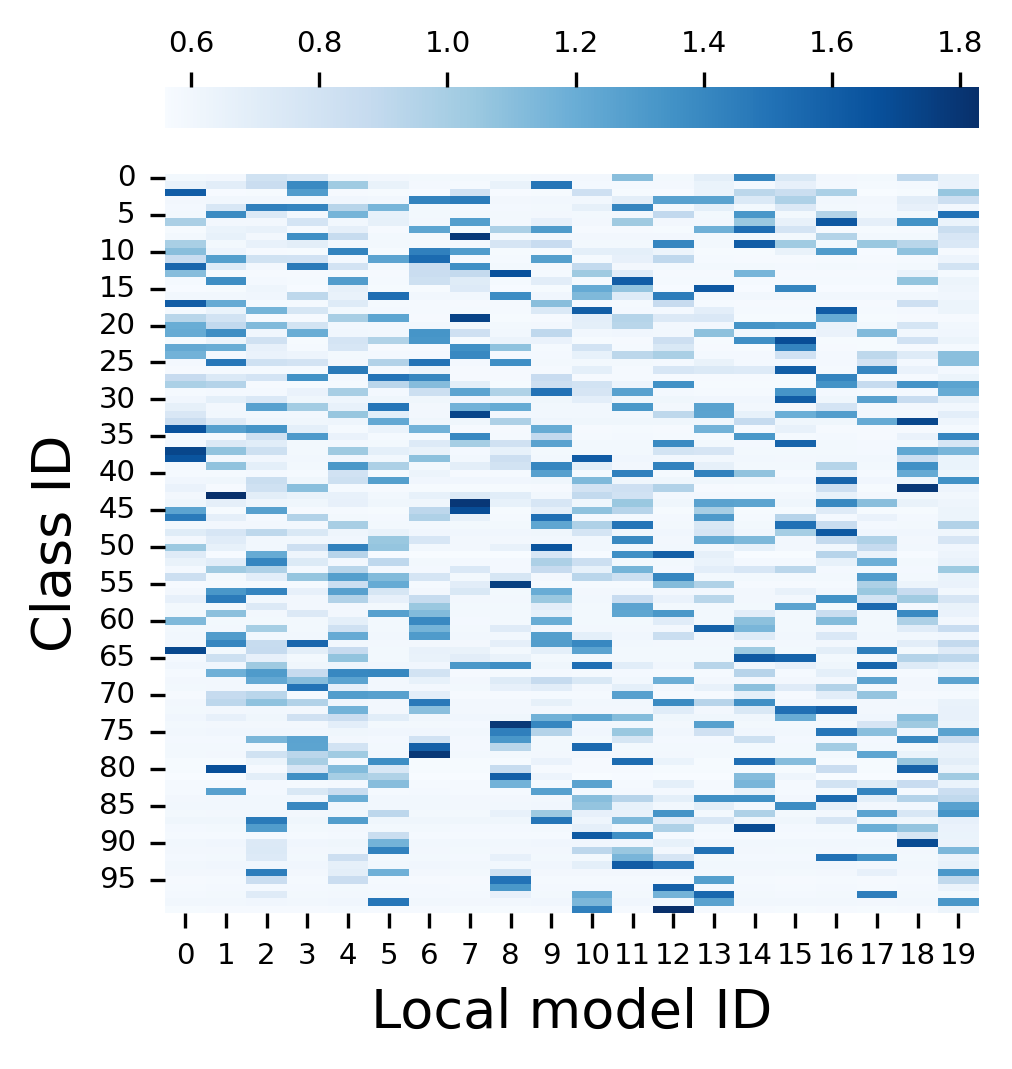} 
        \caption{Local models of \\ \texttt{FedAMP}} \label{fig:heatmap_cifar100_prac1_fedamp}
    \end{subfigure} 
    \\
    \begin{subfigure}[b]{0.15\textwidth}
        \includegraphics[width=\textwidth]{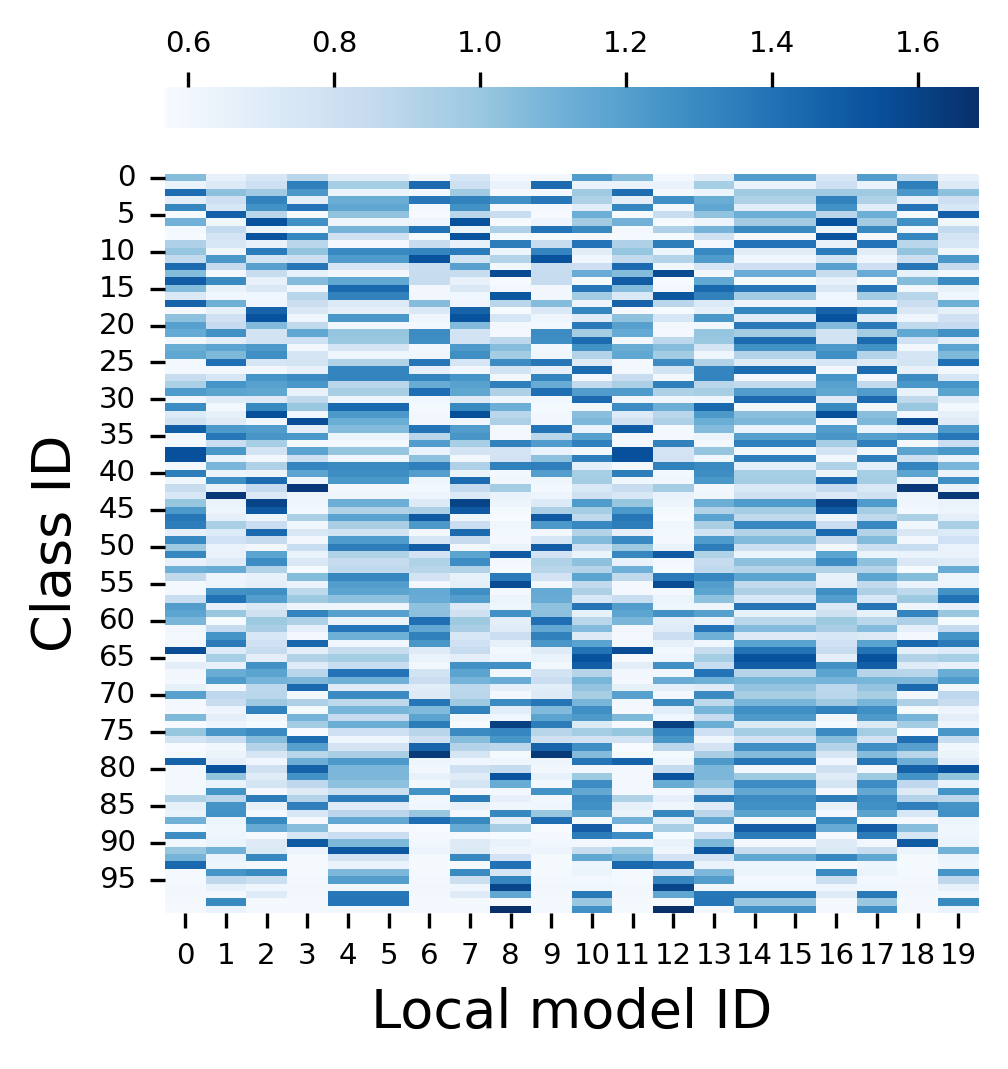}
        \caption{Local models of \\ \texttt{IFCA}} \label{fig:heatmap_cifar100_prac1_ifca}
    \end{subfigure}
    \hfill
    \begin{subfigure}[b]{0.15\textwidth}
        \includegraphics[width=\textwidth]{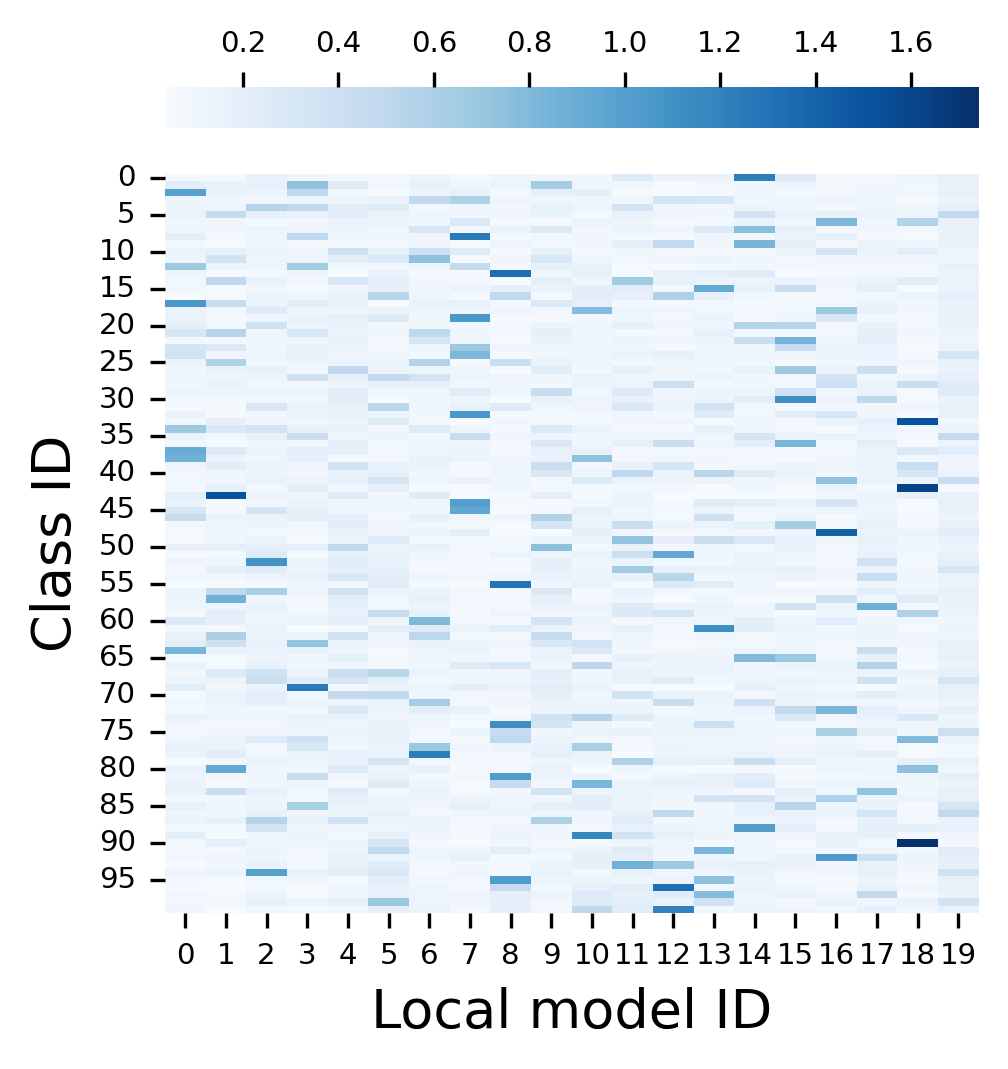}
        \caption{Local models of \\ \texttt{cwFedAvg} w/ \texttt{WDR}} 
        \label{fig:heatmap_cifar100_prac1_cwfedavg_local}
    \end{subfigure} 
    \hfill
    \begin{subfigure}[b]{0.15\textwidth}
        \includegraphics[width=\textwidth]{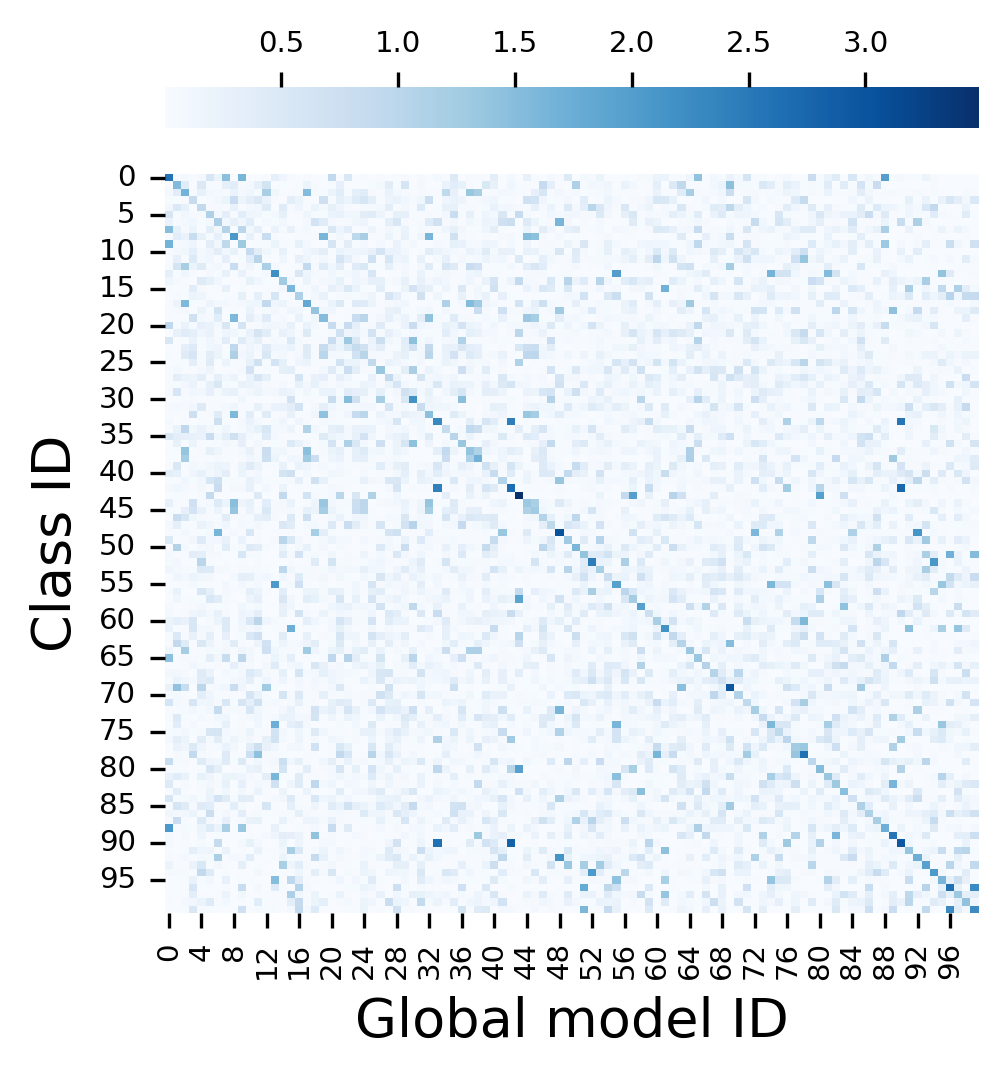}
        \caption{Global models of \\ \texttt{cwFedAvg} w/ \texttt{WDR}} \label{fig:heatmap_cifar100_prac_cwfedavg_wdr_global}
    \end{subfigure}
    \caption{Heatmaps depicting the data distribution and $\ell_2$-norms of output layer weight vectors for the CIFAR-100 practical setting. 
    (a) Each cell represents the number of data samples belonging to class $j$ for client $i$. (b)--(f) Each cell shows the $\ell_2$-norm of output layer weight vector, $\left\|\mathbf{w}_{i,j}\right\|_2$.
    }
    \label{fig:heatmap_cifar100_prac1}
\end{figure}

\subsection{Personalization for Many-Class and Highly Imbalanced Data}
Figure~\ref{fig:heatmap_cifar100_prac1} shows heatmaps of the empirical data distribution and the $\ell_2$-norms of output layer weight vectors ($\|\mathbf{w}_{i,j}\|_2$) for the CIFAR-100 practical setting. The results demonstrate similar patterns observed in the CIFAR-10 pathological setting (Figure~\ref{fig:heatmap_cifar10_pat}). 
Notably, Figure~\ref{fig:heatmap_cifar100_prac1_fedamp} (\texttt{FedAMP}) exhibits a pattern similar to Figure~\ref{fig:heatmap_cifar100_prac1_cwfedavg_local} (\texttt{cwFedAvg} with \texttt{WDR}), which aligns with their shared personalized aggregation-based approach, without \texttt{FedAMP}'s explicit regularization of output layer weights. Figure~\ref{fig:heatmap_cifar100_prac1_ifca} (\texttt{IFCA}), a clustering-based method, reveals distinct cluster formations but does not show personalization patterns.
Figures~\ref{fig:heatmap_cifar100_prac1_cwfedavg_local} and~\ref{fig:heatmap_cifar100_prac_cwfedavg_wdr_global} confirm that our class distribution estimation with \texttt{WDR} works appropriately even for many-class and highly imbalanced data settings.
Additional visualizations for other settings are provided in the supplementary materials.
\section{Limitation and Conclusion} \label{sec:discussion}
The \texttt{cwFedAvg} method requires storing multiple global models on the server, equal to the number of classes. This requires more server memory than \texttt{FedAvg}, but the overhead can be significantly reduced by applying \texttt{cwFedAvg} to only upper layers with higher performance. 

Despite its simplicity, \texttt{cwFedAvg} achieves efficient personalization and provides significant advantages over existing algorithms in cross-device PFL scenarios.
\texttt{cwFedAvg} eliminates the overhead of additional client-side training or downloading other clients' models that conventional PFL methods often require. The computational complexity of the algorithm scales linearly with the number of clients, whereas other PFL methods often require quadratic complexity owing to pairwise information exchange. This scalability makes \texttt{cwFedAvg} particularly suitable for large-scale deployments, as real-world applications typically involve a fixed number of classes while the number of participating clients grows.
Furthermore, as an aggregation module, \texttt{cwFedAvg} with \texttt{WDR} can be integrated into existing PFL frameworks as a replacement for \texttt{FedAvg}, offering potential performance enhancements.

{
    \small
    \bibliographystyle{ieeenat_fullname}
    \bibliography{main}
}

\clearpage
\setcounter{page}{1}
\maketitlesupplementary

\section*{A. Comparison of Aggregation Process}
\label{sec:illustration_comparison}
\begin{figure}[hbt]
    \centering
    \begin{subfigure}[b]{0.235\textwidth}
        \includegraphics[width=\textwidth]{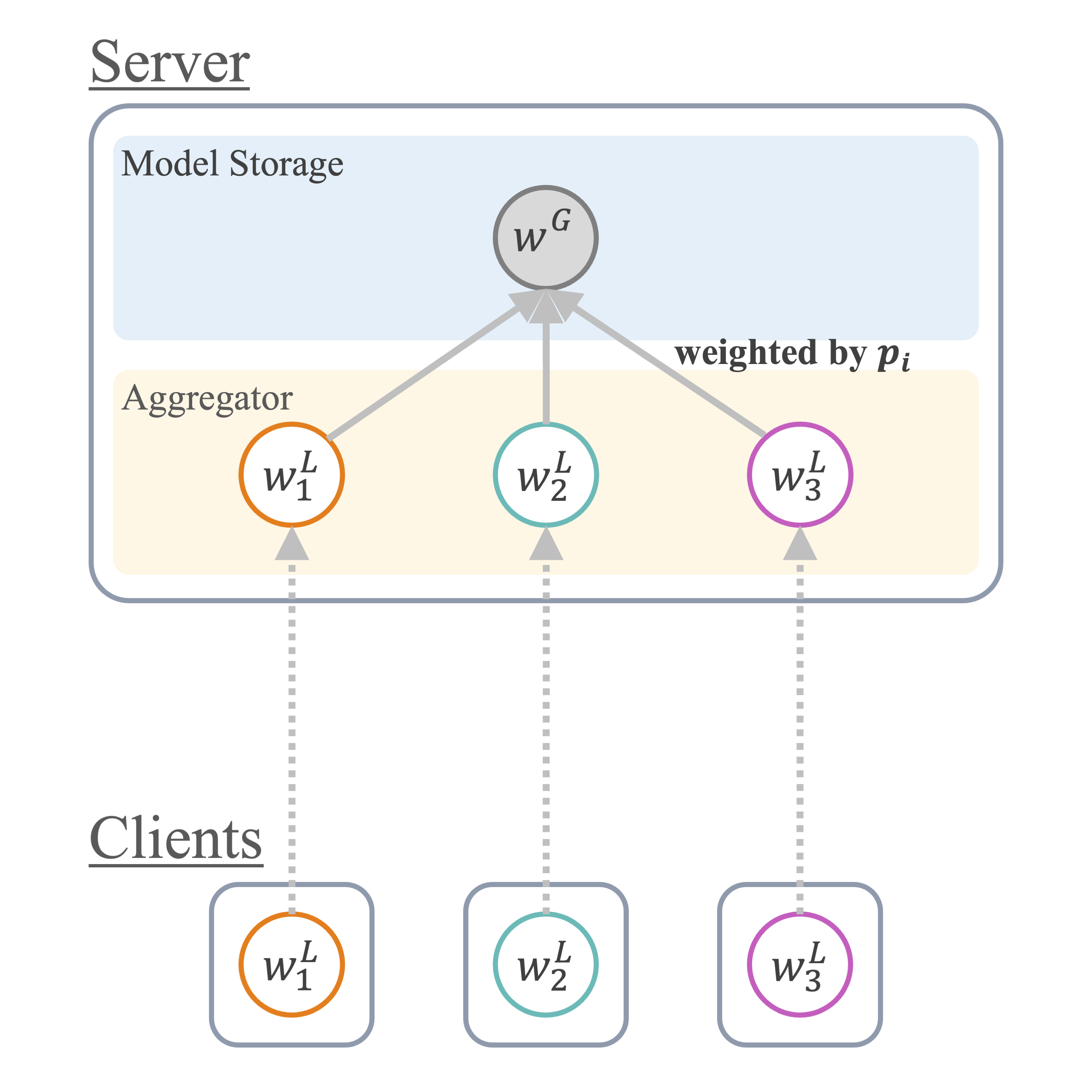} 
        \caption{Local model aggregation \\(\texttt{FedAvg})} \label{fig:fedavg_upload}
    \end{subfigure}
    \begin{subfigure}[b]{0.235\textwidth}
        \includegraphics[width=\textwidth]{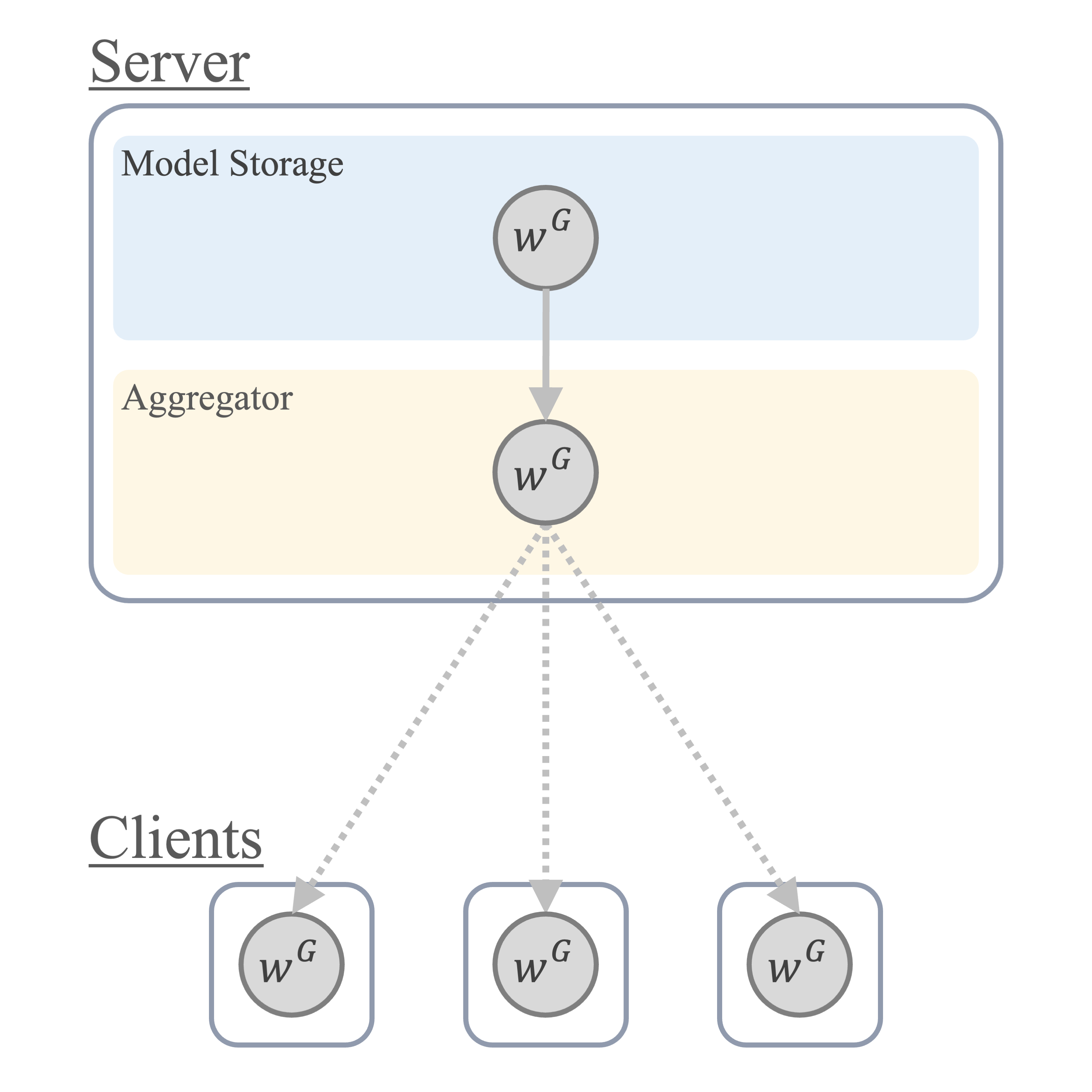} 
        \caption{Global model replication \\ (\texttt{FedAvg})}
        \label{fig:fedavg_download}
    \end{subfigure} 
    \\
    \begin{subfigure}[b]{0.23\textwidth}
        \includegraphics[width=\textwidth]{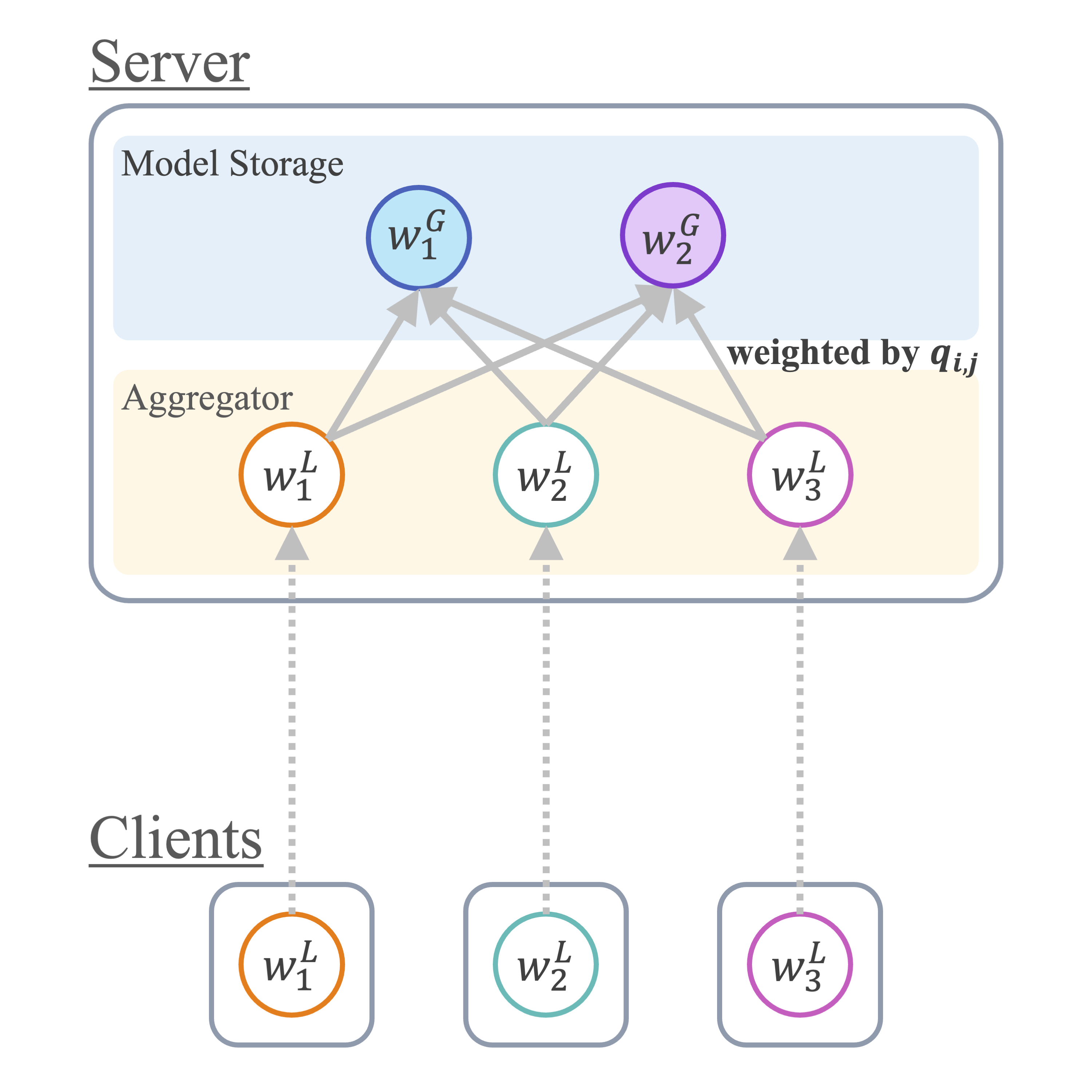} 
        \caption{Class-wise local model \\ aggregation (\texttt{cwFedAvg})}
        \label{fig:local_aggregation}
    \end{subfigure}
    \begin{subfigure}[b]{0.23\textwidth}
        \includegraphics[width=\textwidth]{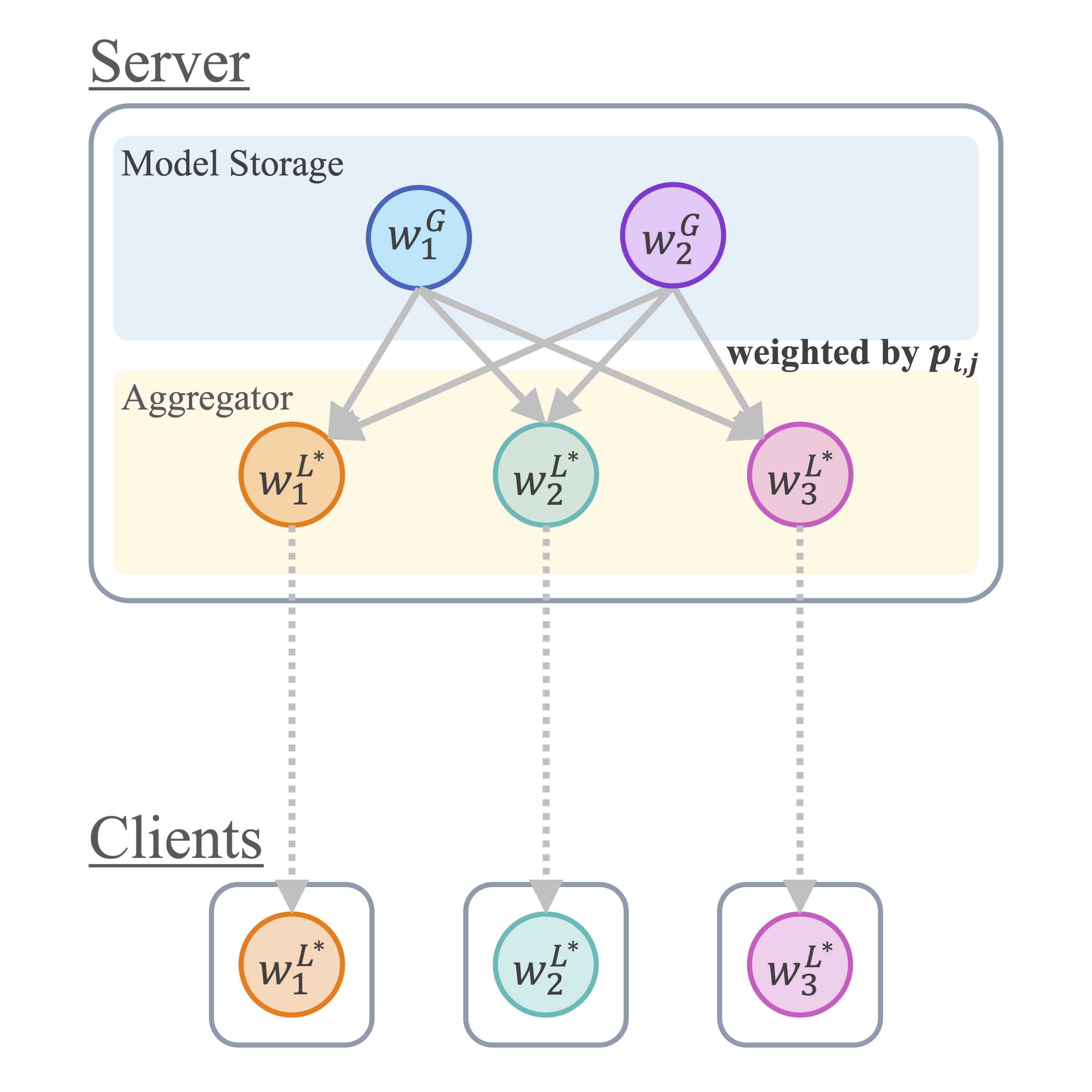} 
        \caption{Class-wise global model \\ aggregation (\texttt{cwFedAvg})}
        \label{fig:global_aggregation}
    \end{subfigure}
    \caption{Comparison of aggregation processes in \texttt{FedAvg} and \texttt{cwFedAvg} with three clients for binary classification task. $*$ denotes the local models updated using class-specific global models.}
    \label{fig:aggregation}
\end{figure}

Figure \ref{fig:aggregation} illustrates the aggregation processes of \texttt{FedAvg} and \texttt{cwFedAvg} using three clients for a binary classification task:
(a) \texttt{FedAvg}: Server aggregates received local models.
(b) Server distributes the aggregated global model to all clients.
(c) \texttt{cwFedAvg}: Server performs class-wise aggregation to create class-specific global models.
(d) Server creates personalized models by combining class-specific global models and distributes them to clients.

\section*{B. Effect of \texttt{WDR} for Many-Class and Highly Imbalanced Data}
\begin{figure}[ht]
    \centering
    \begin{subfigure}[b]{0.294\columnwidth}
        \includegraphics[width=\columnwidth]{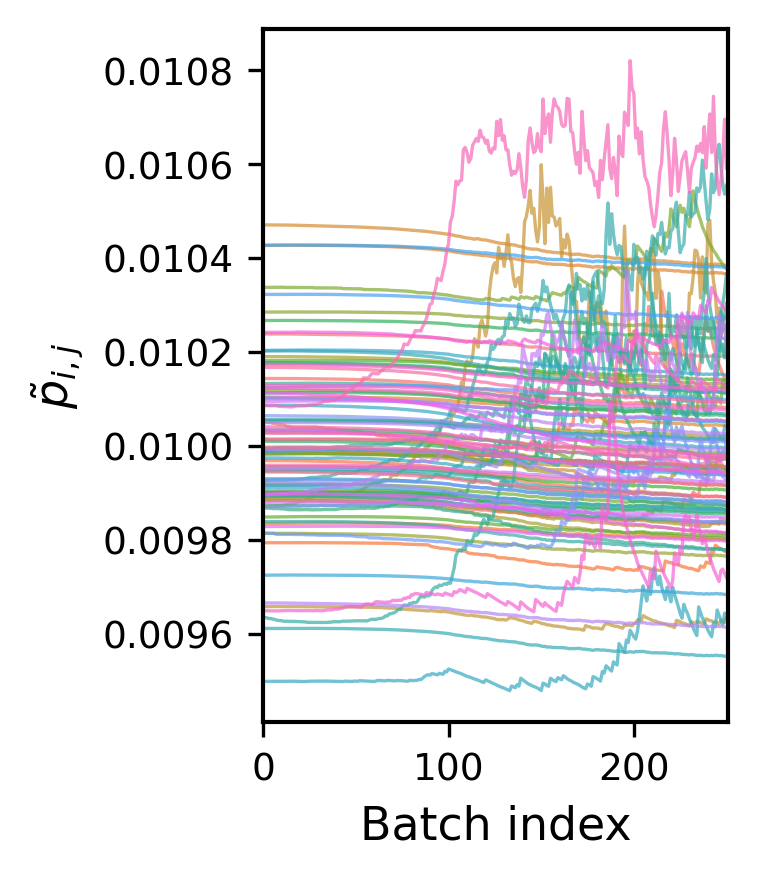} 
        \caption{Train w/o \texttt{WDR}} \label{fig:line_p_cifar100}
    \end{subfigure} 
    \begin{subfigure}[b]{0.275\columnwidth}
        \includegraphics[width=\columnwidth]{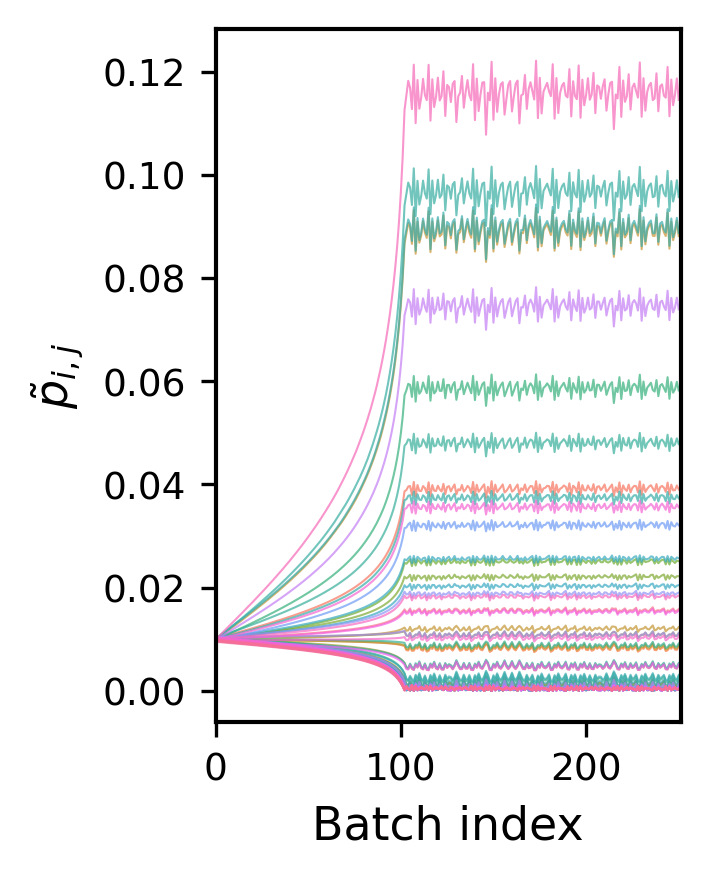} 
        \caption{Train w/ \texttt{WDR}} \label{fig:line_p_wdr_cifar100}
    \end{subfigure} 
    \begin{subfigure}[b]{0.384\columnwidth}
        \includegraphics[width=\columnwidth]{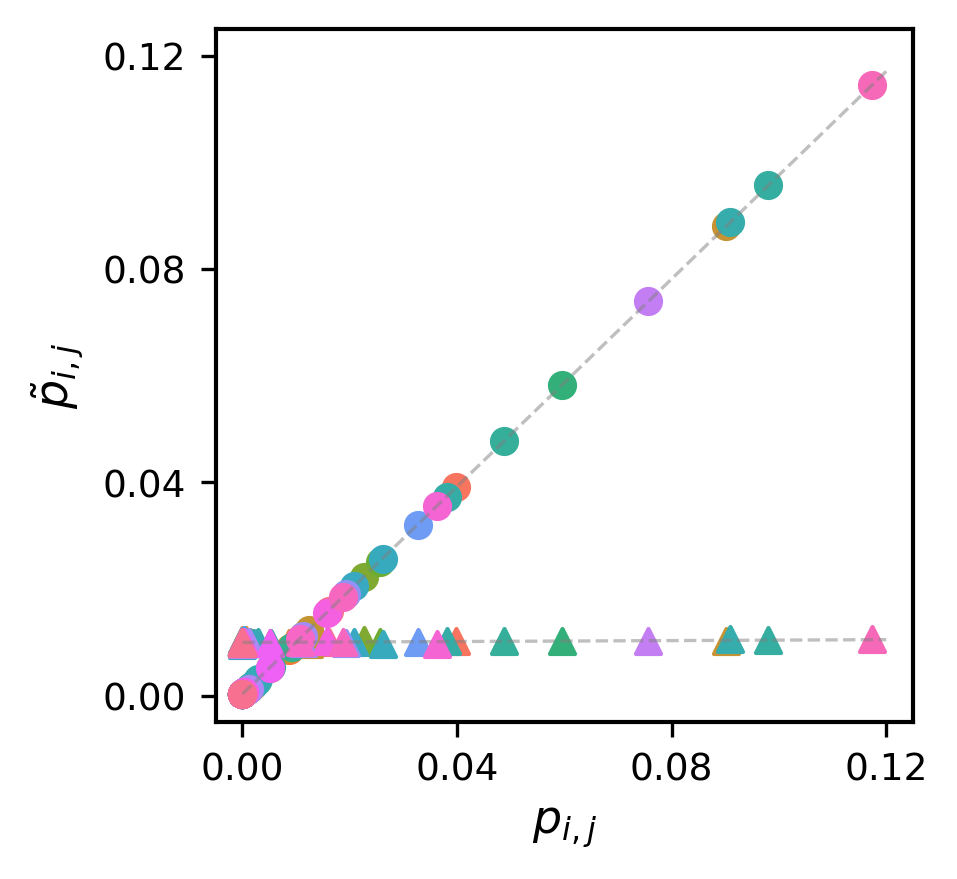} 
        \caption{Correlation} \label{fig:scatter_p_cifar100}
    \end{subfigure}
    \caption{Evolution of $\tilde{p}_{i,j}$ and its correlation with ${p}_{i,j}$ for CIFAR-100 practical setting.}
    \label{fig:norm_correlation_cifar100}
    \vspace{-10pt}    
\end{figure}

Figure~\ref{fig:norm_correlation_cifar100} exhibits similar patterns to the CIFAR-10 pathological setting (Figure~\ref{fig:norm_correlation}). For client ID 11, which contains approximately ten dominant classes, \texttt{WDR} achieves better class separation (Figure~\ref{fig:line_p_wdr_cifar100}), resulting in $\tilde{p}_{i,j}$ values that closely match $p_{i,j}$ (circular markers in Figure~\ref{fig:scatter_p_cifar100}).

\section*{C. Experimental Details}

\subsection*{C.1. CNN Architecture and Hyperparameters} We employ a 4-layer CNN architecture \cite{mcmahan2017communication} composed of two convolutional layers with 5×5 kernels (32 and 64 channels respectively), each paired with 2×2 max pooling. The network terminates with a fully connected layer containing 512 units and ReLU activation, followed by a softmax output layer.
We adopt the hyperparameter settings from Zhang et al. \cite{zhang2023fedala} for baseline algorithms except for \texttt{CFL}, \texttt{IFCA}, \texttt{FedNH} and \texttt{FedUV}. For \texttt{CFL}, \texttt{IFCA}, \texttt{FedNH} and \texttt{FedUV} we follow the configurations specified in their respective papers. A comprehensive list of hyperparameter settings for all baselines is provided in Table \ref{table:hp_settings}.

\begin{table}[ht]
   \centering
   \setlength{\tabcolsep}{7.5pt}
   {\fontsize{9}{11}\selectfont
       \begin{tabular}{ll} 
           \toprule
           Algorithm & Hyperparameter settings \\ 
           \midrule
           \texttt{FedProx} & $\mu$ \begin{footnotesize}(proximal term)\end{footnotesize} $ = 0.001$ \\
           \midrule
           \multirow{3}{*}{\texttt{FedAMP}} & $\alpha_{k}$ \begin{footnotesize}(gradient descent)\end{footnotesize} $ = 1000$ \\
           & $\lambda$ \begin{footnotesize}(regularization)\end{footnotesize} $= 1$ \\
           & $\sigma$ \begin{footnotesize}(attention-inducing function)\end{footnotesize} $= 0.1$ \\
           \midrule
           \multirow{2}{*}{\texttt{CFL}} & $\epsilon_1$ \begin{footnotesize}(norm of averaged updated weight)\end{footnotesize} $=0.4$ \\
           & $\epsilon_2$ \begin{footnotesize}(norm of maximum updated weight)\end{footnotesize} $=0.9$ \\
           \midrule
           \multirow{2}{*}{\texttt{IFCA}} & $k$ \begin{footnotesize}(number of clusters)\end{footnotesize} $ = 2$ for CIFAR-10, \\
           & 8 for CIFAR-100 and Tiny ImageNet \\
           \midrule
           \texttt{FedNH} & $\rho$ \begin{footnotesize}(smoothing parameter)\end{footnotesize} $ = 0.9$ \\
           \midrule
           \multirow{2}{*}{\texttt{FedUV}} & $\mu$ \begin{footnotesize}(classifier variance regularizer)\end{footnotesize} $ = 2.5$ \\
           & $\lambda$ \begin{footnotesize}(Hyperspherical uniformity regularizer)\end{footnotesize} $ = 0.5$ \\
           \bottomrule
       \end{tabular}
   }
   \caption[experiment details]{Hyperparameter settings for the baselines.}
   \label{table:hp_settings}
   \vspace{-10pt}
\end{table}

\subsection*{C.2. Implementation Details} The experiments are implemented in PyTorch 2.4 and conducted on a server with two Intel Xeon Gold 6240R CPUs (96 cores total), 256GB memory, and two NVIDIA RTX A6000 GPUs running Ubuntu 22.04 LTS.

\begin{figure*}[htb]
    \centering
    \begin{subfigure}[b]{0.19\textwidth}
        \includegraphics[width=\textwidth]{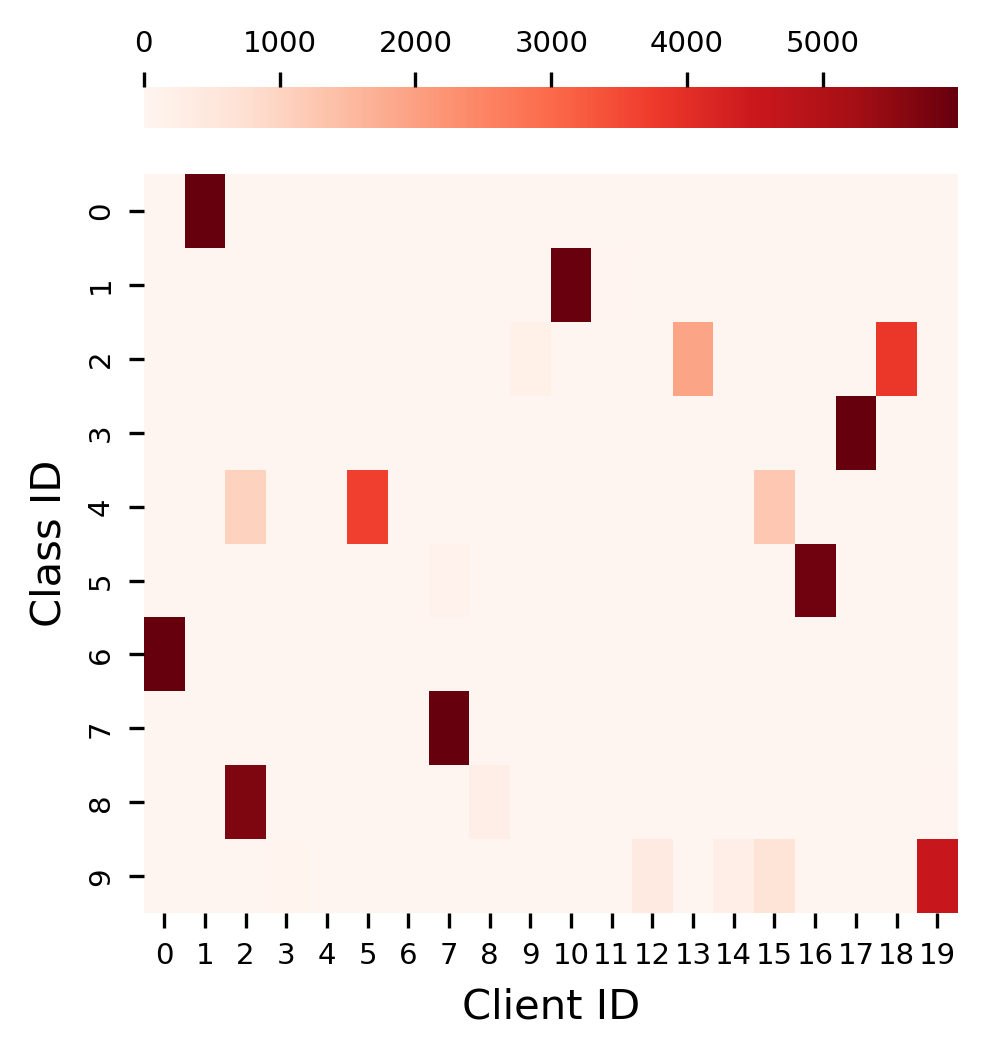}
        \caption{$\alpha=0.01$} 
    \end{subfigure}
    \quad 
    \begin{subfigure}[b]{0.19\textwidth}
        \includegraphics[width=\textwidth]{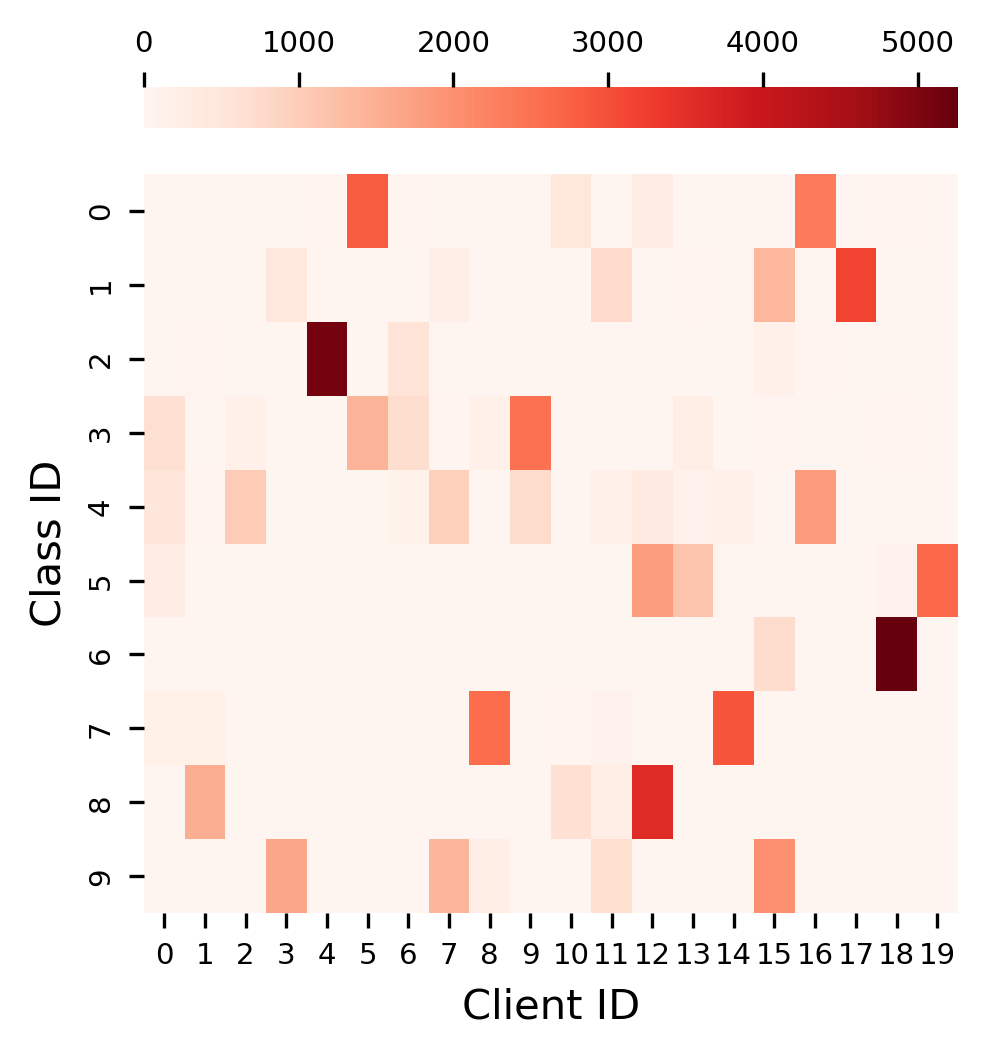}
        \caption{$\alpha=0.1$} 
    \end{subfigure}
    \quad 
    \begin{subfigure}[b]{0.19\textwidth}
        \includegraphics[width=\textwidth]{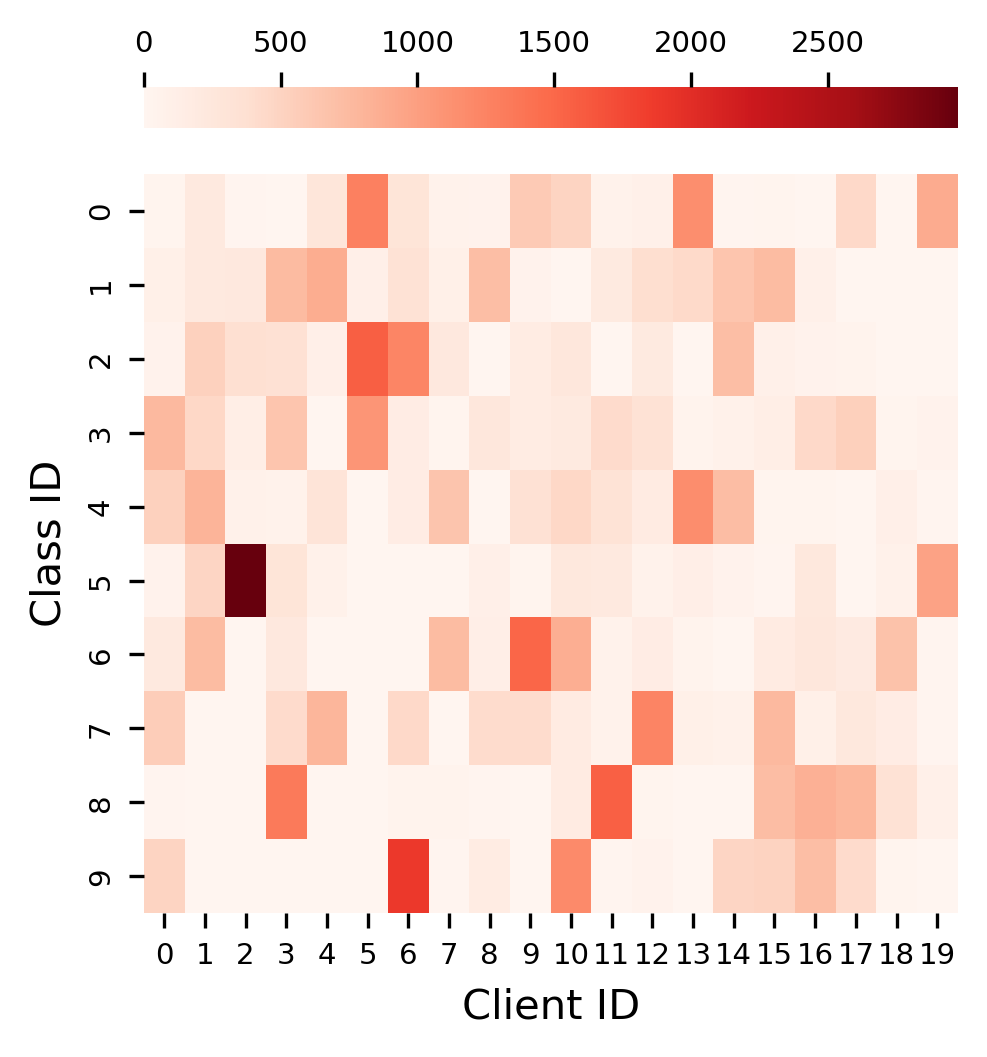}
        \caption{$\alpha=0.5$} 
    \end{subfigure}
    \quad 
    \begin{subfigure}[b]{0.19\textwidth}
        \includegraphics[width=\textwidth]{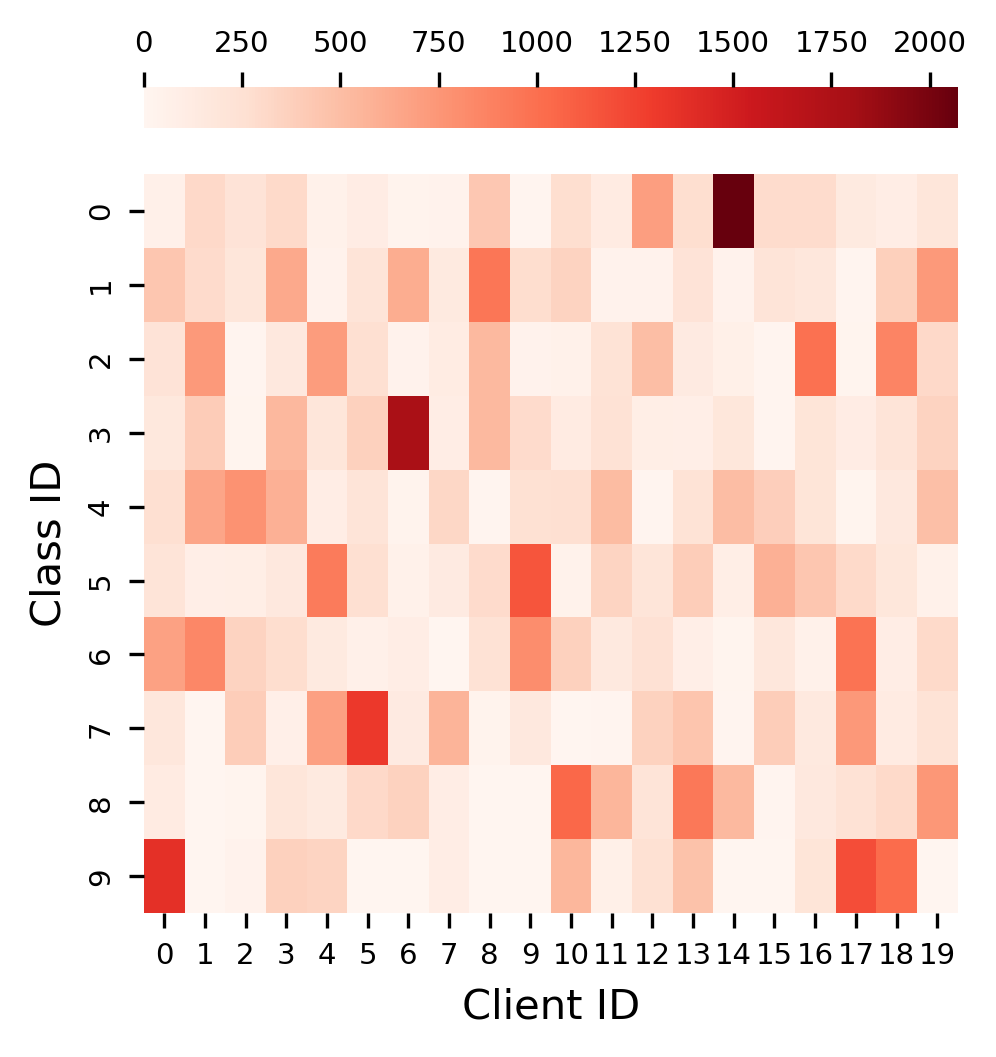}
        \caption{$\alpha=1.0$} 
    \end{subfigure}
    \caption{Data distributions for the CIFAR-10 practical heterogeneous setting.}
    \label{fig:distribution_cifar10}
    \vspace{-10pt}

\end{figure*}

\begin{figure*}[htb]
    \centering
    \begin{subfigure}[b]{0.19\textwidth}
        \includegraphics[width=\textwidth]{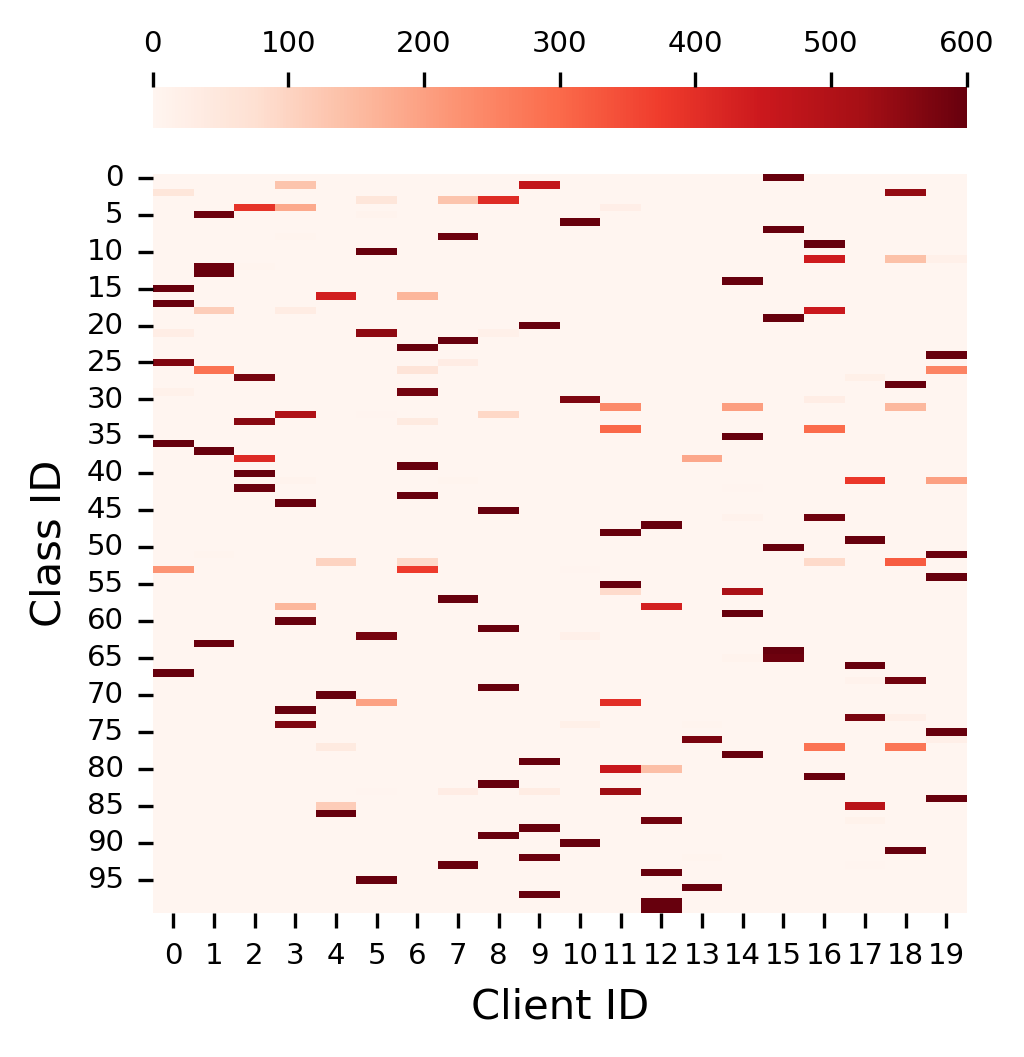}
        \caption{$\alpha=0.01$} 
    \end{subfigure}
    \quad 
    \begin{subfigure}[b]{0.19\textwidth}
        \includegraphics[width=\textwidth]{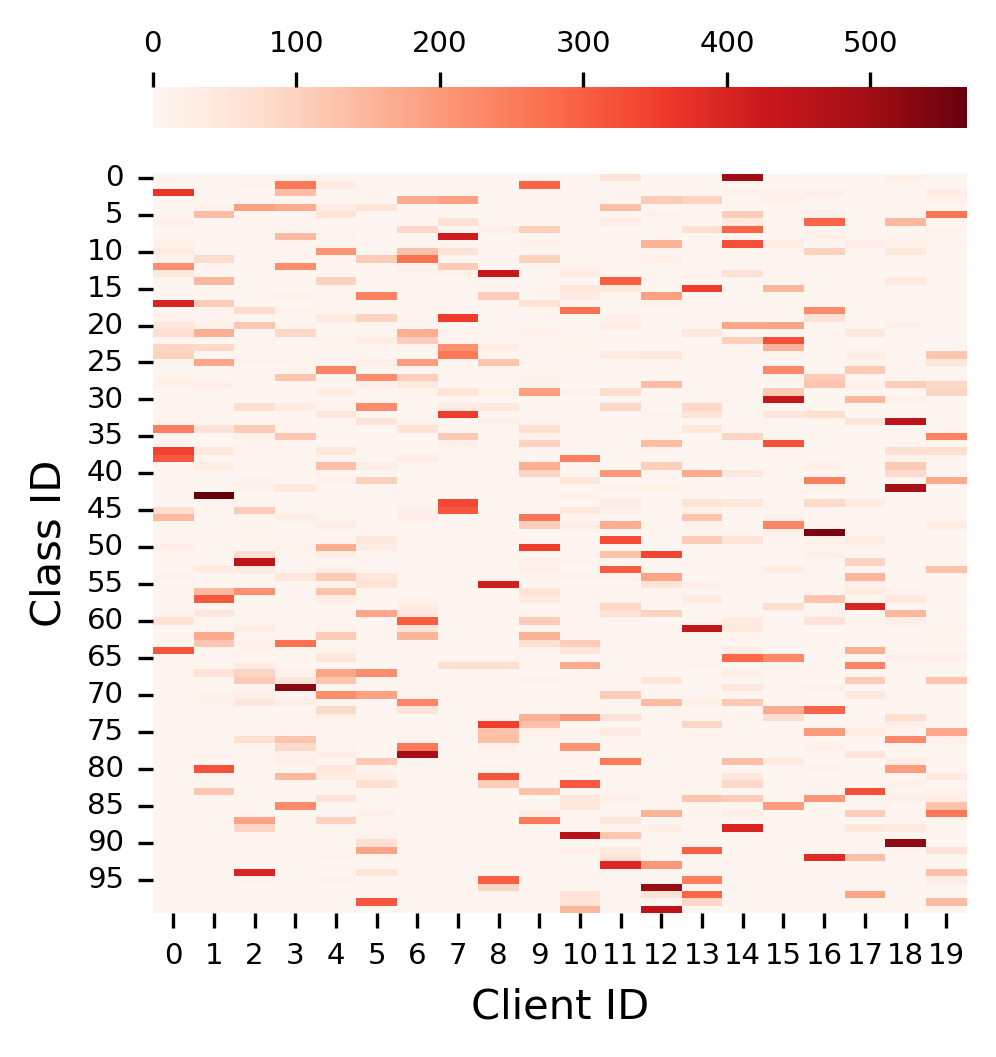}
        \caption{$\alpha=0.1$} 
    \end{subfigure}
    \quad 
    \begin{subfigure}[b]{0.19\textwidth}
        \includegraphics[width=\textwidth]{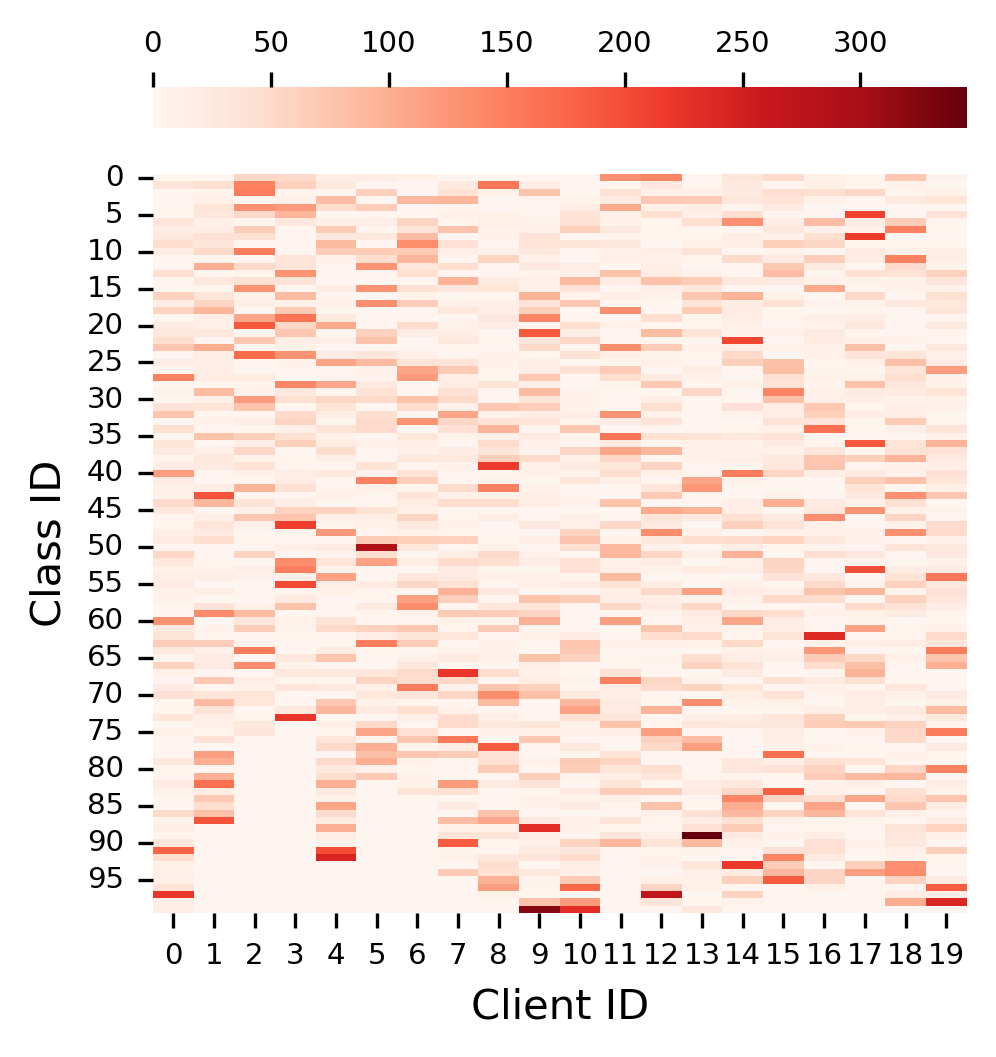}
        \caption{$\alpha=0.5$} 
    \end{subfigure}
    \quad 
    \begin{subfigure}[b]{0.19\textwidth}
        \includegraphics[width=\textwidth]{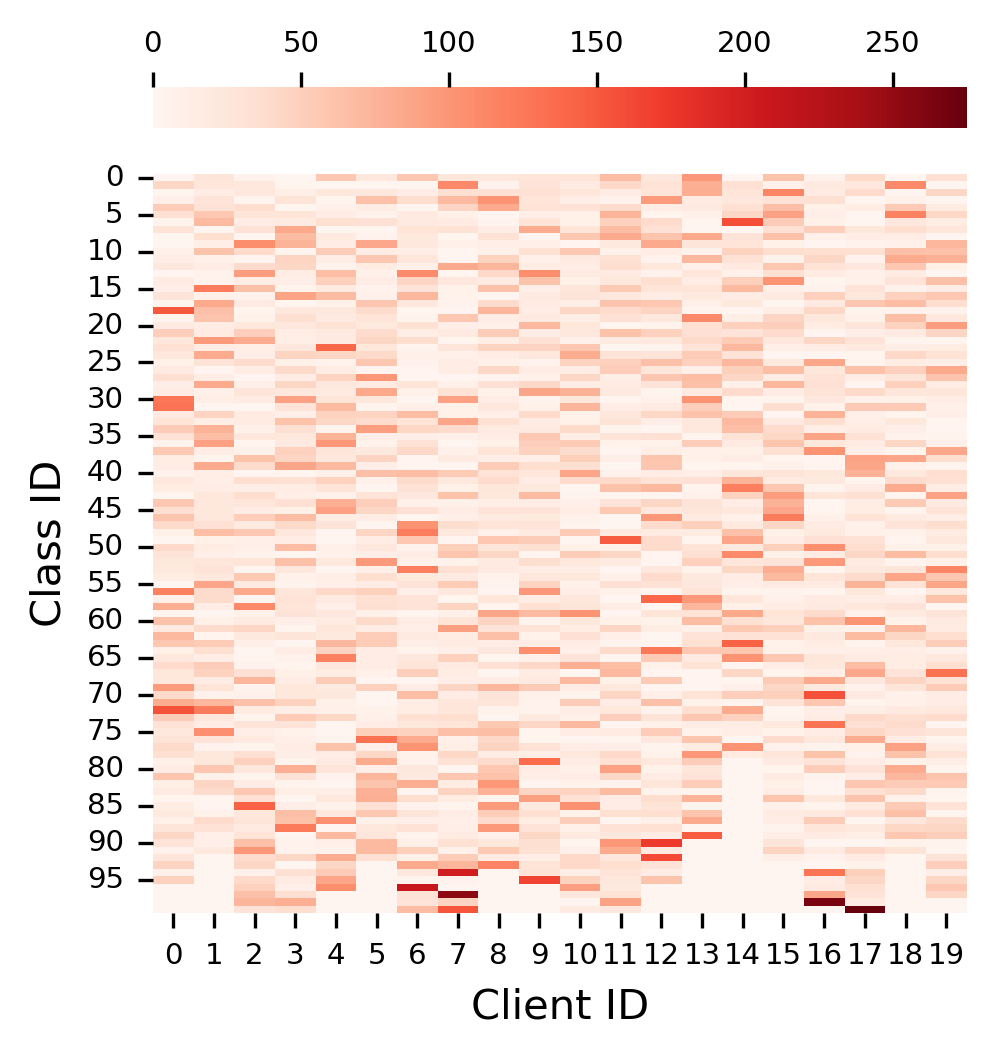}
        \caption{$\alpha=1.0$} 
    \end{subfigure}
    \caption{Data distributions for the CIFAR-100 practical heterogeneous setting.}
    \label{fig:distribution_cifar100}
    \vspace{-10pt}
\end{figure*}

\subsection*{C.3. Data Distributions of Practical Settings} 
Figures \ref{fig:distribution_cifar10} and \ref{fig:distribution_cifar100} show data distributions that vary according to different $\alpha$ values. Each cell in the heatmaps indicates the number of samples per class for each client. Increasing $\alpha$ results in decreased data heterogeneity.

\section*{D. \; Text Dataset Evaluation}
We evaluated our approach on a text dataset to test its effectiveness across various modalities. Table~\ref{table:text_dataset} shows the results for the four highest-performing algorithms, reporting test accuracy on AGNews using FastText.

\begin{table}[h]
    \centering
    \setlength{\tabcolsep}{3pt}
    {\fontsize{9}{11}\selectfont
        \begin{tabular}{rrrr}
            \toprule
            \texttt{FedAvg}        & \texttt{FedAMP}             & \texttt{FedFomo}            & \texttt{cwFedAvg} \\
            \midrule             
            79.57 $\pm$ 0.17  & 97.95 $\pm$ 0.05  & 97.93 $\pm$ 0.09  & \textbf{98.19 $\pm$ 0.01} \\ 
            \bottomrule
        \end{tabular}
    }
    \caption{Classification accuracy(\%) for AG News.}
  \label{table:text_dataset}
  \vspace{-10pt}
\end{table}

\section*{E. \; Memory Cost Comparison}
The memory efficiency of \texttt{cwFedAvg} depends on the ratio of parameter counts in the feature extractor to those in the classifier. 
Table~\ref{tab:memory_cost_comparison} demonstrates the memory cost (number of parameters in millions) differences for ResNet-18 (512 feature dimension in the penultimate layer) with varying class counts. 
While increasing class counts requires higher memory costs, our selective approach (\texttt{cwFedAvg} (Output)) significantly reduces cost compared to the non-selective approach (\texttt{cwFedAvg} (All)).
\begin{table}[ht]
  \centering
  \setlength{\tabcolsep}{4pt}
  {\fontsize{9}{11}\selectfont
  \begin{tabular}{rrrr}
    \toprule
    \# Classes & \texttt{FedAvg} & \texttt{cwFedAvg} (All) & \texttt{cwFedAvg} (Output) \\
    \midrule
    10    & 11.18     & 111.81 & \textbf{11.23} \\
    100   & 11.23     & 1122.67 & \textbf{16.36} \\
    1000  & 11.69     & 11688.42 & \textbf{524.68} \\
    \bottomrule
  \end{tabular} 
  }
  \caption{Memory cost comparison for ResNet-18.}
  \label{tab:memory_cost_comparison}
  \vspace{-10pt}
\end{table}

\section*{F. \; Visualizations of $\ell_2$-norms of Output Layer Weight Vectors}
\begin{figure*}[ht]
    \centering
    \begin{subfigure}[b]{0.18\textwidth}
        \includegraphics[width=\textwidth]{figures/Figure_6a.png}
        \caption{Data distribution of \\ clients } \label{fig:heatmap_cifar10_prac2_data}
    \end{subfigure}
    \hfill
    \begin{subfigure}[b]{0.18\textwidth}
        \includegraphics[width=\textwidth]{figures/Figure_6b.png}
        \caption{Local models of \\ \texttt{FedAvg} } 
    \end{subfigure}
    \hfill
    \begin{subfigure}[b]{0.18\textwidth}
        \includegraphics[width=\textwidth]{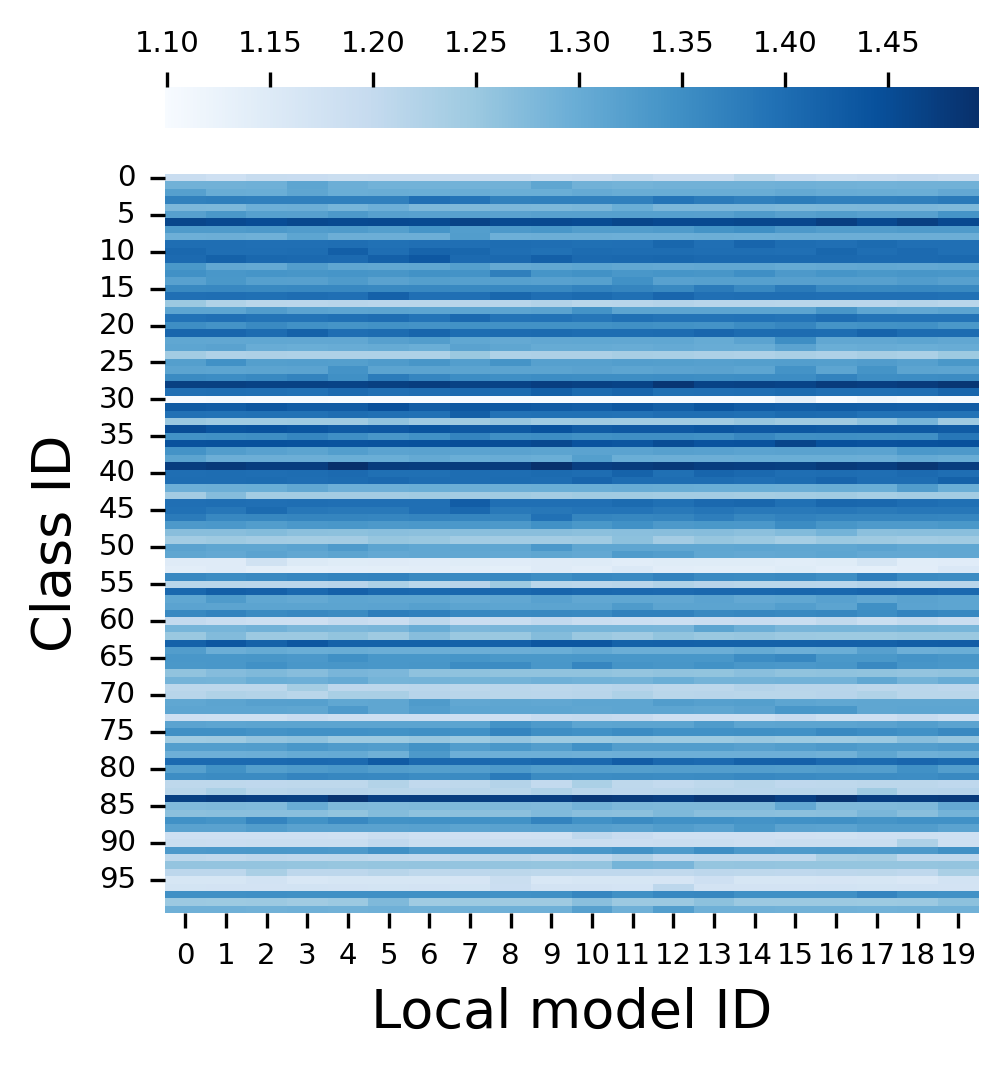}
        \caption{Local models of \\ fine-tuned \texttt{FedAvg}} \label{fig:heatmap_cifar100_prac2_FedAvg_local}
    \end{subfigure}
    \hfill
    \begin{subfigure}[b]{0.18\textwidth}
        \includegraphics[width=\textwidth]{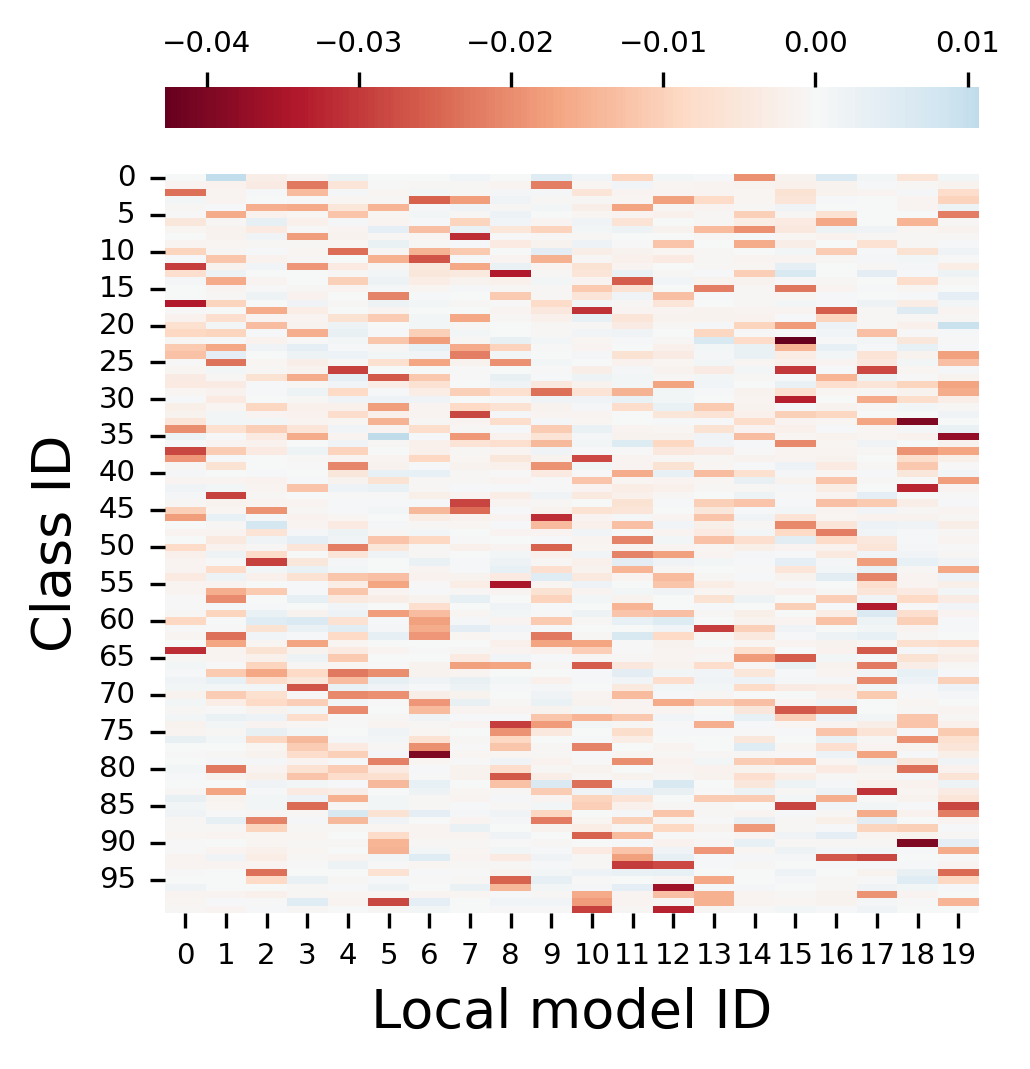}
        \caption{Difference between \\ (b) and (c) } 
    \end{subfigure}
    \hfill
    \begin{subfigure}[b]{0.18\textwidth}
        \includegraphics[width=\textwidth]{figures/Figure_6e.png}
        \caption{Local models of \\ \texttt{FedAMP} } 
    \end{subfigure}
    
    \begin{subfigure}[b]{0.18\textwidth}
        \includegraphics[width=\textwidth]{figures/Figure_6f.png}
        \caption{Local models of \\ \texttt{IFCA} } 
    \end{subfigure}
    \hfill
    \begin{subfigure}[b]{0.18\textwidth}
        \includegraphics[width=\textwidth]{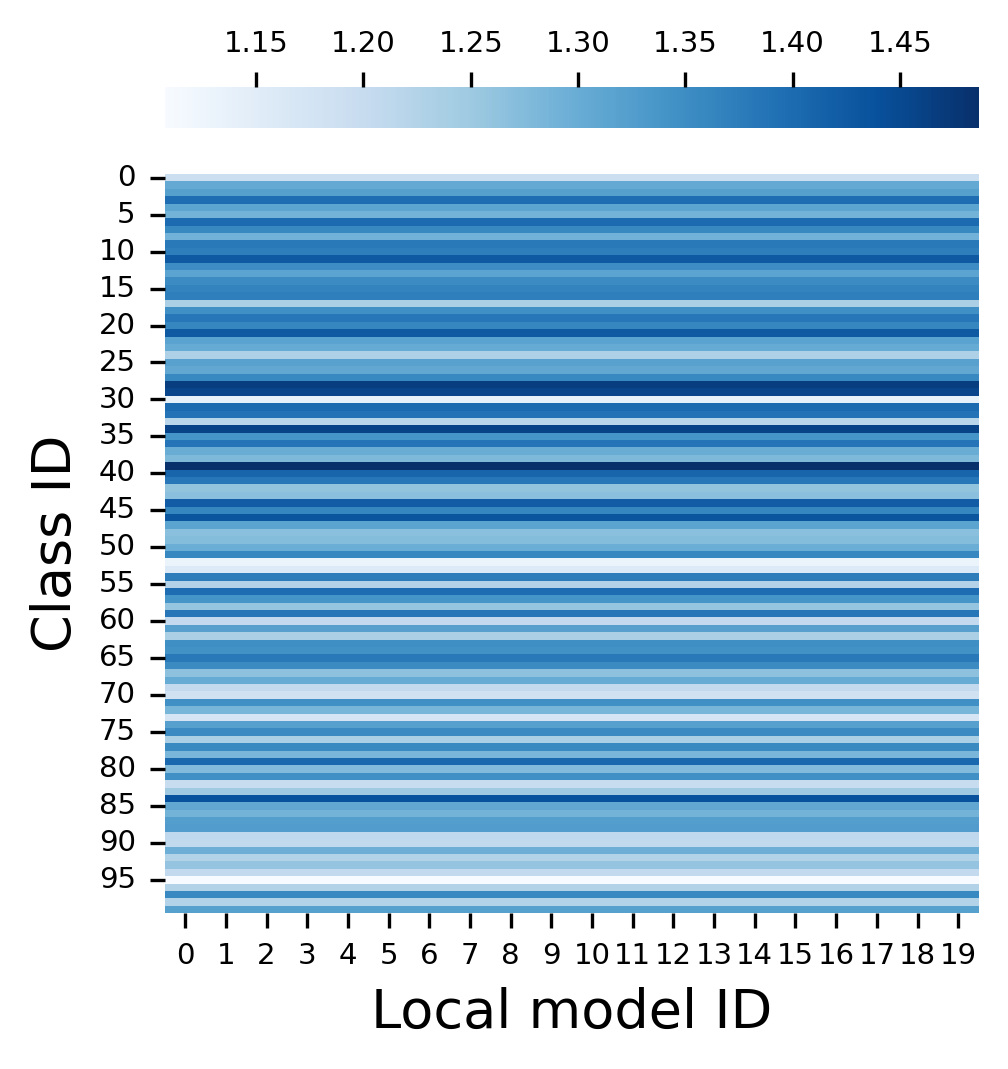}
        \caption{Local models of \\ \texttt{cwFedAvg} w/o \texttt{WDR}} \label{fig:heatmap_cifar100_prac2_cwFedAvg_local}
    \end{subfigure}
    \hfill
    \begin{subfigure}[b]{0.18\textwidth}
        \includegraphics[width=\textwidth]{figures/Figure_6h.png}
        \caption{Local models of \\ \texttt{cwFedAvg} w/ \texttt{WDR}} \label{fig:heatmap_cifar100_prac2_cwFedAvg_local_wdr}
    \end{subfigure}
    \hfill
    \begin{subfigure}[b]{0.18\textwidth}
        \includegraphics[width=\textwidth]{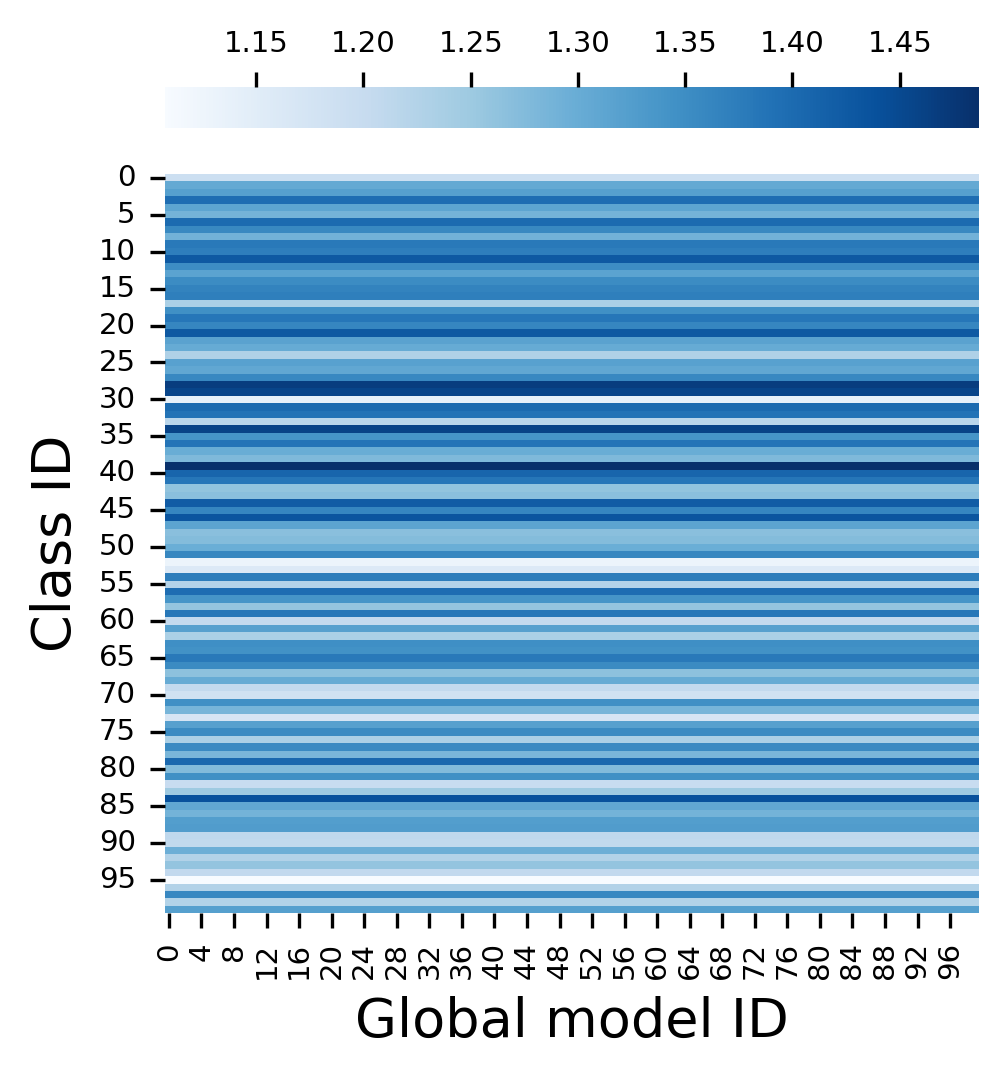}
        \caption{Global models of \\ \texttt{cwFedAvg} w/o \texttt{WDR}} \label{fig:heatmap_cifar100_prac2_cwFedAvg_global}
    \end{subfigure}
    \hfill
    \begin{subfigure}[b]{0.18\textwidth}
        \includegraphics[width=\textwidth]{figures/Figure_6j.png}
        \caption{Global models of \\ \texttt{cwFedAvg} w/ \texttt{WDR}} \label{fig:heatmap_cifar100_prac2_cwFedAvg_global_wdr}
    \end{subfigure} 
    \caption{Heatmaps for the CIFAR-100 practical heterogeneous setting. These heatmaps confirm that the CIFAR-100 practical heterogeneous setting shows very similar patterns as the CIFAR-10 pathological heterogeneous setting.
Notably, Figure \ref{fig:heatmap_cifar100_prac2_cwFedAvg_local} (\texttt{cwFedAvg} without \texttt{WDR}) closely resembles Figure \ref{fig:heatmap_cifar100_prac2_FedAvg_local}.
In contrast, Figure \ref{fig:heatmap_cifar100_prac2_cwFedAvg_local_wdr} (\texttt{cwFedAvg} with \texttt{WDR}) exhibits a pattern similar to Figure \ref{fig:heatmap_cifar10_prac2_data}, suggesting that each model has undergone personalization tailored to its possessed classes. Additionally, we visualize ten class-specific global models of \texttt{cwFedAvg} in Figures \ref{fig:heatmap_cifar100_prac2_cwFedAvg_global} (without \texttt{WDR}) and \ref{fig:heatmap_cifar100_prac2_cwFedAvg_global_wdr} (with \texttt{WDR}). As designed, each global model in Figure \ref{fig:heatmap_cifar100_prac2_cwFedAvg_global_wdr} specializes in a specific single class.}
    \label{fig:heatmap_cifar100_prac2}
\end{figure*}

\begin{figure*}[htb]
    \centering
    \begin{subfigure}[b]{0.18\textwidth}
        \includegraphics[width=\textwidth]{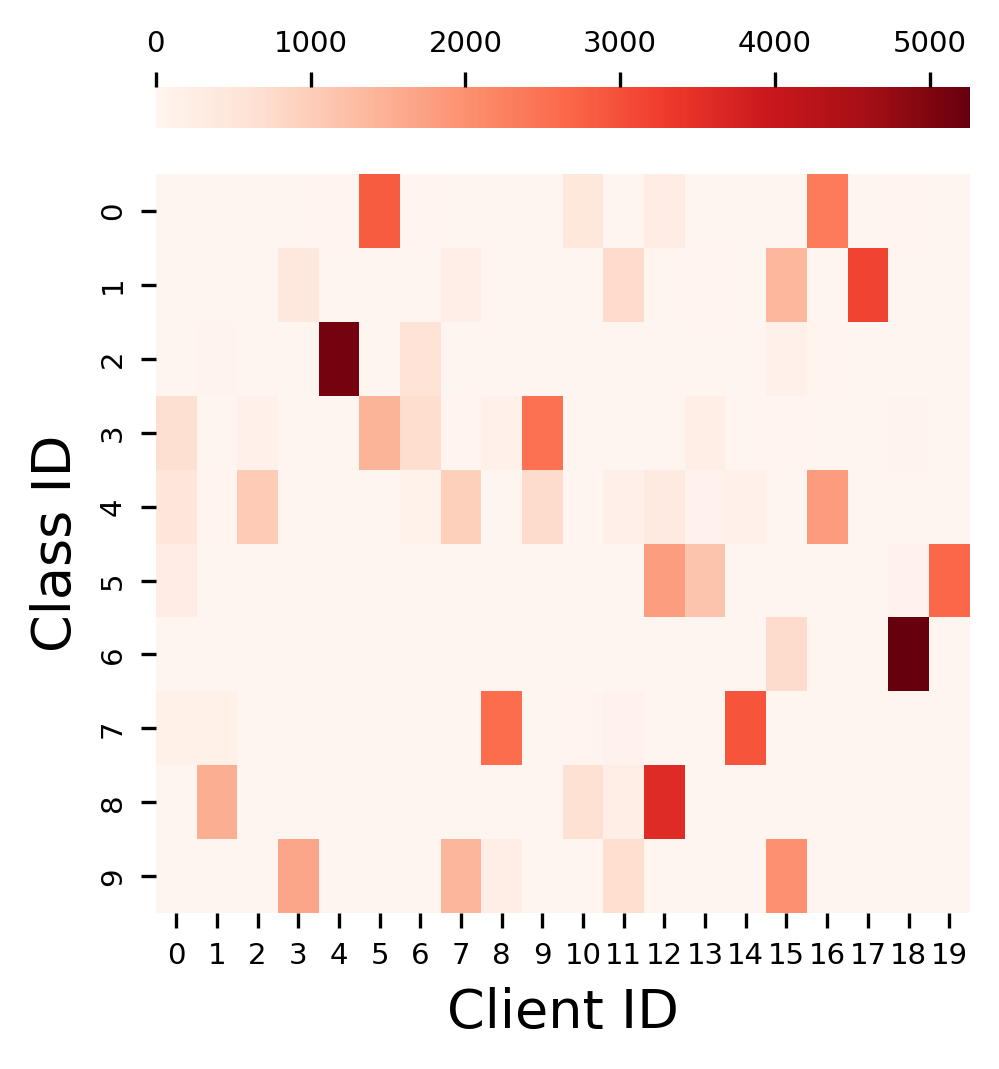}
        \caption{Data distribution of \\ clients } 
    \end{subfigure}
    \quad 
    \begin{subfigure}[b]{0.18\textwidth}
        \includegraphics[width=\textwidth]{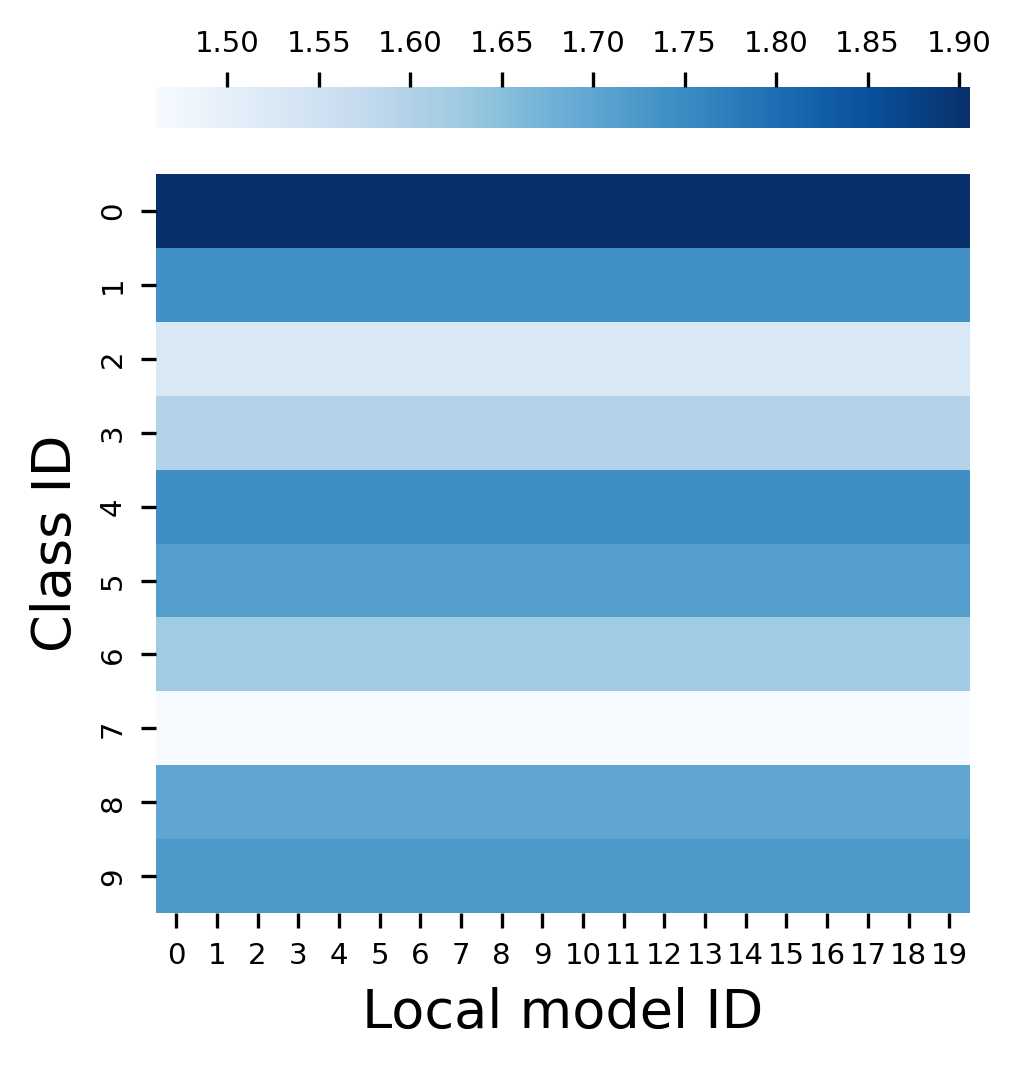}
        \caption{Local models of \\\texttt{FedAvg}} 
    \end{subfigure}
    \quad 
    \begin{subfigure}[b]{0.18\textwidth}
        \includegraphics[width=\textwidth]{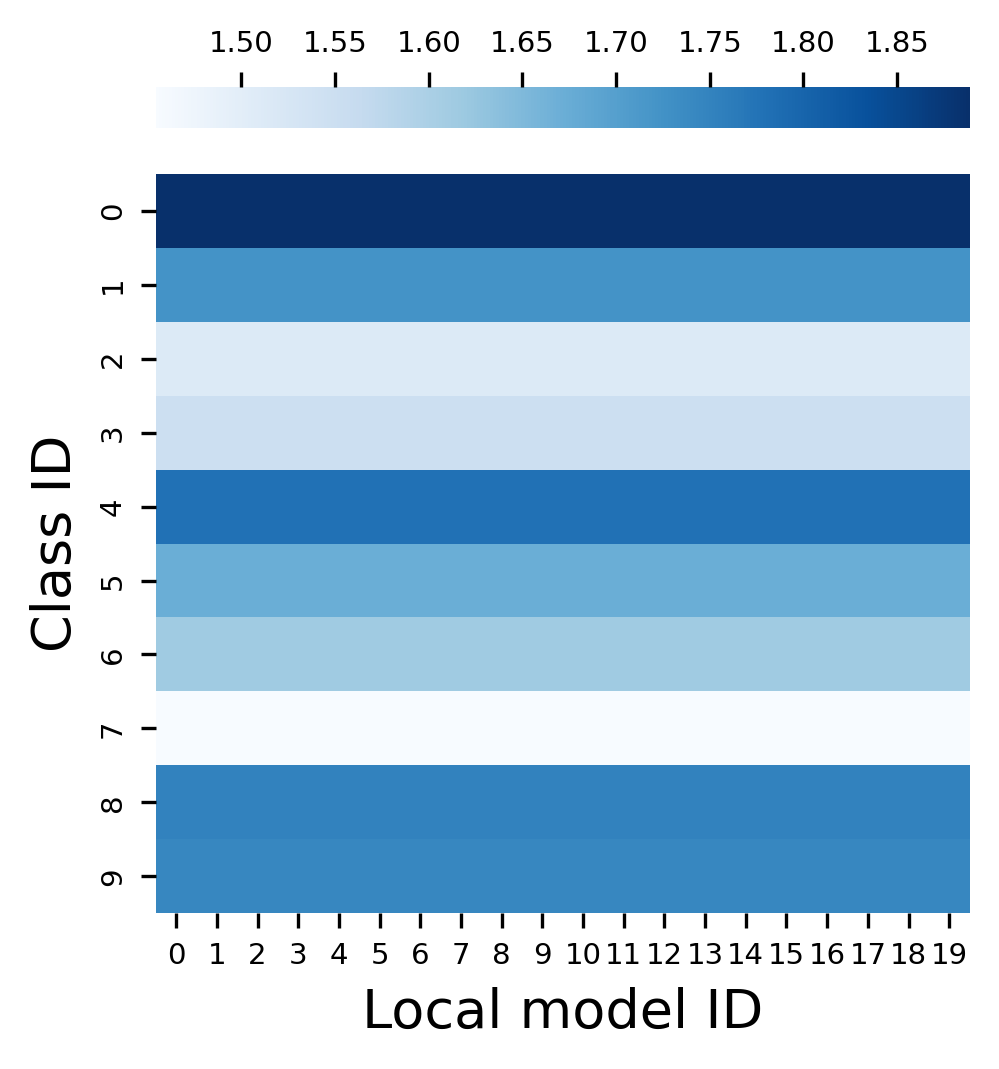}
        \caption{Local models of \\\texttt{FedProx}} 
    \end{subfigure}
    \quad 
    \begin{subfigure}[b]{0.18\textwidth}
        \includegraphics[width=\textwidth]{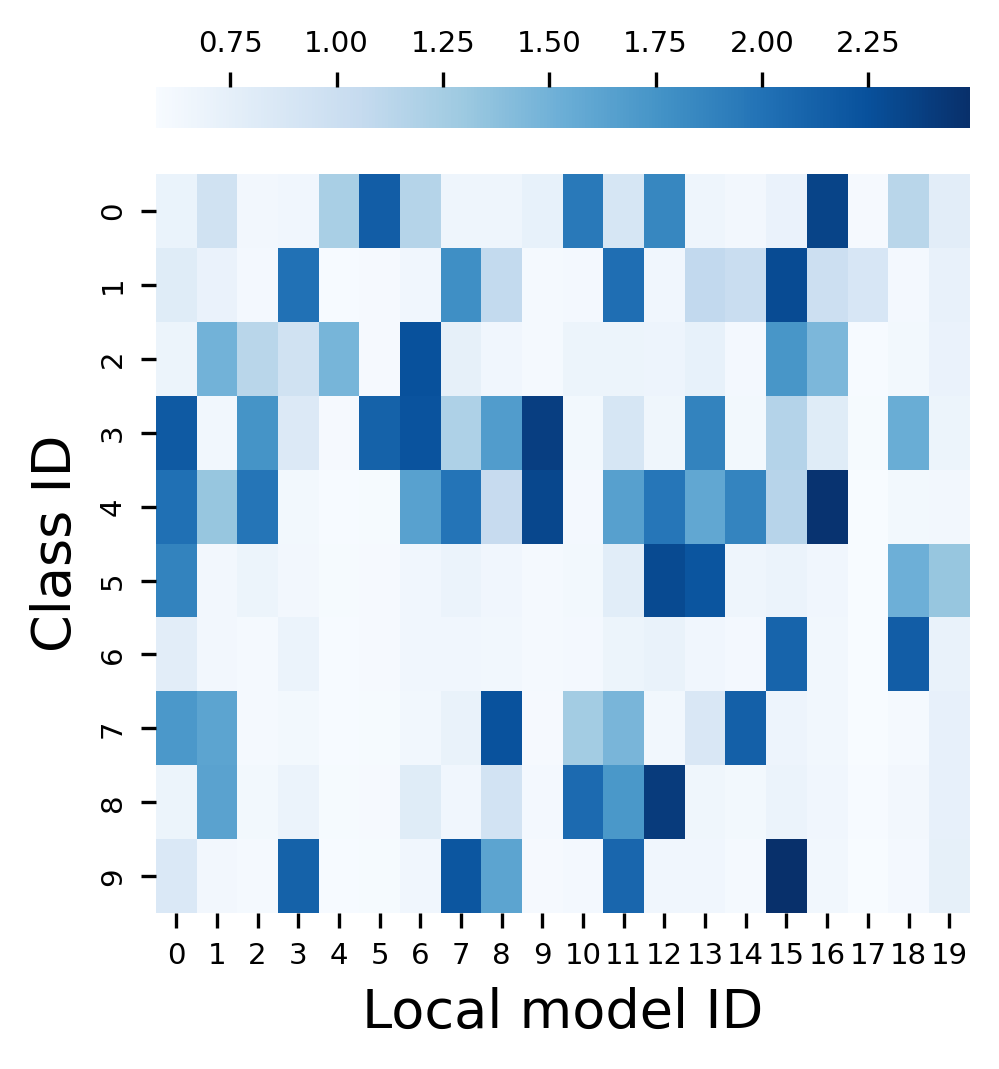}
        \caption{Local models of \\\texttt{FedAMP}} \label{fig:heatmap_cifar10_prac_fedamp}
    \end{subfigure}
    
    \begin{subfigure}[b]{0.18\textwidth}
        \includegraphics[width=\textwidth]{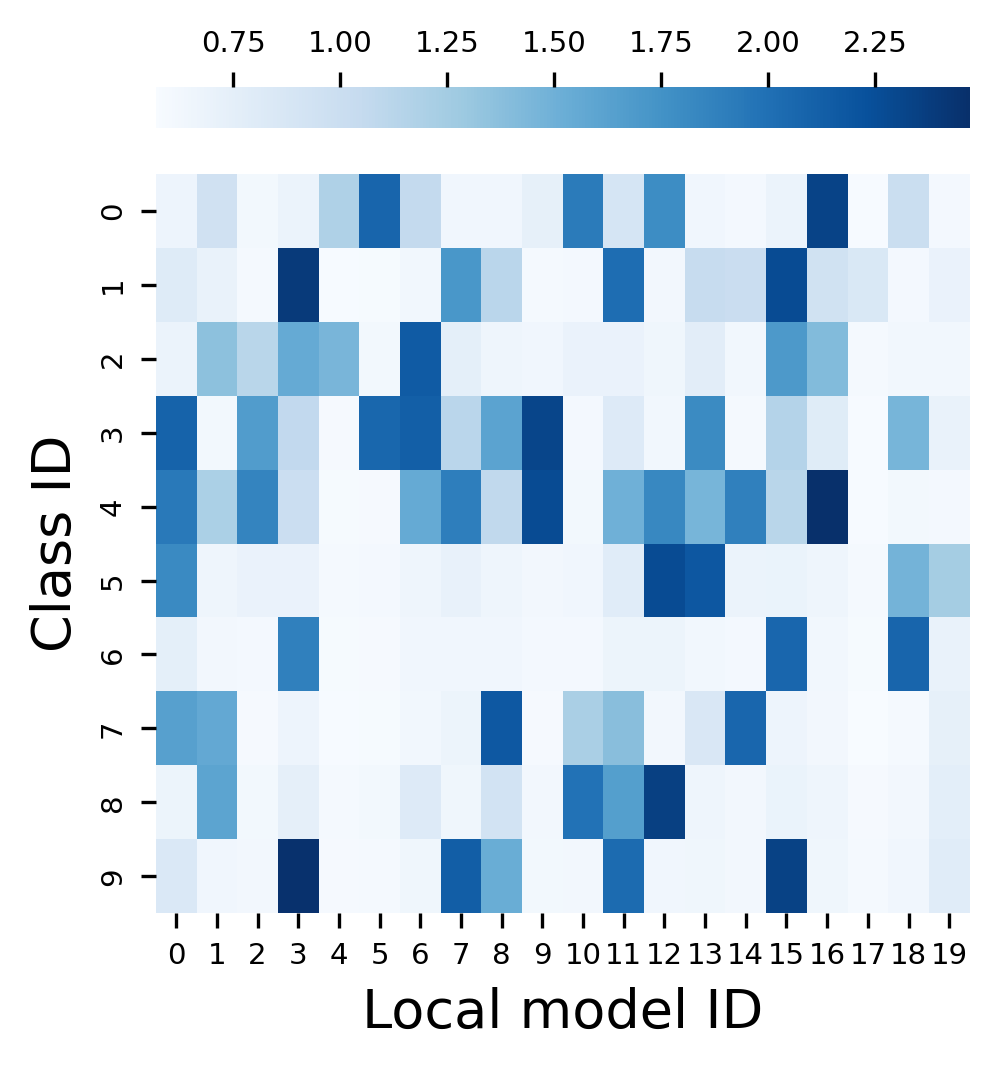}
        \caption{Local models of \\\texttt{FedFomo}} \label{fig:heatmap_cifar10_prac_fedfomo}
    \end{subfigure}
    \quad
    \begin{subfigure}[b]{0.18\textwidth}
        \includegraphics[width=\textwidth]{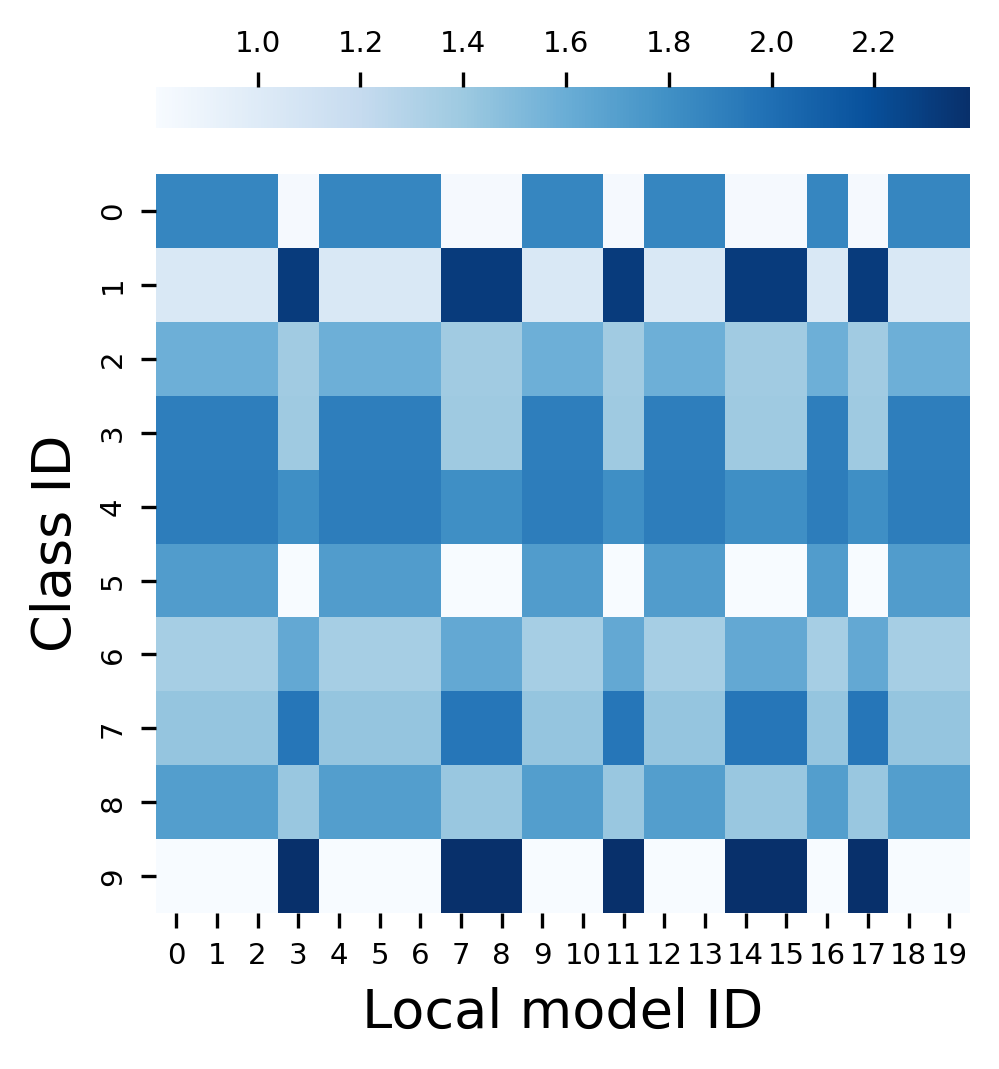}
        \caption{Local models of \\\texttt{CFL}} \label{fig:heatmap_cifar10_prac_cfl}
    \end{subfigure}
    \quad
    \begin{subfigure}[b]{0.18\textwidth}
        \includegraphics[width=\textwidth]{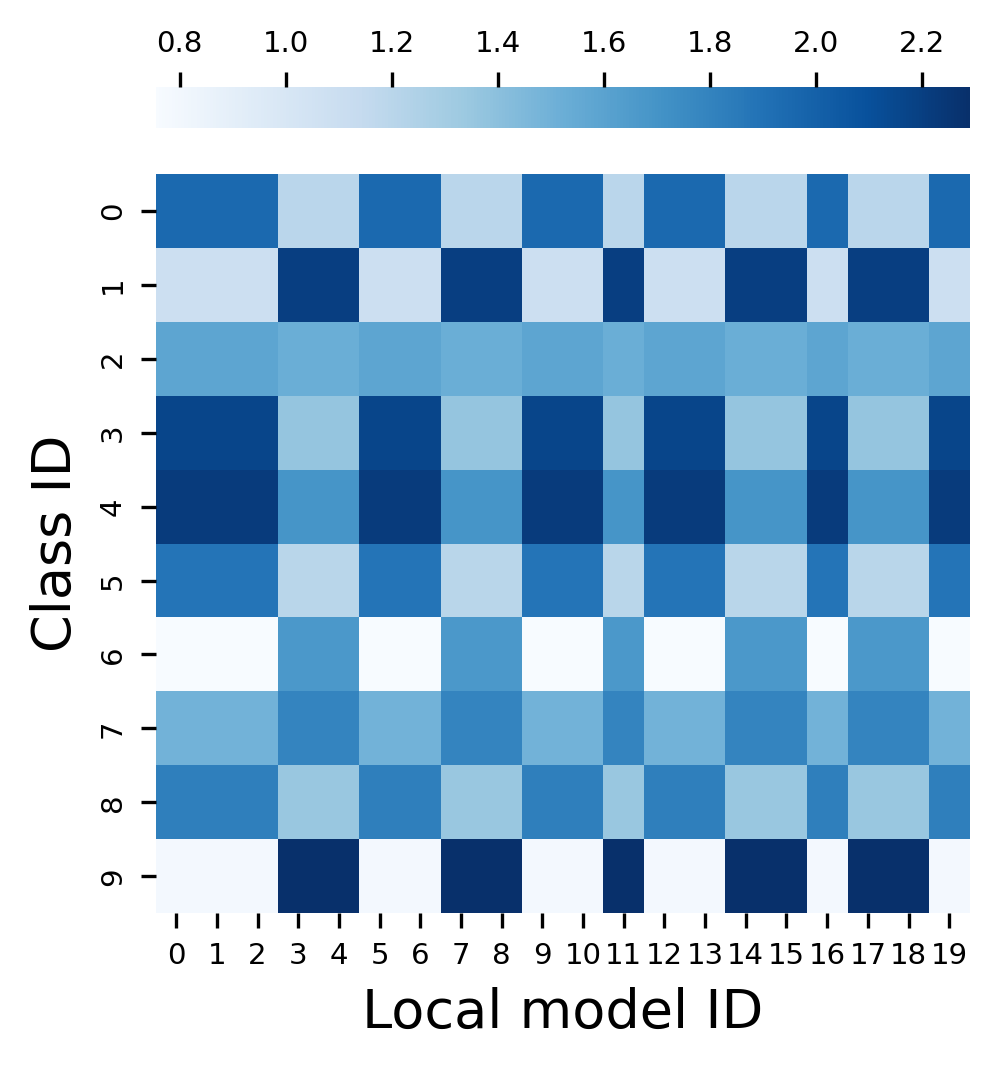}
        \caption{Local models of \\\texttt{IFCA}} \label{fig:heatmap_cifar10_prac_ifca}
    \end{subfigure}
    \quad 
    \begin{subfigure}[b]{0.18\textwidth}
        \includegraphics[width=\textwidth]{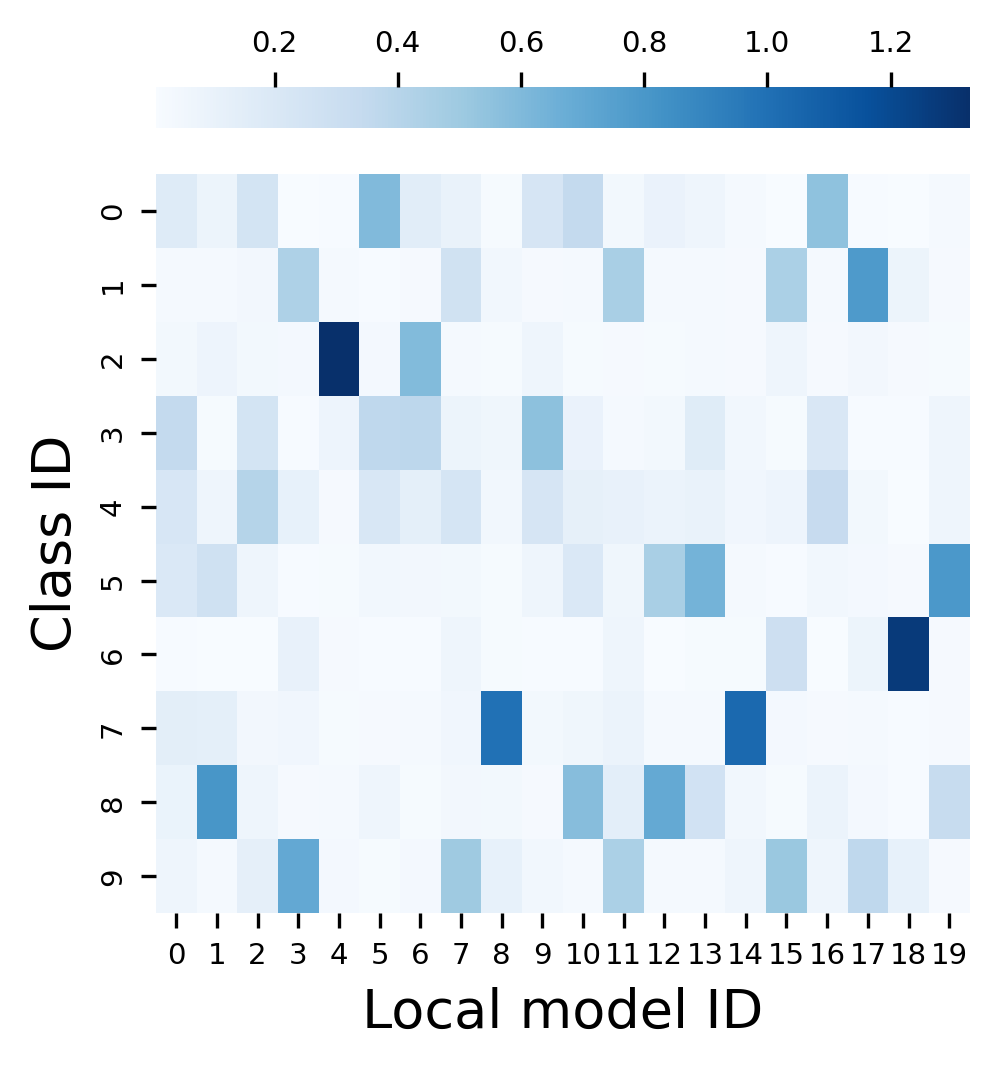}
        \caption{Local models of \\\texttt{cwFedAvg}} 
    \end{subfigure}
    \caption{Heatmaps for the CIFAR-10 practical heterogeneous setting. These heatmaps confirm the observations from the CIFAR-10 pathological heterogeneous setting in Figure \ref{fig:heatmap_cifar10_pat}. Notably, PFL methods such as \texttt{FedAMP} and \texttt{FedFomo} exhibit patterns similar to the data distribution, albeit with less pronounced similarity compared to \texttt{cwFedAvg}. Interestingly, clustering-based PFL methods, such as \texttt{CFL} and \texttt{IFCA}, exhibit distinct patterns, with two clusters evident in the heatmaps. Among the various FL and PFL approaches, \texttt{cwFedAvg} demonstrates the most similar pattern with the true data distribution, suggesting its superior capability in personalizing clients.
}
    \label{fig:heatmap_cifar10_practical}
    \vspace{-10pt}
\end{figure*}

This section explores the applicability of visualizing client $\ell_2$-norms of output layer weight vectors to the CIFAR-100 dataset, which has a significantly higher number of classes than CIFAR-10 (Figure \ref{fig:heatmap_cifar100_prac2}). Additionally, we examine whether the personalization patterns exhibited by the \texttt{cwFedAvg} method can be observed in other PFL algorithms such as \texttt{FedAMP} and \texttt{FedFomo} for CIFAR-10 practical settings (Figure \ref{fig:heatmap_cifar10_practical}). Detailed explanations are included in the figure captions.

\section*{G. \; Convergence Behavior Analysis}
\begin{figure*}[t]
    \centering
        \includegraphics[width=0.90\textwidth]{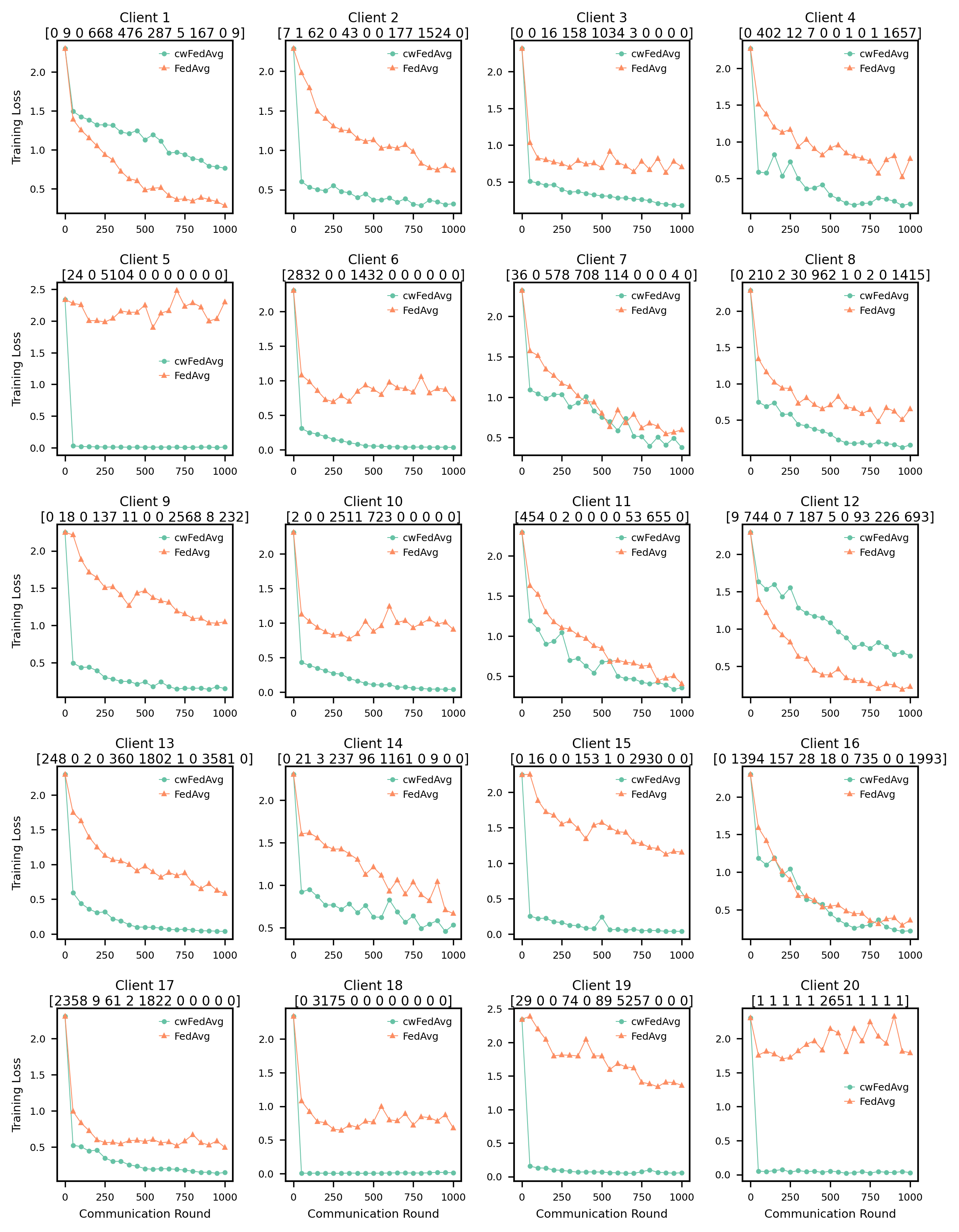}
        \caption{Comparison of Per-Client Convergence Behaviors for CIFAR-10 in Practical Settings ($\alpha=0.1$). Figures clearly show that \texttt{cwFedAvg} converges significantly faster than \texttt{FedAvg} for highly imbalanced distributions, where the number of samples per class is shown below each client ID in the line plots. This superior convergence of \texttt{cwFedAvg} is observed in clients 5, 6, 9, 10, 13, 15, 18, 19, and 20. Conversely, \texttt{FedAvg} demonstrates faster convergence in clients 1 and 12, where the data distribution is less imbalanced.} 
        \label{fig:convergence_per_client_cifar10}
\end{figure*}

\begin{figure*}[ht]
    \centering
        \includegraphics[width=0.88\textwidth]{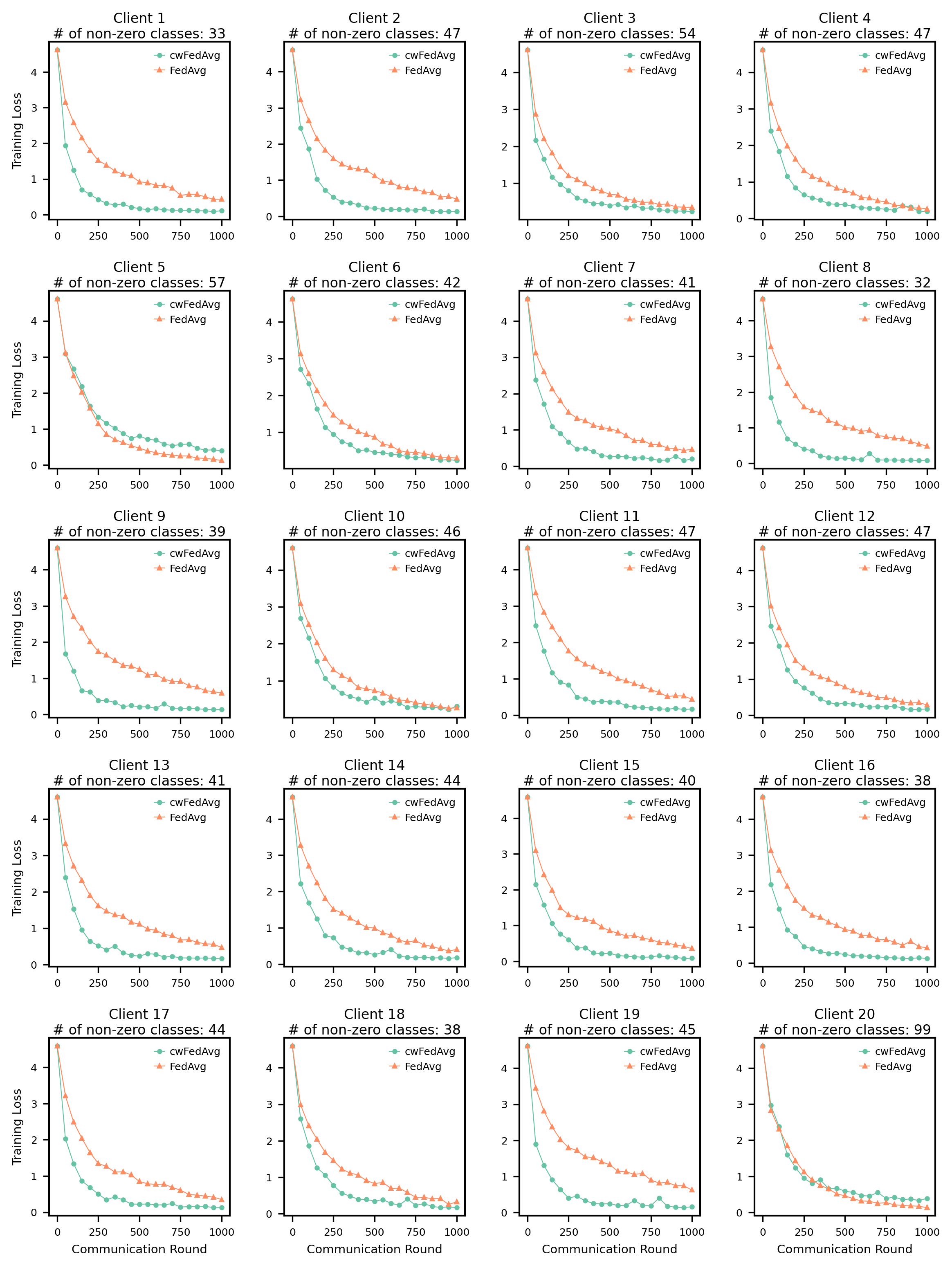}
        \caption{Comparison of Per-Client Convergence Behaviors for CIFAR-100 in Practical Settings ($\alpha=0.1$). Figures reveal convergence characteristics that align with the findings in Figure \ref{fig:convergence_per_client_cifar10}. The line plots, which display the number of non-zero classes under each client ID, demonstrate that \texttt{cwFedAvg} achieves faster convergence than \texttt{FedAvg} in highly imbalanced scenarios. Although this advantage persists across most clients, \texttt{FedAvg} shows superior convergence rates for clients 5 and 20, where data is more evenly distributed.} 
        \label{fig:convergence_per_client_cifar100}
\end{figure*}

In Figure \ref{fig:convergence}, we observe distinct average training loss patterns between \texttt{cwFedAvg} and \texttt{FedAvg}. We further examine the per-client convergence behaviors to analyze how different client data distributions affect the training dynamics of the two methods in Figure \ref{fig:convergence_per_client_cifar10} and \ref{fig:convergence_per_client_cifar100}. Detailed explanations are included in the figure captions.

\end{document}